\documentclass[aos,preprint]{imsart}

\RequirePackage{amsthm,amsmath,amsfonts,amssymb,mathtools,algorithm,algpseudocode,bbm}
\RequirePackage[numbers]{natbib}
\RequirePackage[colorlinks,citecolor=blue,urlcolor=blue]{hyperref}%% uncomment this for coloring bibliography citations and linked URLs
\RequirePackage{graphicx, color, xcolor}%% uncomment this for including figures

\startlocaldefs
\theoremstyle{plain}
\newtheorem{theorem}{Theorem}
\newtheorem{lemma}[theorem]{Lemma}
\theoremstyle{definition}
\newtheorem{definition}{Definition}
\newtheorem{assumption}{Assumption}

\newtheorem{remark}{Remark}
\newcommand\mbe{\mathbb E}
\newcommand\mcr{\mathcal R}
\newcommand\mcb{\mathcal B}
\newcommand\mbp{\mathbb P}
\newcommand\pa{\mathrm{pa}}
\newcommand{\indep}{\mathrel{\perp\!\!\!\perp}}
\newcommand{\notindep}{\mathrel{\not\!\perp\!\!\!\perp}}

\newcommand{\SuppNumbering}{%
  \setcounter{section}{0}
  \setcounter{theorem}{0}
  \setcounter{definition}{0}
  \setcounter{remark}{0}
  \setcounter{equation}{0}
  \setcounter{algorithm}{0}

  \renewcommand{\thesection}{S\arabic{section}}
  \renewcommand{\thetheorem}{\thesection.\arabic{theorem}}
  \renewcommand{\thedefinition}{\thesection.\arabic{definition}}
  \renewcommand{\theremark}{\thesection.\arabic{remark}}
  \renewcommand{\thealgorithm}{\thesection.\arabic{algorithm}}
  \renewcommand{\theequation}{\thesection.\arabic{equation}}

  \renewcommand{\theHsection}{supp.\arabic{section}}
  \renewcommand{\theHtheorem}{supp.\thesection.\arabic{theorem}}
  \renewcommand{\theHdefinition}{supp.\thesection.\arabic{definition}}
  \renewcommand{\theHremark}{supp.\thesection.\arabic{remark}}
  \renewcommand{\theHalgorithm}{supp.\thesection.\arabic{algorithm}}
  \renewcommand{\theHequation}{supp.\thesection.\arabic{equation}}
}

\endlocaldefs

\begin{document}

\begin{frontmatter}
\title{Causal Bandit over Unknown Graphs: Upper Confidence Bounds with Backdoor Adjustment}
%\title{A sample article title with some additional note\thanksref{T1}}
%\runtitle{???}
%\thankstext{T1}{A sample of additional note to the title.}

\begin{aug}
%%%%%%%%%%%%%%%%%%%%%%%%%%%%%%%%%%%%%%%%%%%%%%%
%% Only one address is permitted per author. %%
%% Only division, organization and e-mail is %%
%% included in the address.                  %%
%% Additional information such as            %%
%% identifying the corresponding author must %%
%% be included in in the Acknowledgments     %%
%% section if necessary.                     %%
\author[A]{\fnms{Yijia}~\snm{Zhao}\ead[label=e1]{yijiazhao@ucla.edu}} 
\and
\author[A]{\fnms{Qing}~\snm{Zhou}\ead[label=e2]{zhou@stat.ucla.edu}}

%%%%%%%%%%%%%%%%%%%%%%%%%%%%%%%%%%%%%%%%%%%%%%
%% Addresses                                %%
%%%%%%%%%%%%%%%%%%%%%%%%%%%%%%%%%%%%%%%%%%%%%%
\address[A]{Department of Statistics and Data Science, University of California, Los Angeles
\printead[presep={\\ }]{e1,e2}}

%\address[B]{Department of Statistics, University of California, Los Angeles\printead[presep={,\ }]{e2}}

\end{aug}

\begin{abstract}
The causal bandit problem seeks to identify, through sequential experimentation, an intervention that maximizes the expected reward in a causal system modeled by a directed acyclic graph (DAG). Existing methods typically assume that the causal graph is known or impose restrictive structural assumptions. In this paper, we study causal bandit problems when the causal graph is unknown.
We first consider Gaussian DAG models without latent confounders. By combining observational and experimental data collected sequentially during the bandit process, we identify candidate backdoor adjustment sets for each intervention arm. These sets enable estimation of causal effects and construction of upper confidence bounds that integrate information from both data sources. Based on these estimates, we propose a new algorithm, termed backdoor-adjustment upper confidence bound (BA-UCB), for sequential intervention selection.
We establish finite-time upper bounds on the cumulative regret of BA-UCB, showing improved rates and substantially relaxed dependency on the number of intervention arms compared to standard multi-armed bandit methods. We further extend the methodology and theoretical guarantees to settings with latent confounders, where the observed variables are modeled by an acyclic directed mixed graph. Simulation studies demonstrate that BA-UCB achieves substantially lower cumulative regret and favorable computational efficiency relative to existing approaches.
\end{abstract}


\end{frontmatter}
%%%%%%%%%%%%%%%%%%%%%%%%%%%%%%%%%%%%%%%%%%%%%%
%% Please use \tableofcontents for articles %%
%% with 50 pages and more                   %%
%%%%%%%%%%%%%%%%%%%%%%%%%%%%%%%%%%%%%%%%%%%%%%
%\tableofcontents

%%%%%%%%%%%%%%%%%%%%%%%%%%%%%%%%%%%%%%%%%%%%%%
%%%% Main text entry area:

\section{Introduction}
\label{sec:intro}

\subsection{Multi-armed bandit}

The multi-armed bandit (MAB) problem has been extensively studied in statistics, operations research, and machine learning~\citep{auer2002MAB,lai1985asymptotically,berry1985bandit,bubeck2012regret,lattimore2020bandit}. It is formulated as a sequential decision-making process in which a player selects one arm at each round, based on past actions and observed rewards. The objective is to identify the arm with the highest expected reward while minimizing cumulative regret incurred from selecting sub-optimal arms, through a careful trade-off between exploration and exploitation. This problem has broad applications in online recommendation systems~\citep{li2010contextual}, adaptive clinical trials~\citep{bather1985allocation,villar2015multi}, and dynamic pricing~\citep{besbes2009dynamic}, among others.

Under the classic setting of the MAB problem~\citep{auer2002MAB, lai1985asymptotically}, the action set $\mathcal A$ consists of $K$ independent arms and each arm $a\in \mathcal A$ corresponds to a real-valued distribution of the reward variable. It is an overly simplified setting in which the entire system is regarded as a black box, relying solely on reward feedback to guide learning. Various algorithms have been developed in this setting, among which the upper confidence bound (UCB) algorithm~\citep{auer2002MAB} and Thompson sampling~\citep{thompson1933} are two classical examples. However, real-world decision-making processes often involve additional complexity, motivating numerous further developments of the MAB problem. For example, the contextual bandit problem incorporates observed contextual features before each decision, allowing personalization and adaptive learning~\citep{langford2008}; the combinatorial bandit focuses on actions formed by combinations of basic arms, leading to exponentially large action spaces~\citep{cesa2012combinatorial}; and the budgeted bandit associates each action with a cost and requires the player to operate under a fixed budget~\citep{tran2010epsilon}. 

\subsection{Causal bandit}

A major source of complexity in real-world decision-making stems from the causal dependency among variables. In many domains, such as healthcare, social sciences, and economics, arms and rewards are connected through underlying causal mechanisms, typically represented by a causal directed acyclic graph (DAG). In this causal bandit (CB) framework~\citep{lattimore2016causal}, a causal DAG specifies the joint distribution of all variables in the system, including the reward. Interventions on variables correspond to arms, allowing the causal bandit problem to be cast naturally as a multi-armed bandit task. Unlike the standard MAB setting, however, leveraging the additional causal structure can enable more efficient identification of the optimal intervention.

Most existing work on causal bandit relies on full or partial knowledge of the underlying causal graph to achieve improvements over standard MAB algorithms. For instance, Lattimore et al.~\cite{lattimore2016causal}, Sen et al.~\cite{sen2017identifying}, and Yabe et al.~\cite{yabe2018causal} study the causal bandit problem assuming complete knowledge of the DAG and absence of unobserved confounders. Although Lee and Bareinboim~\cite{lee2018structural} and Maiti et al.~\cite{maiti2022causal} consider scenarios with unobserved confounders, they still assume that the full graph is known. Other approaches further impose structural restrictions while requiring the graph to be provided, such as the budgeted bandit under no-backdoor graphs~\citep{nair2020budgeted} and the linear causal bandit~\citep{lu2020regret}. More recent work in this category includes causal bandits for linear structural equation models~\citep{varici2023} and causal bandits with general causal models and soft interventions~\citep{yan2024}, both of which continue to assume a known causal graph. However, in practice, the underlying causal DAG is rarely known, which severely limits the application of these methods.

A few recent causal bandit methods have been developed without assuming the underlying causal graph is provided. For discrete variables, the optimal intervention is always among the parents of the reward variable~\citep{lee2018structural}. A natural approach is therefore to first identify the parent set of the reward variable and then apply MAB algorithms with a reduced action set consisting of only the identified parents. Many recent works follow this strategy, including the central node-upper confidence bound (CN-UCB) algorithm~\citep{lu2021causal} and the randomized parent search (RAPS) algorithm~\citep{konobeev2023}. When the variables have continuous domains, the problem is significantly more challenging, since the optimal intervention is not necessarily among the parents of the reward. De Kroon et al.~\cite{dekroon2022causal} tackle the problem for discrete and Gaussian models. However, their method relies on the existence of a set of non-intervenable variables that d-separate the interventions and the reward variable, and no theoretical guarantee on the cumulative regret was provided. Huang and Zhou~\cite{Jireh2022BBB} propose a Bayesian framework that utilizes joint inferences from experimental and observational data without additional assumptions on the underlying graph structure. But the Bayesian framework requires updating posterior probabilities over all possible parent sets, which is extremely computationally intensive and becomes impractical even for moderately sized action sets. Yan and Tajer~\citep{yan2024linear} propose a graph-agnostic linear causal bandit (GA-LCB) algorithm for linear structural equation models with unknown graph structure, assuming no latent confounders. Their approach emphasizes learning the graph structure and model parameters prior to reward evaluation, which incurs additional computational cost and may be unnecessary for identifying the optimal intervention.

%In contrast, our work assumes no prior knowledge of the causal graph and aims to identify the variable with the largest causal effect on the reward. This objective not only targets maximizing the reward, but also facilitates a better understanding of the system, which is important for resource allocation and policy-making, especially when only limited interventions are feasible. 

\subsection{Contributions of this work}

 To motivate our work, we revisit the farming example constructed by Lattimore et al.~\cite{lattimore2016causal}. Suppose a farmer seeks to maximize crop yield and knows that the outcome is affected by temperature, moisture level, and soil nutrient levels, but the causal relationships among these factors are unknown. Due to limited time and budget, she can intervene on at most one factor per season: for example, using artificial heating or shading to control temperature, irrigation or rain covers to control moisture, or adjusting fertilizer composition to modify nutrient levels. Each intervention is costly and yields limited experimental data per season. In contrast, historical observational data, collected without interventions, may be readily available. An important practical question is whether such observational data can be leveraged to help identify the variable with the largest causal impact on crop yield, thereby reducing the need for costly experimentation.
Motivated by this example, our objective is to identify the variable with the largest causal effect on the reward, and we therefore consider the set of arms to be a collection of atomic interventions. In such continuous domains, the optimal intervention is not necessarily associated with a parent of the reward node, making it unclear which aspects of the underlying causal structure should be learned and how such information can be used to reduce the action space. Further challenges include how to efficiently integrate observational data to identify the most influential variable and how to accommodate potential latent confounders.

To address these challenges, we develop the Backdoor Adjustment Upper Confidence Bound (BA-UCB) algorithm, which identifies the optimal intervention without requiring prior knowledge of the causal graph or imposing restrictive structural assumptions. We first focus on settings in which all observed variables, including the reward, follow a Gaussian DAG model. Rather than attempting to recover the full DAG, BA-UCB exploits both observational and experimental data to identify candidate adjustment sets and integrates information from the two data sources to construct upper confidence bounds for each arm within a sequential decision-making framework. We then extend BA-UCB to settings with latent confounders, in which the observed variables are modeled by an acyclic directed mixed graph (ADMG) containing both directed and bidirected edges. In this case, valid adjustment sets may not exist for all arms, introducing additional methodological challenges.

We establish finite-time upper bounds on the cumulative regret of the BA-UCB algorithm. Specifically, suppose that $m$ experimental samples and $n$ observational samples are collected at each round. In the absence of latent confounders, and assuming a sufficiently large amount of prior observational data, we show that the worst-case cumulative regret is bounded by
$O(\sqrt{{T\log T}/{(m+n})})$,
where $T$ denotes the total number of rounds. In contrast, the regret bound for the standard MAB algorithm is $O(\sqrt{KT\log T})$~\citep{russo2014learning}, where $K=|\mathcal{A}|$ is the number of arms. Even if $m+n$ experimental samples are generated per round, the standard MAB regret bound remains at least
$O(\sqrt{KT\log T/(m+n)})$. One sees that BA-UCB substantially weakens the dependence of cumulative regret on $K$ and effectively incorporates observational data, which are typically much less costly to obtain than experimental data. When latent confounders are present, we show that, with a sufficiently large amount of prior observational data, the cumulative regret of BA-UCB is bounded by
$O(\sqrt{T\log T/(m+n)})+O(\sqrt{K_1T\log T/m})$, where $K_1$ denotes the number of variables for which a valid adjustment set does not exist or cannot be identified. When $K_1$ is small, the dependence of the regret on the total number of arms $K$ is again substantially reduced, highlighting the benefit of incorporating observational data into sequential decision-making. Finally, through extensive simulation studies, we demonstrate that BA-UCB achieves lower cumulative regret than the standard UCB algorithm~\citep{auer2002MAB} and the CN-UCB algorithm~\citep{lu2021causal}, while being computationally more efficient than the Bayesian backdoor bandit algorithm~\citep{Jireh2022BBB}.

The remainder of the article is organized as follows. Section~\ref{sec:setup} introduces the causal bandit framework under Gaussian DAG models without latent confounders. Section~\ref{sec:alg} presents the BA-UCB algorithm, and Section~\ref{sec:regret} develops its finite-sample regret analysis. Section~\ref{sec:experiments} reports empirical results based on simulated data. Section~\ref{sec:confounder_case} extends the BA-UCB framework to settings with latent confounders. Section~\ref{sec:discussion} concludes with a summary of the main findings and directions for future work. Proofs and some technical details are provided in the supplementary materials.

\section{Problem formulation}
\label{sec:setup}

In a causal graphical model~\citep{pearl2009causality}, the causal relationships among a set of random variables $\mathcal X=\{X_1,\cdots, X_p\}$ is represented by a DAG $\mathcal G$ over $p$ nodes $V=\{1,\cdots, p\}$, each corresponding to one random variable. %, and a joint probability distribution $P$ over $\mathcal X$ that factorizes over $\mathcal G$. 
A directed edge $i\to j$ in the causal DAG $\mathcal G$ means that $i$ is a direct cause of $j$, in which case $i$ is called a parent of $j$ and $j$ is called a child of $i$. The parent set of $j$, denoted as $\mathrm{pa}(j)$, is the set of nodes that have an edge directed into node $j$, i.e., $\mathrm{pa}(j)=\{i\in V: i\to j\}$. Similarly, the child set of $ j$ is denoted as $\mathrm{ch}(j)=\{i\in V: j\to i\}$. Each child-parent relationship in $\mathcal G$ is modeled by a function $X_j=f_j(X_{\mathrm{pa}(j)},\varepsilon_j)$, $j=1,\cdots,p$, where $\varepsilon_1,\cdots,\varepsilon_p$ are exogenous errors.

Through Section~\ref{sec:experiments}, we focus on Gaussian DAGs whose functional relationships are given by a linear structural equation model (SEM),
\begin{equation}\label{eq:GaussianBN}
    X_j=\sum_{i\in \mathrm{pa}(j)} \beta_{ij}X_i + \varepsilon_j,\quad j=1,\cdots,p,
\end{equation}
where $\varepsilon_j\overset{\mathrm{indep.}}{\sim}\mathcal N(0,\omega_j^2)$ with $\omega_j^2>0$ for all $ j\in V$ and $\beta_{ij}\neq 0$ for all $i\in\mathrm{pa}(j)$. %We consider the problem of finding the variable with the largest causal effect on the reward variable designated as $Y = X_p$. Given our interest in finding the single variable with largest causal effect in magnitude on $Y$, we only need to consider atomic interventions. 
Without loss of generality, the reward variable is designated as $Y = X_p$ (noting that the nodes are not sorted). The causal effect of $X_i$ on $Y$ is defined via an atomic intervention that 
%An atomic intervention is an intervention on a single node $X_i\in\mathcal X$ which 
sets the variable $X_i$ to a fixed value $x_i$, denoted as $do(X_i=x_i)$. In general, the causal effect $\gamma_i$ of $X_i$ on $Y$ is determined by the mapping $x_i\mapsto \mathbb P[Y|do(X_i=x_i)]$. For Gaussian DAGs, %the causal effect $\gamma_i$ is defined as
\begin{equation}
   \gamma_i:=\frac{\partial }{\partial x_i}\mathbb E(Y|do(X_i=x_i))
\end{equation}
and by linearity $\mathbb E(Y|do(X_i=x_i))=\gamma_ix_i$. Thus, from the intervention $do(X_i=x_i)$ for any $x_i\neq 0$, we can determine the causal effect $\gamma_i=\mathbb E(Y|do(X_i=x_i))/x_i$. %i.e., any intervention $do(X_i=x_i)$ with $x_i\neq 0$ can be rescaled to have the same causal effect as $do(X_i=\pm 1)$. 
Without loss of generality, we only consider $do(X_i=\pm 1)$ as the arms in our causal bandit problem, so that the reward magnitude across arms is on the same scale.

%\qzcmt{this part can be shortened; jump to the arm set.} 
Treating atomic interventions as arms, we formulate the problem of finding the variable with the largest causal effect under the framework of a multi-armed bandit problem. The arm set is 
\begin{equation}\label{eq:arm_set}
    \mathcal A = \left\{do(X_i=x_i): x_i= -1\text{ or }1, i=1,\ldots,p-1\right\}.
\end{equation}
%where the relationship among all the arms are modeled by a Gaussian DAG. Note that 
Since $\mathbb E[Y| do(X_i=-1)]=-\mathbb E[Y| do(X_i=1)]$ under the linear SEM~\eqref{eq:GaussianBN}, in most practical applications we only need to choose $x_i$ to be either $-1$ or $1$. Consequently, the size of the arm set can be effectively reduced to $K=|\mathcal A|=p-1$. Note that our algorithm and theoretical results apply to the case with both $do(X_i=1)$ and $do(X_i=-1)$ as separate arms. % \qzcmt{is it correct? if not, change to "can be easily modified to accommodate"}
 
For simplicity, let arm $i$ correspond to the intervention $do(X_i=x_i)$ and identify the action set $\mathcal A$ with $\overline V:=\{1,\ldots,p-1\}$. For any arm $i\in \mathcal A$, the expected reward is defined as $\mu_i=\mathbb E[Y| do(X_i=x_i)]$, which has the same magnitude as the causal effect $\gamma_i$ since $|x_i|=1$ by design. The optimal reward is denoted by $\mu^*=\max_{i\in\mathcal A}\mu_i$ corresponding to the optimal arm $i^*=\operatorname{argmax}_{i\in\mathcal A}\mu_i$. The goal is to find the optimal arm while minimizing the expected loss in this process. Let $A_t\in\mathcal{A}$ denote the arm chosen at round $t$. The expected loss is measured by the cumulative regret up to round $T$,
\begin{equation}\label{eq:cumregret_def}
    \mathcal R_T\coloneqq\mu^*T-\sum_{t=1}^T\mathbb E \left(\mu_{A_t}\right)=\sum_{i\in \mathcal A}\Delta_i\mathbb E \left(n_i(T)\right),
\end{equation}
where $\Delta_i=\mu^*-\mu_i$ and $n_i(t)=\sum_{s=1}^t  1\{A_s=i\}$ denotes the number of times that arm $i$ has been chosen up to round $t$. A sequential decision-making policy that minimizes the cumulative regret has to handle the trade-off between exploration and exploitation.

Many algorithms have been proposed for the standard MAB problem with theoretical guarantees on cumulative regret, among which upper confidence bound (UCB) methods and Thompson sampling are two of the most widely studied. For normally distributed rewards, the UCB decision rule~\cite{auer2002MAB} takes the form
\begin{equation}\label{eq:ucb_alg}
    A_{t+1}\in \underset{i \in \mathcal A}{\operatorname{argmax}}
    \left\{\hat{\mu}_{i}(t)+\hat{\sigma}_{i}(t)\, c \sqrt{\log t}\right\},
\end{equation}
where $\hat{\mu}_i(t)$ is an estimator of the expected reward for arm $i$, $\hat{\sigma}_i^2(t)$ is an estimator of the variance of $\hat{\mu}_i(t)$, and $c>0$ is a tuning constant. When the standard MAB framework is applied to the causal bandit setting, only experimental data are collected at each round, and $\hat{\mu}_i(t)$ and $\hat{\sigma}_i(t)$ correspond, respectively, to the sample mean of $Y$ under the intervention $do(X_i=x_i)$ and its associated standard error, computed from data available up to time $t$.
This approach treats each intervention arm in isolation and does not exploit the causal relationships among the variables. In contrast, the BA-UCB method developed in the next section explicitly incorporates causal structures by leveraging observational data collected prior to or alongside the bandit process, leading to more efficient estimation and improved decision-making.

\section{The BA-UCB algorithm}
\label{sec:alg}

%In many real-life situations, obtaining interventional data can be costly. So we consider the  setting where we also collect observational data at each round and incorporate this side information into decision making. 

\subsection{The main idea}

Let $\beta_{i}(Y\sim X_i+X_S)$ denote the regression coefficient of $X_i$ when regressing $Y$ on $(X_i,X_S)$, i.e., $\mathbb E(Y|X_i,X_S)=\beta_{i}X_i+\beta_{S}^\top X_S$. For Gaussian DAGs, if $X_S\subseteq \mathcal{X}\setminus\{X_i,Y\}$ satisfies the backdoor criterion relative to $(X_i, Y)$~\citep{pearl2009causality}, the causal effect $\gamma_i$ of $X_i$ on $Y$ can be identified via backdoor adjustment,
\begin{equation}\label{eq:beta_obs}
   \gamma_i=\beta_{i}(Y\sim X_i+X_S).
\end{equation}
Therefore, $\gamma_i$ can be estimated using not only experimental data under intervention on $X_i$ but also observational data as long as a backdoor adjustment set $S$ is provided. %\qzcmt{Add a short overview of the main idea here: many applications there are large amount of prior observational data, generate observational data is usually much less costly than experimental data, combine them would improve estimate accuracy, use both types of data can identify backdoor adjustment sets. Moved the first paragraph of Section~\ref{subsec:backdoor_id} here, which is a part of the main idea.}

In many practical applications, a large amount of observational data is often available, while generating experimental data is typically much more costly. Leveraging both sources of data has at least two major benefits: (i) Observational data can be used with experimental data to identify valid backdoor adjustment sets, enabling causal identification using observational data alone; (ii) the abundance of observational data can reduce the variance of causal effect estimators, thereby improving accuracy. This motivates our work to integrate observational and experimental evidence in causal bandit settings.

In Section~\ref{subsec:backdoor_id}, we propose a method to identify backdoor adjustment sets for each node assuming the DAG is unknown. Once a backdoor adjustment set is identified, we can obtain an estimate of the causal effect $\mu_i$ and its variance $\sigma_i^2$ from either experimental or observational data. The two estimates can then be combined to improve estimation accuracy. Suppose that we have collected prior observational data of size $n_0\geq 0$ before the sequential decision process, and will collect $n$ observational data and $m$ experimental data at each round. Note that $n_0=0$ if prior data is not available. Recall that $n_i(t)=\sum_{s=1}^t 1\{A_s=i\}$ is the number of rounds at which intervention $do(X_i)$ is performed. Denote by $\mathcal D_i(t)$ the experimental data generated under $do(X_i)$ and $\mathcal D_o(t)$ the observational data collected, both up to round $t$. Then, $N_i(t)=mn_i(t)$ is the size of $\mathcal D_i(t)$ and $N_o(t)=n_0+nt$ is the size of $\mathcal D_o(t)$.

\subsection{Identification of backdoor adjustment sets}
\label{subsec:backdoor_id}

It is straightforward to obtain a consistent estimator of $\mu_i$ and its variance from $\mathcal D_i(t)$ as follows:
\begin{equation}\label{eq:est_int}
    \hat\mu_{i,\mathrm{int}}(t)=\frac{1}{n_i(t)}\sum_{s=1}^t\bar Y_s 1\{A_s=i\}, \quad \widehat{\operatorname{Var}}(\hat{\mu}_{i,\mathrm{int}}(t))=\frac{s_{i,t}^2}{N_i(t)},
\end{equation}
where $\bar Y_t$ is the mean of rewards collected under intervention at round $t$ and $s_{i,t}^2$ is the sample variance of $Y$ generated under $do(X_i=x_i)$ up to round $t$. From the property of Gaussian DAGs, the intervention distribution is $Y|do(X_i=x_i)\sim \mathcal N(\mu_i,\widetilde\sigma_i^2)$ for some $\widetilde\sigma_i^2>0$, and therefore $\hat\mu_{i,\text{int}}(t)\sim \mathcal N(\mu_i, \widetilde\sigma_i^2/N_i(t))$ is an unbiased estimator of $\mu_i$.

Estimates of causal effects $\mu_i$ can also be obtained from observational data by backdoor adjustment \eqref{eq:beta_obs}. For any subset $S$ of nodes, let $\mathbf{{X}}_{i,S}=[\mathbf X_i\mid  \mathbf X_S]$ denote a matrix of dimension $N_o(t)\times (s+1)$, where $s = |S|$, consisting of the columns corresponding to $X_i$ and $X_S$ in the observational dataset $\mathcal D_o(t)$. Denote by $\hat\beta=[\hat\beta_{i},\ \hat\beta_S]$ the least-squares coefficient vector for the linear regression $Y\sim X_i+X_S$ using $\mathcal D_o(t)$. Construct the following estimators of the reward $\mu_i=\mathbb{E}(Y\mid do(X_i=x_i))$ and its variance,
\begin{equation}\label{eq:est_obs}
 \hat\mu_{i,\mathrm{obs}}(t,S) =\hat \beta_{i}\cdot x_i,\quad \widehat{\operatorname{Var}}(\hat{\mu}_{i,\mathrm{obs}}(t,S))=\frac{\rho_i(t,S)\parallel \mathbf Y- \mathbf X_i\hat\beta_{ i}-\mathbf X_S\hat\beta_{S}\parallel^2}{N_o(t)(N_o(t)-s-1)},
\end{equation}
where $\rho_i(t,S)$ is the $(1,1)$-th element of the matrix $(\mathbf{X}_{i,S}^\top \mathbf{X}_{i,S}/N_o(t))^{-1}$. 
%Since $Y|X_i,X_S\sim \mathcal N(X_i\beta_{i}+\beta_S^\top X_S, \zeta_{i,S}^2)$, where $\zeta_{i,S}^2=\mathrm{Var}(Y\mid X_i,X_S))$, 
It is easy to see that 
\[
\hat\mu_{i,\mathrm{obs}}(t,S)\mid \mathbf{{X}}_{i,S}\sim \mathcal N(\beta_{i}x_i, \rho_i(t,S)\zeta_{i,S}^2/N_o(t)),
\]
where $\beta_i=\beta_{i}(Y\sim X_i+X_S)$ and $\zeta_{i,S}^2=\operatorname{Var}(Y\mid X_i,X_S)$.
To simplify notation, let
\begin{equation}\label{eq:mu_S_def}
    \mu_i(S)=\beta_{i}x_i,
\end{equation}
where $x_i\in\{-1,1\}$ is the chosen intervention level in $do(X_i=x_i)$. %the data-generating model so that $\mu_i(S)\geq 0$, i.e., $x_i=\mathrm{sign}(\beta_i)$.} 
If $X_S$ satisfies the backdoor criterion relative to $(X_i,Y)$, then $\mu_i(S)=\mu_i$ and $\hat\mu_{i,\mathrm{obs}}(t,S)$ is an unbiased estimator of $\mu_i$ with variance approximated by \eqref{eq:est_obs}. It is a natural idea to combine the two estimators in \eqref{eq:est_int} and \eqref{eq:est_obs} to improve the statistical efficiency in the upper confidence bounds for $\mu_i$ in a bandit algorithm.

Next, we discuss how to make a data-driven decision on whether a subset $S$ is a valid adjustment set. Let $\mathcal B_i:=\{S\subseteq \overline V\setminus\{i\}:\mu_i(S)=\mu_i\}$ be the collection of sets that yield the true causal effect of $X_i$ on $Y$, which we call valid backdoor adjustment sets.
Since experimental data and observational data are independent, %the distribution of  
\[\hat\mu_{i,\mathrm{obs}}(t,S)- \hat\mu_{i,\mathrm{int}}(t)\mid\mathbf{{X}}_{i,S}\sim \mathcal N\left(\mu_i(S)-\mu_i, \frac{\rho_i(t,S)\zeta_{i,S}^2}{N_o(t)}+\frac{\widetilde\sigma_i^2}{N_i(t)} \right).\]
%\qzcmt{$\widetilde\sigma_i^2$ not defined yet?}
As $t$ increases, this Gaussian distribution will concentrate around $\mu_i(S)-\mu_i$, which equals $0$ if $S$ is a valid backdoor adjustment set relative to $X_i$ and $Y$. We assume that for any set $S$, either $|\mu_i(S)-\mu_i|\geq 2\delta$ for some $\delta>0$ or $\mu_i(S)=\mu_i$. Under this assumption, if $S\notin\mathcal B_i$, the difference $\hat\mu_{i,\mathrm{obs}}(t,S)-\hat\mu_{i,\mathrm{int}}(t)$ will deviate from $0$ significantly as $t$ becomes large. This allows us to distinguish valid sets $S\in\mathcal B_i$ from invalid ones $S'\notin\mathcal B_i$ by checking whether the difference between the two estimates is close to zero.

One might choose $S^*=\operatorname{argmin}_{S}|\hat\mu_{i,\mathrm{obs}}(t,S)-\hat\mu_{i,\mathrm{int}}(t)|$ %\qzcmt{notation not defined or inconsistent with prior paragraph} 
at each time step $t$ as a candidate adjustment set relative to $X_i$ and $Y$. This idea turns out to be too greedy and leads to suboptimal choices for a bandit problem. To encourage more exploration, we instead construct confidence intervals $CI_i(t,S)$ at round $t$ for $\mu_i(S)-\mu_i$ as
\begin{equation}\label{eq:CI}
   \hat\mu_{i,\mathrm{obs}}(t,S) - \hat\mu_{i,\mathrm{int}}(t) \pm c_2\sqrt{\log t}\sqrt{ \widehat{\operatorname{Var}}(\hat{\mu}_{i,\mathrm{obs}}(t,S))+ \widehat{\operatorname{Var}}(\hat{\mu}_{i,\mathrm{int}}(t))},
\end{equation}
where $c_2>0$ is some constant. We choose sets $S$ whose corresponding confidence intervals cover $0$ as our candidate backdoor adjustment sets. If such sets do not exist, we choose the set whose confidence interval has the smallest distance to $0$, where the distance for an interval $[a,b]$ is defined as $d(0,[a,b])=\min\{|x|:a\leq x\leq b\}$. Formally, we identify a collection of candidate backdoor adjustment sets relative to $(X_i, Y)$ at time step $t$ as
\begin{equation}\label{eq:backdoor_id}
    \widehat{\mathcal {B}}_i(t) = \operatorname{argmin}_{S}\ d(0, CI_i(t,S)).
\end{equation}
Note that $\widehat{\mathcal {B}}_i(t)$ may consist of multiple sets whose confidence intervals cover 0.
Without any restriction, the number of candidate sets $S$ can be large especially when the number of nodes $p$ is large. %, among which a lot of sets don't satisfy the backdoor criterion. 
Let $M_i$ be the regression neighborhood of node $i$, i.e., the support of the regression of $X_i$ on $X_{-i}$, %$M_i=\{j\in V\setminus i: X_i\notindep X_j|X_{V\setminus\{i,j\}}\}$, 
and let $\mathcal N_i$ denote the collection of all subsets of $M_i$. For a Gaussian DAG formulated by \eqref{eq:GaussianBN}, the regression neighborhood $M_i$ contains the parents of node $i$. Since the parent set $\mathrm{pa}(i)\subseteq M_i$ always satisfies the backdoor criterion, we restrict $S\in \mathcal N_i$ in the  minimization~\eqref{eq:backdoor_id} to reduce the number of candidate sets.

\subsection{Algorithm description}
\label{subsec:alg}

In this section, we describe the BA-UCB algorithm for causal bandit. Unlike most existing methods, our algorithm does not require prior knowledge or exploration of the underlying DAG. The main idea is to use estimates from both experimental data and observational data to identify backdoor adjustment sets alongside the decision process and then construct upper confidence bounds from both sources. 

For a basic MAB problem, upper confidence bounds are of the form $\hat\mu_i(t)+\hat\sigma_i(t)c\sqrt{\log t}$, where $\hat\mu_i(t)$ and $\hat\sigma_i(t)$ are the mean and standard deviation of the sample mean of $Y$ respectively. Since $\mu_i$ and $\sigma_i^2$ can be estimated from both experimental data as in \eqref{eq:est_int} and observational data as in \eqref{eq:est_obs}, we consider using weighted average of the two estimates to construct UCBs, i.e.,
\begin{equation}\label{eq:UCB}
\begin{aligned}
     \hat{\mu}_{i}(t,S)&=\frac{N_{o}(t) \hat{\mu}_{i,\mathrm{obs}}(t,S)+N_{i}(t) \hat{\mu}_{i, \mathrm{int}}(t)}{N_{o}(t)+N_{i}(t)},\\
\hat\sigma_i^2(t,S)&=\frac{N_o^2(t)\widehat{\operatorname{Var}}(\hat{\mu}_{i,\mathrm{obs}}(t,S))+N_i^2(t)\widehat{\operatorname{Var}}(\hat{\mu}_{i,\mathrm{int}}(t))}{(N_o(t)+N_i(t))^2}.
\end{aligned}
\end{equation}
This effectively combines the two types of data to achieve higher estimation accuracy.

As discussed in Section~\ref{subsec:backdoor_id}, we restrict $S\in \mathcal N_i$ to find candidate backdoor adjustment sets relative to $X_i$ and $Y$. We use observational data $\mathcal D_o(t)$ to estimate the regression neighborhood $M_i$ of each node $i\in \overline V$. %\qzcmt{for almost all values of $B$ and $\Omega$ (are these notatons defined?)} %Lebesgue measure $1$. % We apply a nodewise neighborhood regression method to estimate the Markov blanket \citep{Meinshausen_2006}. 
% \qzcmt{reference too general; better to cite Meinshausen and Buhlmann 2006 AOS} %The conditional distributions are $X_i|\mathbf X_{-i}\sim \mathcal N(\bar\mu_i,\bar\sigma_i^2)$, where $\bar\mu_i=\sum_{j:j\neq i}\bar\beta_{ij}X_j$ and $\bar\beta_{ij}= 0$ iff $X_i\indep X_j|\mathbf X_{-ij}$. So we can estimate the Markov blanket of $X_i$ by running a linear regression of $X_i$ on all other variables and choosing those with significant coefficients.
Given the estimated regression neighborhood $\widehat{M}_i$, we identify candidate backdoor adjustment sets $S\subseteq\widehat{M}_i$ as in 
\eqref{eq:backdoor_id}, sample one set $S$ from the candidate sets if $|\widehat{\mathcal B}_i(t)|\geq 2$, and construct the weighted version of UCBs to choose the intervention for the next round. The procedure is summarized in Algorithm~\ref{alg:BA-UCB}. In practice, estimated regression neighborhoods and candidate backdoor adjustment sets will not vary significantly unless we collect enough new data. Therefore, to reduce computation, we update them every $\kappa$ iterations.

%\qzcmt{In Algorithm 1, where do you calculate (7), in Line 11 or after Line 15? Please clarify}

\begin{algorithm}
   \caption{BA-UCB for Gaussian DAGs}
   \label{alg:BA-UCB}
\begin{algorithmic}[1]
\Require Observational data $\mathcal{D}_o(0)$ of size $n_0$, parameters $c,c_2, c_3,n, m, \kappa$, rounds $T$.
\For{$t=0,\cdots, T-1$}
\If{there is a node $i$ which has been intervened less than $\lceil\frac{c_3\log t+1}{m} \rceil$ times}\label{alg:line:min_time}
    \State Choose intervention node $A_{t+1} = i$;
\Else
\For {$i=1,\cdots, p-1$}
 \If{$t \equiv 0 \pmod{\kappa}$}
  \State Estimate the regression neighborhood $\widehat{M}_i$ from $\mathcal D_o(t)$; \label{alg:line:MB}%of each node $i$; %and find the collection of sets $\widehat{\mathcal N}_i$;
  \State Identify candidate adjustment sets $\widehat{\mathcal B}_i = {\operatorname{argmin }}_{S \subseteq \widehat{M}_i} d(0, CI_i(t,S))$; \label{alg:line:backid}
  \EndIf
  \State Sample set $S$ from $\widehat{\mathcal B}_i$, calculate $\hat{\mu}_{i,\mathrm{obs}}(t,S)$ and $\hat\sigma_{i,\mathrm{obs}}(t,S)$ as in \eqref{eq:est_obs};\label{alg:line:sample}
   \State Calculate estimates $\hat\mu_i(t,S)$, $\hat\sigma_i^2(t,S)$ as in \eqref{eq:UCB};\label{alg:line:estimate}
\EndFor
    %\State Choose $A_{t+1}=\underset{i \in \{1, \ldots, p-1\}}{\operatorname{argmax}}\{\hat{\mu}_{i}(t,S)+$
    %\State \quad \quad \quad $\hat{\sigma}_{i}(t,S) c \sqrt{\log t}\}$; 
    \State $A_{t+1}=\operatorname{argmax}_i \{\hat{\mu}_{i}(t,S)+\hat{\sigma}_{i}(t,S) c \sqrt{\log t}\}$;
\EndIf 
\State Generate $m$ data points under intervention $A_{t+1}$ and $n$ observational data. \label{alg:line:collect}
\State Update $\hat\mu_{i,\mathrm{int}}(t+1)$ and $\widehat{\operatorname{Var}}(\hat\mu_{i,\mathrm{int}}(t+1))$ as in \eqref{eq:est_int}.
\EndFor
\Ensure Final intervention $A_T$ and estimated reward $\hat \mu^*$.
\end{algorithmic}
\end{algorithm}

\begin{remark}\label{remark:one-regression}
   According to Rule 2 of $do$ Calculus by Pearl~\cite{pearl2009causality}, if $X_S$ blocks all backdoor paths from $X$ to $Y$, we have $P(Y|do(X=x),X_S)= P(Y|X=x, X_S)$, i.e., conditioning on the backdoor adjustment set $S$, the intervention distribution $[Y| do(X=x),X_S]$ is identical to the conditional distribution $[Y| X=x,X_S]$. Hence, given a backdoor adjustment set $S$ for node $i$, we can perform a single regression $Y\sim X_i + X_S$ on both experimental data under $do(X_i)$ and observational data to estimate $\mu_i$. Accordingly, in Algorithm~\ref{alg:BA-UCB} Line~\ref{alg:line:sample}, we would run one linear regression $Y\sim X_i+X_S$ using both experimental data $\mathcal{D}_i(t)$ and observational data $\mathcal{D}_o(t)$. 
   In Line~\ref{alg:line:estimate}, we calculate $\hat{\mu}_{i}(t,S)$ and $\hat\sigma_i^2(t,S)$ as in \eqref{eq:est_obs} but with the combined data. This is another way to combine the two types of data to generate estimates and construct UCBs. However, this single-regression approach exhibits inferior empirical performance, as shown by the simulation results in Section~\ref{subsec:sim_results}. The underlying issue is that when $S$ is not a valid adjustment set, the estimator of $\mu_i$ from the regression $Y \sim X_i + X_S$ can be severely biased, particularly when applied to experimental data. Consequently, the single-regression estimator based on the pooled dataset $\mathcal{D}_i(t) \cup \mathcal{D}_o(t)$ could be very different from both the observational estimator $\hat{\mu}_{i,\mathrm{obs}}(t,S)$ and the interventional estimator $\hat{\mu}_{i,\mathrm{int}}(t)$, even though these two estimators are themselves close. In contrast, the weighted-average estimator in \eqref{eq:UCB} combines the two estimators in a controlled manner, yielding a more robust $\hat{\mu}_i(t)$ and superior performance in practice.
\end{remark}

\section{Regret analysis}
\label{sec:regret}

In this section, we provide theoretical analysis on the cumulative regret $\mathcal{R}_T$~\eqref{eq:cumregret_def} of our proposed BA-UCB algorithm. Since using the same data to identify backdoor adjustment sets in Line~\ref{alg:line:backid} and to construct combined estimates in Line~\ref{alg:line:estimate} of Algorithm~\ref{alg:BA-UCB} complicates the analysis, we consider sample splitting to simplify the proof. More specifically, we collect $2n_0$ observational data at the beginning of Algorithm~\ref{alg:BA-UCB}, $2m$ experimental data and $2n$ observational data at each round in Line~\ref{alg:line:collect}, and use half of the data to identify candidate backdoor adjustment sets in Line~\ref{alg:line:backid} and the other half to calculate estimates in Line~\ref{alg:line:estimate}. All theoretical results in this section are established under this sample-splitting version of Algorithm~\ref{alg:BA-UCB}. %The only difference from our setting in \ref{sec:alg} and \ref{sec:experiments} is that we double the sample sizes and use separate data for the two procedures. 
In practice, the algorithm works well without sample splitting, as demonstrated by the numerical experiments in Section~\ref{sec:experiments}.

\subsection{Main results}\label{subsec:regret_analysis}

The underlying model assumptions for Algorithm~\ref{alg:BA-UCB} are stated formally in Assumption~\ref{ass:gaussian_dag}.
\begin{assumption}\label{ass:gaussian_dag}
The random vector $X=(X_1,\cdots,X_p)$ satisfies the linear SEM~\eqref{eq:GaussianBN} associated with a DAG $\mathcal G$.
\end{assumption}

As explained in Section~\ref{subsec:backdoor_id}, we also assume a gap between $\mu_i(S)$ for $S\notin \mathcal B_i$ and the true causal effect $\mu_i$, which is formally stated as Assumption~\ref{ass:id}. 

\begin{assumption}[Identifiability]\label{ass:id}
    There exists $\delta>0$, such that either $|\mu_i(S)-\mu_i|\geq 2\delta$ or $\mu_i(S)=\mu_i$, $\forall S\in\mathcal N_i$ and $i\in\{1,\cdots, p-1\}$.
\end{assumption}

\begin{remark}\label{remark:gap_condition}
Since we do not need to recover the entire graph, the faithfulness
assumption is not required in our work. Note that backdoor adjustment relies only on the recursive factorization of the joint distribution according to a DAG, which is
implied by the SEM in \eqref{eq:GaussianBN}. %\ref{ass:id} could be implied by faithfulness assumption and thus is a weaker assumption.
\end{remark}

\begin{figure}
\includegraphics[width=0.4\textwidth]{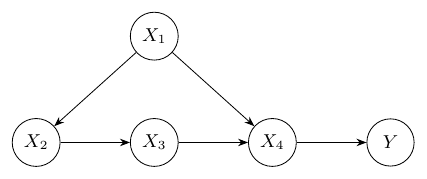}
    \caption{An example DAG to illustrate Assumption~\ref{ass:id}.}
    \label{fig:dag}
\end{figure}

 %By Assumption~\ref{ass:id}, there is a gap between the true causal effect $\mu_i$ identified with a valid backdoor adjustment set and $\mu_i(S)$ using an incorrect set $S$. 
 If $S$ satisfies the backdoor criterion, then $\mu_i(S)$ identifies the causal effect of $X_i$ on $Y$; in particular, the parent set $\mathrm{pa}(i)$ of $X_i$ always satisfies the backdoor criterion, and thus $\mu_i(\mathrm{pa}(i))$ is the true causal effect. Under Assumption~\ref{ass:id}, for any $S'$ that is not a valid backdoor adjustment set, there is a gap such that $|\mu_i(S')-\mu_i|\geq 2\delta$. Using Figure~\ref{fig:dag} as an example, suppose we want to estimate the causal effect of $X_2$ on $Y(=X_5)$, i.e., $\gamma_{2}=\beta_{23}\beta_{34}\beta_{45}$. In this example, the regression neighborhood $M_2$ of $X_2$ consists of its parent $X_1$ and child $X_3$, and thus $\mathcal N_2=\{\emptyset, \{X_1\}, \{X_3\},\{X_1,X_3\}\}$. Among these subsets, only $\{X_1\}$ satisfies the backdoor criterion relative to $(X_2, Y)$, i.e., $\mathcal{B}_2=\{\{X_1\}\}$. From the SEM~\eqref{eq:GaussianBN}, the conditional expectations of $Y$ given $X_2$ and different sets in $\mathcal N_2$ can be derived as follows:
 \begin{align*}
     &\mathbb{E}\left[Y \mid  X_2\right]=\beta_{45}\left( \frac{\beta_{12}\beta_{14} \omega_1^2}{\beta_{12}^2 \omega_1^2+\omega_2^2}+\beta_{34} \beta_{23}\right) X_2,\\
     &\mathbb{E}\left[Y \mid X_1, X_2\right]=\left(\beta_{45} \beta_{14}\right) X_1+\left(\beta_{45} \beta_{34} \beta_{23}\right) X_2,\\
     &\mathbb{E}\left[Y \mid X_2, X_3\right]=\beta_{45}\left(\frac{\beta_{12}\beta_{14} \omega_1^2}{\beta_{12}^2 \omega_1^2+\omega_2^2} X_2+\beta_{34} X_3\right),\\
    & \mathbb{E}\left[Y \mid X_1, X_2, X_3\right]=\beta_{45}\left(\beta_{14} X_1+\beta_{34} X_3\right).
 \end{align*}
 Then the minimum difference between the true causal effect $\gamma_{2}=\beta_{23}\beta_{34}\beta_{45}$ and $\mu_2(S)$, $S\in\mathcal N_{2}\setminus\mathcal B_2$ is 
 $$\min\left\{\left| \frac{\beta_{12}\beta_{14}\beta_{45} \omega_1^2}{\beta_{12}^2 \omega_1^2+\omega_2^2}\right|,\left|\beta_{45}\left( \frac{\beta_{12}\beta_{14} \omega_1^2}{\beta_{12}^2 \omega_1^2+\omega_2^2}-\beta_{34} \beta_{23}\right)\right|,\left|\beta_{23}\beta_{34}\beta_{45} \right|\right\}.$$ 
 Since the parameters $\beta_{ij}$ are non-zero for directed edges, this difference becomes $0$ only when $\beta_{12}(\beta_{14}-\beta_{23} \beta_{34}\beta_{12})\omega_1^2=\beta_{23} \beta_{34}\omega_2^2$, a set of measure zero in the parameter space. This shows the rationale behind Assumption~\ref{ass:id}.

    % \qzcmt{We can be more concrete, writing down the exact expression, in terms of $\beta_{ij}$ and $\sigma_j^2$, of $\beta_2(Y\sim X_2+S)$ for each $S\in \mathcal{N}_2$ in this example.}

To make sure our algorithm works properly, we also need a mild assumption on the number of observational data for estimation of regression neighborhoods and $\hat{\beta}_i$ in~\eqref{eq:est_obs} from backdoor adjustment sets. This is stated in Assumption~\ref{ass:n0}.

\begin{assumption}\label{ass:n0}
$n_0+nK\geq p$.
\end{assumption}
Recall that $K=p-1$ in our setting. This assumption is easily satisfied when $n_0$, $n\geq 1$ or $n_0\geq p$. Let $s_0$ denote the maximum of in- and out-degrees among all nodes except the reward variable. In fact, if we use Lasso-based neighborhood regression and restrict the search of candidate backdoor adjustment sets to subsets with cardinality  $\leq s_0$ for computational efficiency, then Assumption~\ref{ass:n0} can be relaxed to $n_0+nK\geq s_0+2$. 

Recall a few relevant distributions introduced earlier in the paper.
The joint distribution of $X$ defined by the linear SEM \eqref{eq:GaussianBN} is $\mathcal N( 0,\Sigma)$, where $\Sigma=( I- B)^{-\top}\Omega( I-  B)^{-1}$, $\Omega=\mathrm{diag}(\omega_1^2,\cdots,\omega_p^2)$ and $ B=(\beta_{ij})_{p\times p}$.
The distribution of the reward variable $Y$ under intervention is   
\[
Y\mid do(X_i=x_i)\sim \mathcal N(\mu_i,\widetilde\sigma_i^2),
\]
and the conditional distribution $[Y\mid X_i, X_S]$ is 
\begin{align*}
Y\mid X_i,X_S \sim \mathcal N\left(\mu_i(S)X_i+\beta_S^\top X_S, \zeta_{i,S}^2\right).
\end{align*}
Let $\gamma_{\min}$,$\gamma_{\max}$, and $\operatorname{tr}$ denote the smallest eigenvalue, the largest eigenvalue, and the trace of a matrix, respectively. Define a few constants:
\begin{equation}\label{eq:def_notation}
    \begin{aligned}
\widetilde{\eta}_i^2&=\underset{S\in\mathcal B_i}{\max}\left\{ \frac{9\zeta_{i,S}^2}{\gamma_{\min}(\Sigma)},\tilde\sigma_i^2\right\},\quad\quad
\widetilde\eta^2=\max_i\widetilde{\eta}_i^2,\\
\widetilde\phi^2 &= \underset{i}{\max}\,\underset{S\in\mathcal B_i}{\max}\max\left\{\frac{9\zeta_{i,S}^2}{\gamma_{\min}(\Sigma)\tilde\sigma_i^2},\frac{25\tilde\sigma_i^2\gamma_{\max}(\Sigma)}{9\zeta_{i,S}^2}\right\},\\
 \psi^2&=\max_i\max_{S'\in \mathcal N_i\setminus\mathcal B_i}\max\left\{\frac{9\zeta_{i,S'}^2}{\gamma_{\min}(\Sigma)},\widetilde\sigma_i^2\right\}.
    \end{aligned}
\end{equation}

Now we establish a case-dependent upper bound on the cumulative regret of the BA-UCB algorithm. Here, ``case-dependent'' refers to a bound that explicitly depends on the reward gaps $\Delta_i=\mu^*-\mu_i$. 
\begin{theorem}\label{thm1}
Under Assumptions~\ref{ass:gaussian_dag}, \ref{ass:id} and~\ref{ass:n0}, the cumulative regret $\mathcal{R}_T$ of Algorithm~\ref{alg:BA-UCB}, with parameters $c\geq 4\sqrt{2}\widetilde\phi$, $c_2=2\sqrt{3}$, and $c_3\geq \max\left\{64,32\psi^2/\delta^2\right\}$, after $T$ rounds is at most
\begin{equation}\label{eq:thm1}
\begin{aligned}
   \log T\left(\frac{20c^2}{n+m}\sum_{i:\mu_i<\mu^*}\frac{\widetilde\eta_i^2}{\Delta_i}+\frac{c_3\sum_{i=1}^K\Delta_i}{m}\right)+C_3 \sum_{i=1}^K\Delta_i,
\end{aligned}
\end{equation}
where $C_3>0$ is a constant that does not depend on $T$.
\end{theorem}
In general, $C_3$ depends on $\Sigma$, $n_0$ and $n$, and its exact expression is provided in Section~\ref*{appendix:constant} of the supplementary materials. See Section~\ref*{appendix:DAGunknown} in the supplementary materials for a proof of this theorem. %The right side of \eqref{eq:thm1} is a case-dependent bound that depends on $\Delta_i=\mu^*-\mu_i$.

Theorem~\ref{thm1} shows the advantage of incorporating observational data into the UCB framework. The standard UCB algorithm has an upper bound of a similar form, such as in Theorem~4 of Auer et al.~\cite{auer2002MAB}, which is
\[
\frac{\log T}{M}\left(256\sum_{i:\mu_i<\mu^*} \frac{\widetilde\sigma_i^2}{\Delta_i}+8\sum_{i=1}^K\Delta_i\right)+\text{const}\cdot\sum_{i=1}^K\Delta_i
\]
assuming $M$ data points are generated under each round of intervention.
%The key difference is that the leading term in our bound \eqref{eq:thm1} \qzcmt{eq ref incorrect here and also other places in this section.} is $\frac{20c^2}{n+m}\sum_{i:\mu_i<\mu^*}\frac{\widetilde\eta_i^2}{\Delta_i}\log T$ when the reward gaps $\Delta_i$ are small. 
Setting $n+m=M$ in the bound~\eqref{eq:thm1} will result in the same order of the leading term $O(\log T/M)$, showing that observational data, while much less costly, is as useful as experimental data in terms of reducing the regret for the BA-UCB algorithm. Increasing $n$, the size of the observational data in each round, directly reduces this dominating term and thus lowers the bound. Moreover, the benefit of having an initial observational dataset  is reflected in the constant $C_3$, which also lowers the cumulative regret.

In the MAB literature, regret analyses typically fall into two categories: case-dependent bounds that have an explicit dependency on $\Delta_i$, and case-independent bounds that hold uniformly over all problem instances. Since our algorithm is developed under Assumption~\ref{ass:gaussian_dag}, dependency on the underlying distribution of $X$ is unavoidable. Nevertheless, in line with the standard practice, we also derive an upper bound that does not depend explicitly on the gaps $\Delta_i$.

To do that, we assume that all rewards are bounded, i.e., $\mu_i\in[-1,1]$ without loss of generality, and accordingly, that all constructed upper confidence bounds are clipped to $[-1,1]$. While this bound does not involve $\Delta_i$, it still depends on other problem-specific parameters defined by the underlying distribution (e.g., through $\Sigma$).

\begin{theorem}\label{thm2}
    Assume $\mu_i\in[-1,1],\,\forall i\in\{1,\cdots,p-1\}$. Then under Assumptions~\ref{ass:gaussian_dag}, \ref{ass:id} and~\ref{ass:n0}, the cumulative regret $\mathcal{R}_T$ of Algorithm~\ref{alg:BA-UCB}, with parameters $c\geq 4\widetilde\phi$, $c_2=2\sqrt{3}$, and $c_3\geq \max\left\{32, 32\psi^2/\delta^2\right\}$, after $T$ rounds is at most
\begin{equation}\label{eq:thm2}
\begin{aligned}
    2c\widetilde\eta \sqrt{\log T}&\min\Bigg\{\frac{\sqrt{(n+m)KT+K^2n_0}-\sqrt{K^2n_0}}{n+m},\ \frac{\sqrt{n_0+nT}-\sqrt{n_0}}{n} 1\{n\geq 1\}\Bigg\}\\
   &+\frac{2Kc_3\log T}{m}
        +C_4,
\end{aligned}
\end{equation}
where $C_4$ is a constant that does not depend on $T$.
\end{theorem}
In general, $C_4$ depends on $\Sigma$, $n_0$, $m$, and $n$, and the exact expression is provided in Section~\ref*{appendix:constant} of the supplementary materials. See Section~\ref*{appendix:DAGunknown} in the supplementary materials for a proof of this theorem.

\subsection{Implications}\label{subsec:regret_implication}

To better understand how observational data benefit the sequential decision process, we take a closer look at \eqref{eq:thm2}. Since $T$ is usually much larger than $K$, the first term in \eqref{eq:thm2} is the leading term. If $n_0$ is large relative to $T$, more specifically $n_0\geq T(n+m)$, we have  
\begin{equation*}
\begin{aligned}
     \frac{\sqrt{(n+m)KT+K^2n_0}-\sqrt{K^2n_0}}{n+m} 
&=\frac{\sqrt{K^2n_0}}{n+m}\left(\sqrt{1+\frac{(n+m)T}{Kn_0}}-1\right)\\
&\leq \frac{T}{2\sqrt{n_0}}\leq \frac{1}{2}\sqrt{\frac T {n+m}},
\end{aligned}
\end{equation*}
where the first inequality is due to $(1+x)^{1/2}\leq 1+x/2$ for $x>0$. In this case, the leading term of the cumulative regret is $O(\sqrt{T\log T/(n+m)})$, which does not depend on $K$ and decreases with the total number $2(m+n)$ of experimental and observational data collected at each round. The cumulative regret of a standard UCB algorithm that generates $2(n+m)$ experimental data at each round is $O(\sqrt{KT\log T/(n+m)})$. This shows that by using a large prior observational data with backdoor adjustment, we reduce the regret substantially by a factor of $O(\sqrt{K})$ and remove its dependency on the action set size $K$. Moreover, the value of observational data generated at each round is comparable to that of experimental data in terms of minimizing the cumulative regret.

If $n_0$ is small relative to $T$, since $\sqrt{x+a}-\sqrt{x}\leq \sqrt{a}$ when $a>0$, the leading term in~\eqref{eq:thm2} is bounded by the minimum between $O(\sqrt{KT\log T/(n+m)})$ and $O(\sqrt{T\log T/n})$. %When $K/(n+m)\leq 1/n$, i.e., $K<1+m/n$ is not a large number, the leading term of the cumulative regret is $O(\sqrt{KT\log T/(n+m)})$; 
In general, it is cheaper to generate observational data than experimental data and thus $m/n$ is often chosen to be small, say $m/n\leq 1$. Therefore, for many problems, we expect to have $K> 1+m/n$ and in this case the leading term of \eqref{eq:thm2} is $O(\sqrt{T\log T/n})$, which does not depend on $K$ either. If the action set is large, i.e., $K\gg 1+m/n$, then the BA-UCB bound $O(\sqrt{T\log T/n})+O(K\log T/m)$ will be much smaller than the bound of the standard UCB $O(\sqrt{KT\log T/(n+m)})$ discussed above, under the typical setting that $T\gg K^2\log T$. 
%Note that in this case,
%the regret bound of our algorithm is $O(\sqrt{T\log T/n})+O(K\log T/m)$ and thus experimental data is still needed at each round ($m\geq 1$).

In summary, as long as we collect enough observational data with either a large initial $n_0$ or during the bandit process ($n\geq 1$), we can relax the dependence of the upper bound \eqref{eq:thm2} for the cumulative regret on the number of arms $K$, and substantially reduce the order of the cumulative regret bound compared to the traditional MAB algorithms. 
    
The regret analysis demonstrates the advantage of the BA-UCB algorithm using observational data for the causal bandit problem. With backdoor adjustment, observational data can be used to estimate the causal effects of all arms simultaneously, which substantially relaxes the dependence of the cumulative regret on the number of arms. Note that, even with a sufficiently large amount of observational data, one can at most recover the equivalence class of a DAG, represented by a completed partially directed acyclic graph (CPDAG). In the Gaussian case, however, one cannot reduce the action set given a CPDAG to a small subset of the nodes in general. Thus, our approach is fundamentally different from the existing works~\citep{lu2021causal, dekroon2022causal} that use prior data or knowledge to reduce the action set before a sequential bandit algorithm.

\subsection{Proof outline}

Complete proofs of Theorem~\ref{thm1} and \ref{thm2} can be found in the supplementary materials. Here, we give an outline of the main steps of the proofs. 

% \qzcmt{General comment: should include more math expressions in the outline, replacing some of the tedious text descriptions}

We first establish case-dependent and $\Delta$-independent upper bounds on the cumulative regret assuming the underlying causal DAG is given. When the graph structure is given, a valid adjustment set of each node $i$ is its parent set $\mathrm{pa}(i)$, which is used to generate estimates of causal effects from observational data in our BA-UCB algorithm. Then the central component is to show the concentration of the estimated mean $\hat\mu_i(t,\mathrm{pa}(i))$ and variance $\hat\sigma_i^2(t,\mathrm{pa}(i))$ in~\eqref{eq:UCB} around their expectations. By applying certain tail bounds, %summarized in~\eqref{eq:t_bound} and~\eqref{eq:chi_bound}, 
we control the deviation probabilities of these estimators after a sufficiently large number of rounds (i.e., $t$ is large). Given the gaps $\Delta_i$ between the optimal and suboptimal arms, together with the concentration properties of the estimators, we can upper bound the probability $\mathbb P(A_t=i)$ for each sub-optimal arm $i$ for a large enough $t$, leading to a finite upper bound on the cumulative regret.

When the DAG structure is unknown, the regret analysis proceeds as above once valid backdoor adjustment sets have been identified. Conditional on a valid adjustment set, the cumulative regret bound coincides with that in the known-DAG scenario. Let $S(i,t)$ denote the sampled adjustment set for node $i$ at time step $t$ in Algorithm~\ref{alg:BA-UCB} Line~\ref{alg:line:sample}. The main additional challenge lies in bounding the probability that the algorithm fails to identify a valid adjustment set at a given round, i.e., $\mbp(S(i,t)\notin \mathcal B_i)$. % \qzcmt{define the notation $S(i,t)$}. 
% \qzcmt{Revise hereafter according to the proof on page 16 in supplementary materials. Control the probabilities of three events.}
Such an error may arise from inaccurate estimation of the regression neighborhood in Line~\ref{alg:line:MB} or incorrect identification of candidate backdoor adjustment sets in Line~\ref{alg:line:backid}. We can control the error probability by bounding: (i)~$\mbp(\widehat{M}_i \neq M_i)$, (ii)~$\mbp(0\notin CI_i(t, \mathrm{pa}(i)))$, and (iii)~$\mbp(0\in CI_i(t,S')\text{ for some }S'\in\mathcal N_i\setminus \mathcal B_i)$.

%To bound the probability of the first source of error, we analyze the Markov blanket estimation procedure based on node-wise linear regression. 
Under the Gaussian linear SEM~\eqref{eq:GaussianBN}, %the true regression neighborhood $M_i$ of each node corresponds to the set of variables with nonzero regression coefficients in the neighborhood regression $X_i\sim X_{-i}$. For each node $X_i$, 
we show that for each $i$, the estimated neighbor set $\widehat{M}_i$ satisfies %obtained by thresholding the $t$-statistics from the regression, 
%\begin{align*}
    $\mathbb{P}\big(\widehat{M}_i \neq M_i\big)=O(t^{-2}),$
%\end{align*}
provided that the observational sample size $N_o(t)$ grows at least on the order of $\log t$. %logarithmically with $t$ and the threshold is chosen proportional to $\sqrt{\log t}$. 
This result follows from Gaussian concentration inequalities for the $t$-statistics in linear regression and a union bound across all variables. Once $M_i$ is recovered, the probability that $S(i,t)$ in Line~\ref{alg:line:sample} does not belong to the family of valid backdoor adjustment sets $\mathcal{B}_i$ can be upper bounded using the concentration properties of the difference $\hat\mu_{i,\mathrm{obs}}(t,S) - \hat\mu_{i,\mathrm{int}}(t)$ %\qzcmt{use the symbols in the equation} 
in~\eqref{eq:CI}. More specifically, we show that
\begin{align*}
    \mbp\left(0\in CI_i(t, S)\right)&\leq O(t^{-2}), \quad\forall S\in \mathcal N_i\setminus\mathcal B_i,\\
    \mbp\left(0\notin CI_i(t, S)\right)&\leq O(t^{-2}),\quad \forall S\in\mathcal B_i,
\end{align*}
provided that the sample sizes $N_i(t)$ and $N_o(t)$ both grow at least on the order of $\log t$. Combining the error bounds of the three events, we arrive at $\mbp(S(i,t)\notin \mathcal B_i)=O(t^{-2})$, and hence $\sum_{t=1}^\infty\mathbb{P}(S(i,t)\notin\mathcal B_i)$ is finite. Therefore, the increase in the cumulative regret due to these errors is bounded by a constant, and the dominant term in the regret bound remains the same as in the DAG-known case.

\section{Numerical experiments}
\label{sec:experiments}

In this section, we evaluate the empirical performance of the BA-UCB algorithm (without sample splitting) through simulation. In our main experiments, we compare BA-UCB with three competing methods: the standard UCB, the central node (CN)-UCB~\citep{lu2021causal}, and the Bayesian backdoor bandit (BBB)-UCB~\citep{Jireh2022BBB}. 

\subsection{Simulation settings}
\label{subsec:implement}

%In this section, we describe how we generated the Gaussian DAGs and implementation details of the algorithms.
We simulate DAGs with the number of nodes $p\in \{10,20,30,50\}$. We randomly generate a DAG using the function \verb|random.graph|  in the \verb|bnlearn| package, sample coefficients $\beta_{ij}$ from $[-1,-0.5]\cup [0.5,1]$ for $i\in \mathrm{pa}(j)$ and standard deviations $\omega_i$ from $[\sqrt{0.2},\sqrt{0.5}]$, and  normalize each variable to have unit variance. %To reliably generate distributions with meaningful causal effects and reward signals, 
We set maximum in-degree to be 3 and choose the last variable in the topological sort as the reward variable $Y$ to minimize the number of trivial arms. 
In each simulation, we randomly generated 100 Gaussian DAGs, ran each bandit algorithm 5 times on each DAG with $T=5000$ time steps, and recorded the average cumulative regrets. 
 
Since the average execution time of BBB-UCB was $8.2$ seconds for each iteration on a Gaussian DAG with 10 nodes, it is prohibitive to run it for $5000$ time steps on $100$ Gaussian DAGs. So for $p=10$, we directly used the Gaussian DAGs and the corresponding simulation results provided by Huang and Zhou~\cite{Jireh2022BBB}, available at \href{https://github.com/jirehhuang/bcb}{https://github.com/jirehhuang/bcb}. Since the computation time increases exponentially with the number of nodes, we exclude BBB-UCB from our comparison for $p\geq 20$.

The CN-UCB algorithm was proposed for discrete network settings, aiming to first identify the parents of the reward variable and then restrict the action set of UCB to the identified parent nodes. To mimic the mechanism under Gaussian settings, we first apply the PC algorithm~\citep{spirtes1991algorithm} to identify neighbors of the reward variable $Y$ using observational data, where a neighbor of $X_i$ could be  a parent ($X_i\to Y$) or be connected via an undirected edge ($X_i-Y$), as the PC algorithm in general learns a partially directed acyclic graph. Then, we apply the UCB algorithm with these identified neighbors as the arms. The significance level of the PC algorithm for conditional independence test was set to $\alpha=0.1$, which is slightly higher than the common cutoff of $0.05$, so that true neighbors are less likely to be missed. Each node was intervened on at least 5 times at the beginning to allow the UCB algorithm to accumulate enough data. Note that, unlike discrete settings, the optimal intervention with the largest causal effect may not be a parent node of the reward under Gaussian settings. Consequently, the performance of this procedure depends on whether the optimal arm is in $\mathrm{pa}(Y)$. 

For our BA-UCB algorithm, we chose $\tau = 1/\sqrt{N_{o}(t)}$ as the threshold for p-values when estimating the regression neighborhood through linear regression. After several trials, we set $c=c_2=1/\sqrt{2}$ for all settings, where $c$  and $c_2$ are the constants in \eqref{eq:ucb_alg} and \eqref{eq:CI}, respectively. To reduce computational cost, we only considered $\{S\in \widehat{\mathcal N}_i:|S|\leq 3\}$ in the identification of candidate backdoor adjustment sets in Algorithm~\ref{alg:BA-UCB}. We estimated the regression neighborhood $M_i$, calculated $\widehat{\mathcal B}_i$, and updated the estimates from observational data every $\kappa$ steps, where $\kappa=20,50,100, 200$ for $p=10,20,30,50$, respectively, so that the computation time of $5000$ iterations was under 5 minutes for all settings. The numerical results were all generated on a macOS system with M1 chip, 8-core CPU, and 16GB memory. % \qzcmt{notations here inconsistent with those in the alg descripiton, e.g. $c_2$ and $\kappa$}
% How I implement BA-UCB, choice of paramenters.

\subsection{Simulation results of BA-UCB}
\label{subsec:sim_results}

% \qzcmt{Write a short paragraph to give an outline of this subsection.}

In this subsection, we present simulation results to evaluate different aspects of the BA-UCB algorithm. We begin with the setting where the true graph is given, which highlights the benefit of information sharing through observational data. We then compare two estimation approaches, i.e., the single regression versus weighted average (see Remark~\ref{remark:one-regression}), and show that the empirical performance of the weighted average approach is better. Finally, we assess whether observational and experimental estimates are consistent and whether the combined estimate converges to the true reward, and validate the design of our algorithm by showing the accuracy in identifying a true adjustment set.

\subsubsection{Results when the true DAG is given}\label{subsec:dagGiven}

\begin{figure}
\includegraphics[width=0.8\textwidth]{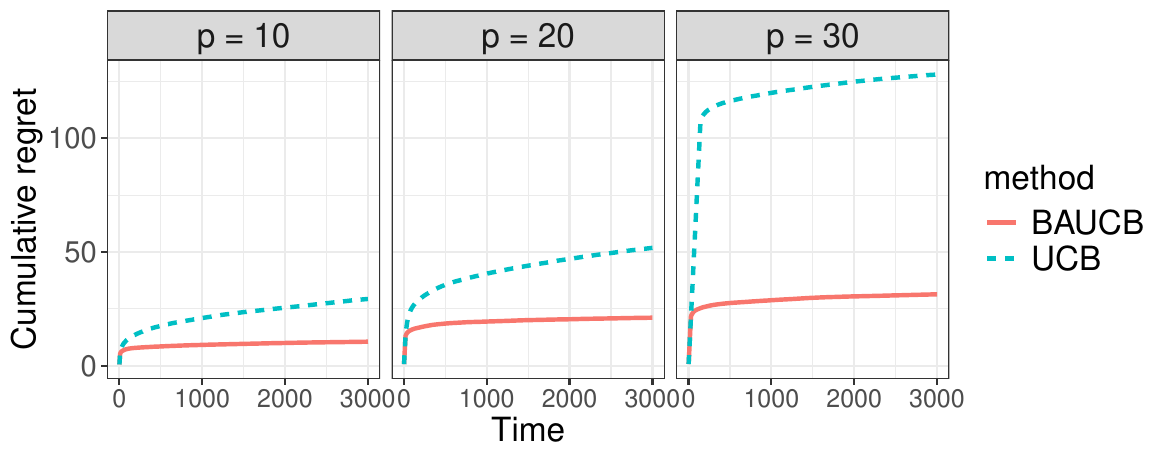}
\caption{Comparison of empirical cumulative regrets over 3000 time steps between the BA-UCB and standard UCB algorithms when DAG is given for $p=10,20,30$. }
    \label{fig:dag_given}
\end{figure}

To highlight the benefit of having observational data, we examined the special case in which the underlying graph is given, for example through domain knowledge. In this setting, we ran our BA-UCB algorithm using the parent set of a node as the backdoor adjustment set and compared it to the standard UCB algorithm. For a fair comparison, BA-UCB collected $n=5$ observational samples and $m=5$ experimental samples per round, starting with no prior data, whereas UCB collected $m+n=10$ experimental samples per round. Each variable was intervened upon at least once in both algorithms to ensure proper initialization. We recorded the median cumulative regret across five runs of each generated graph and reported the average over 100 random DAGs in Figure~\ref{fig:dag_given}.

As shown in Figure~\ref{fig:dag_given}, BA-UCB effectively avoids being trapped in sub-optimal arms and achieves much smaller empirical cumulative regret very quickly, while using the same total number of collected samples as the standard UCB algorithm. Moreover, the performance gap widens as the graph size $p$ increases, highlighting the efficiency of information sharing through observational data. Here, information sharing refers to the fact that, via valid backdoor adjustment, observational data can simultaneously inform the estimation of causal effects across all arms. As the number of arms grows, this information sharing becomes increasingly valuable, resulting in higher efficiency gains for BA-UCB in larger graphs.

\subsubsection{Single regression vs. weighted average estimates}\label{subsec:singleregression}

Hereafter, we assume the underlying DAG is unknown for all numerical results.
 We compare BA-UCB with estimates constructed by the single regression approach described in Remark~\ref{remark:one-regression} and by the weighted average as in \eqref{eq:UCB}. The two approaches only differ in the way to combine observational and experimental data for constructing upper confidence bounds. Given a backdoor adjustment set, the single regression approach runs one linear regression with observational data and experimental data combined together, in contrast to the weighted average of separate estimates from the two sources of data as in Line~\ref{alg:line:estimate} of Algorithm~\ref{alg:BA-UCB}. 

In the simulations, both approaches used $m=n=5$, a minimum intervention time of $1$ for each arm, and 20 initial experimental samples per arm. The number of initial observational samples was set to $n_0=180,380,580$ for $p=10,20,30$, respectively. For each generated graph, we reported the median cumulative regret over five runs and averaged the results across 100 random DAGs.

\begin{figure}
%\begin{center}
\includegraphics[width=0.8\textwidth]{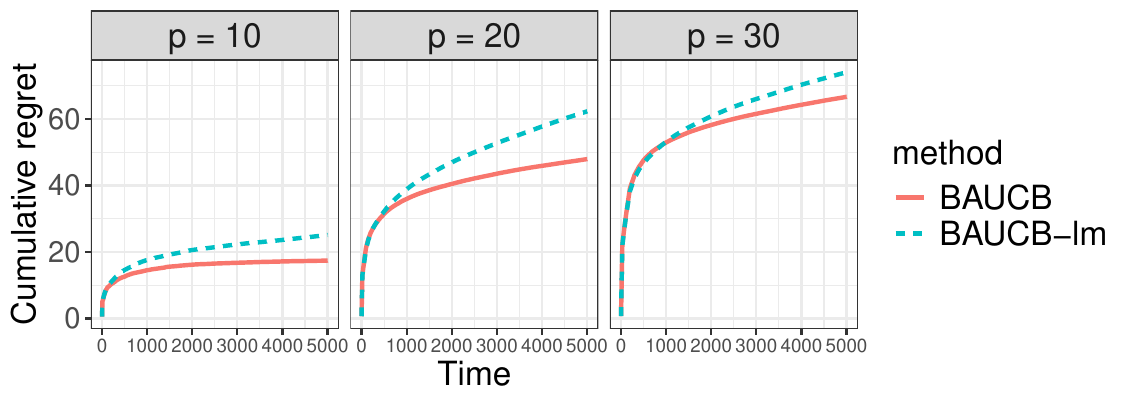}
    \caption{Comparison of empirical cumulative regret over $T=5000$ time steps for BA-UCB on DAGs of $p=10,20,30$. BAUCB-lm uses estimates from the single linear regression and BAUCB uses weighted average. }
    \label{fig:compare_lm}
%    \end{center}
%    \vskip -0.3in
\end{figure}

As shown in Figure~\ref{fig:compare_lm}, the weighted-average approach consistently achieves lower cumulative regret than the single-regression approach across all graph sizes considered. In BA-UCB, adjustment sets are identified based on the criterion that the observational estimate $\hat\mu_{i,\mathrm{obs}}(t)$ obtained using a candidate set as in \eqref{eq:est_obs} is close to the corresponding interventional estimate $\hat\mu_{i,\mathrm{int}}(t)$ as in \eqref{eq:est_int}. However, this does not guarantee that the identified set is a valid backdoor adjustment set, especially at the early stage of the algorithm. When an invalid adjustment set is selected, it is still possible for both $\hat\mu_{i,\mathrm{obs}}(t)$ and $\hat\mu_{i,\mathrm{int}}(t)$ to remain close to each other, and to the true causal effect $\mu_i$, so that their weighted average remains accurate.

In contrast, the single-regression approach relies directly on the identified adjustment set when estimating the causal effect. If the selected adjustment set $S$ is invalid, the regression $Y\sim X_i+X_S$ fitted on the combined data $\mathcal D_o(t)\cup \mathcal D_i(t)$ can yield estimates that deviate substantially from both $\hat\mu_{i,\mathrm{obs}}(t)$ and $\hat\mu_{i,\mathrm{int}}(t)$, leading to inferior empirical performance. By decoupling experimental estimate $\hat\mu_{i,\mathrm{int}}(t)$ from adjustment set selection, the weighted-average approach is therefore more robust to adjustment set mis-specification. Consequently, for the subsequent results, we only focus on the weighted-average approach.

\subsubsection{Consistency of optimal arm estimation}\label{subsec:optimalarm}

To assess whether BA-UCB behaves as designed, we examined the consistency of the estimated reward for the optimal arm by analyzing the results from the simulations in Section~\ref{subsec:singleregression}. Specifically, we evaluated whether the observational estimator $\hat\mu_{\mathrm{obs}}^*(t)$ and the interventional estimator $\hat\mu_{\mathrm{int}}^*(t)$, defined in \eqref{eq:est_obs} and \eqref{eq:est_int}, converge over time, and whether the combined estimator $\hat\mu^*(t)$ in \eqref{eq:UCB} converges to the true optimal reward $\mu^*$ as $t$ increases.

\begin{figure}
\includegraphics[width=0.7\textwidth]{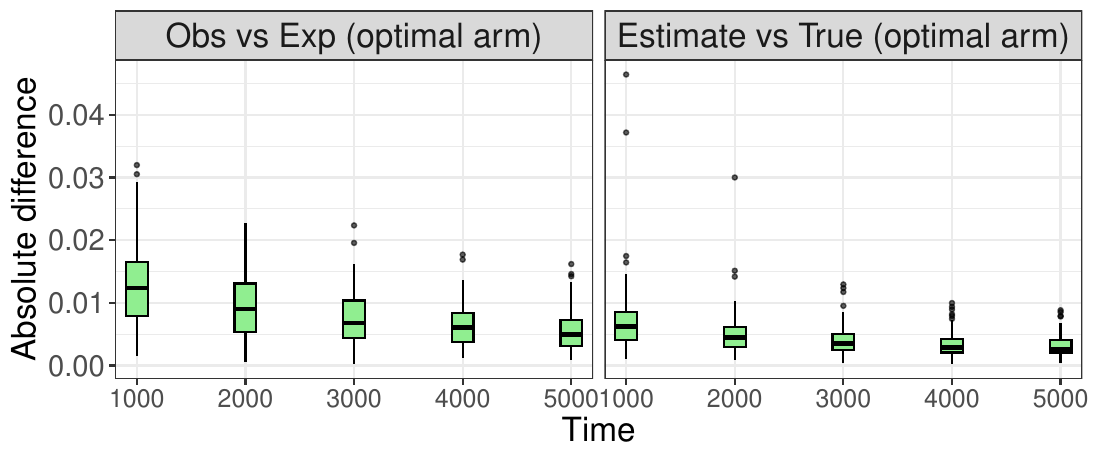}
\caption{Estimation accuracy of the optimal arm over time. The left panel shows the absolute difference between reward estimates based on observational and experimental data, while the right panel reports the absolute difference between the estimated reward and the true value. The box-plots summarize results across 100 randomly generated DAGs with $p=10$. %and the gray line connects the medians across time points. 
% \qzcmt{axis numbers (1000, 0.01, etc) in some figures like this one seem a bit too small, please adjust.}
}
\label{fig:abs_diff}
\end{figure}

Figure~\ref{fig:abs_diff} summarizes estimation accuracy for the optimal arm. In both panels, the median error decreases steadily over time, indicating that observational and experimental estimates become increasingly aligned and that the estimated reward converges toward the true reward value. This behavior is consistent with improved identification of backdoor adjustment sets and leads to diminishing simple regret as the algorithm progresses. Outliers observed at early stages are expected, as adjustment set identification is noisy when limited data are available. 
% Outliers observed at later stages can be attributed to the simulation design, which enforces only a small minimum number of experimental samples. Specifically, each arm is required to have at least 25 experimental samples, consisting of 20 initial experimental samples and at least one intervention round (Line~\ref{alg:line:min_time} in Algorithm~\ref{alg:BA-UCB}), which contributes $m=5$ additional samples. Increasing the minimum intervention time is expected to enable more reliable identification of backdoor adjustment sets and more accurate estimation of the optimal reward.

% \qzedit{[QZ: I noticed that you used begin center inside a figure. This is not necessary, you can remove the centering and vskip commands. their template automatically centering figures and keep proper spacing; I changed Figure 3.]}

\begin{figure}
\includegraphics[width=0.7\textwidth]{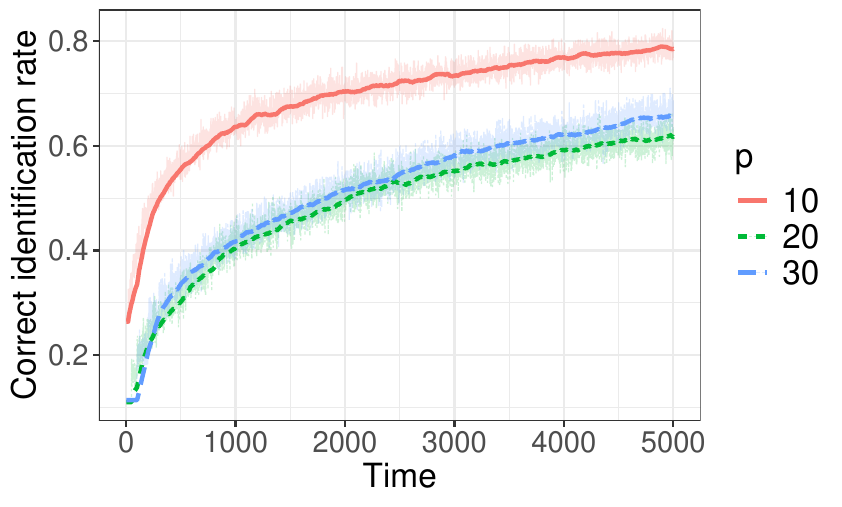}
\caption{Average identification rate of a correct adjustment set for the optimal arm at each time step. Solid curves are smoothing averages with a 100-step window for better visualization.} %for numbers of observed variables $p=9,18,27$, corresponding to total variable counts of $10$, $20$, and $30$, respectively.}}
    \label{fig:parent_id_no_latent}
\end{figure}

To evaluate the accuracy of adjustment set identification, we recorded the frequency of the sampled adjustment set for the optimal arm being a true adjustment set over 5 independent runs and 100 graphs at each time step. Figure~\ref{fig:parent_id_no_latent} reports this average identification rate over time. An identified adjustment set is regarded as correct if $|\mu_i(S)-\mu_i|<0.01$, where $\mu_i(S)$ is defined in \eqref{eq:mu_S_def}. Figure~\ref{fig:parent_id_no_latent} shows a clear increasing trend over time for all graph sizes $p$, demonstrating that BA-UCB progressively improves its identification of valid adjustment sets as more data are collected. Nevertheless, %due to the enforced minimum number of experimental samples in our simulation design and the fact that 
because BA-UCB chooses an adjustment set even when the confidence interval $CI_i(t,S)$ in \eqref{eq:CI} does not cover 0, incorrect adjustment sets may still be selected, especially at early stages.

\subsection{Comparisons}

% \qzcmt{Write a short paragraph to give an outline of this subsection.}

In this subsection, we compare our BA-UCB algorithm with existing causal bandit algorithms, including BBB-UCB, CN-UCB, and the standard UCB. We first present results for small graphs with $p=10$ to illustrate the differences in performance and computational cost across methods. We then move to larger graphs with $p=20,30,50$, excluding BBB-UCB due to its high computational cost. These comparisons demonstrate that BA-UCB attains consistently lower regret than UCB and CN-UCB, while offering substantially greater scalability than BBB-UCB.

%We first compared our BA-UCB algorithm with the other competing methods on small graphs with $p=10$. 
Recall that $n_0$ is the number of initial observational data, $n$ is the number of observational data collected at each round, and $m$ is the number of experimental data collected at each round. In the comparison on small graphs with $p=10$, the results of the BBB-UCB algorithm were generated with $n_0=320$ and $m=1$. We chose $n_0=320$ and $n=m=1$ for our BA-UCB algorithm, under which our algorithm took much less time than the BBB-UCB algorithm for $T=5000$ steps. For the CN-UCB algorithm, we set $n_0=320$ and $m=2$. For the UCB algorithm, we collected $4$ experimental data from each arm at the beginning, i.e., $36$ initial experimental data in total, before incurring regrets, and set $m=2$ for each round. Assuming the cost of collecting experimental data is around 10 times the cost of observational data, the initial cost of the UCB algorithm was comparable to that of the other three algorithms. Note that $m=2$ experimental data were simulated at each round of CN-UCB and UCB, while we only generated $m=1$ experimental data and $n=1$ observational data for our BA-UCB algorithm. Although these three algorithms collected the same number of data points at each round,  the comparison setting here is in favor of the CN-UCB and the UCB algorithms because experimental data are much more informative and also more expensive than observational data.

\begin{figure}
\includegraphics[width=0.6\textwidth]{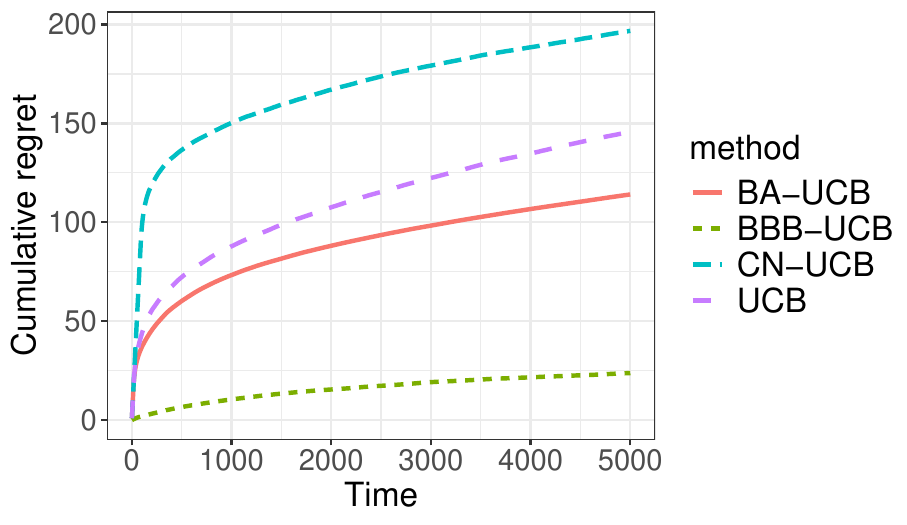}
\caption{Comparison of empirical cumulative regret over $T=5000$ time steps among the BA-UCB, BBB-UCB, CN-UCB, and UCB algorithms on DAGs with $p=10$.}
\label{fig:comp2}
\end{figure}

 As shown in Figure~\ref{fig:comp2}, our BA-UCB algorithm achieved a much lower average cumulative regret than the standard UCB algorithm and the CN-UCB algorithm. %Note that the data sizes at each round were $n=m=1$ for BA-UCB and $m=2$ for UCB and CN-UCB. Our proposed BA-UCB algorithm may cost much less in practice, since generating observational data is much cheaper than experimental data. 
 CN-UCB is seen to have poor performance because for 22 of the 100 Gaussian DAGs, the optimal arm does not correspond to intervention on a parent node of the reward variable. BBB-UCB achieved excellent performance at the cost of significantly higher computation time and resources. More specifically, BBB-UCB required exact computation of the parent set posterior probabilities at each time step, of which the complexity is $O(p\cdot 3^p)$. This is not feasible even for a moderate $p\geq 20$. 

\begin{figure}
\includegraphics[width=0.8\textwidth]{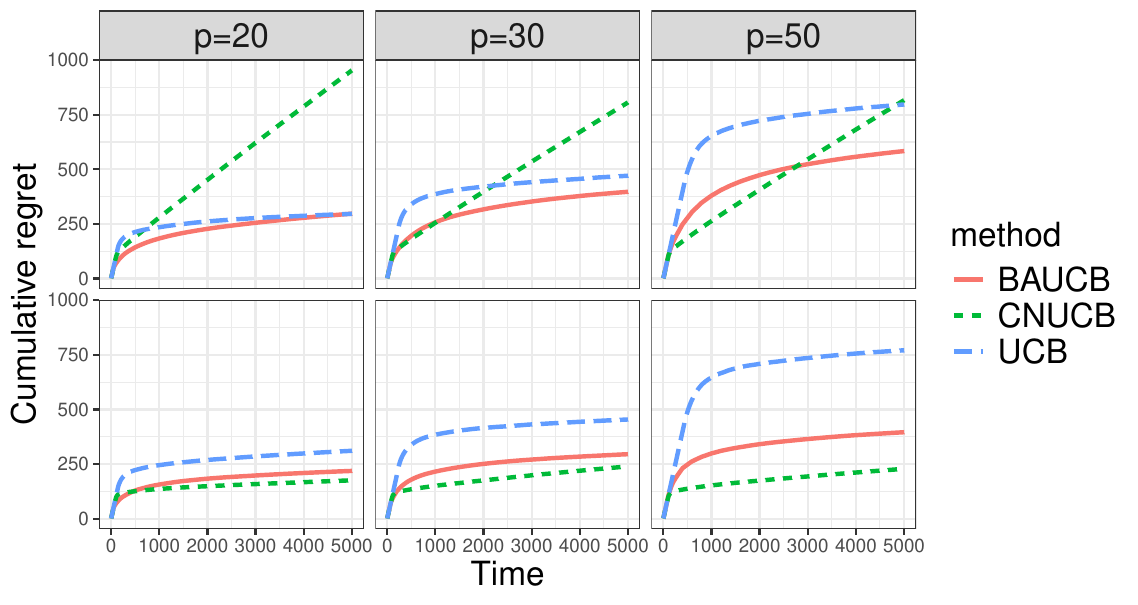}
\caption{Comparison of empirical cumulative regret over 5000 time steps among BA-UCB, CN-UCB, and UCB algorithms for $p=20,30,50$. The top panel reports the cases where the optimal arm is not a parent of the reward variable and the lower panel reports the cases where the optimal arm is a parent of the reward variable.} 
\label{fig:comp1}
\end{figure}

Since the BBB-UCB algorithm does not scale well with $p$, we excluded it from our next comparison for $p = 20, 30, 50$. The parameters were set as $n_0 = 100$ and $m=n=1$ for BA-UCB; $n_0 = 5100$ and $m = 1$ for CN-UCB; $m = 2$ for UCB after a burning period of the first 50 steps. Note that the total number of data generated was the same for all three algorithms over $T=5000$ steps. Since observational data are much cheaper to generate, BA-UCB and CN-UCB would be much less costly than the UCB algorithm in practical applications. 
%The results are reported in Figure~\ref{fig:comp1}.
Given that the performance of CN-UCB strongly depends on whether or not the optimal intervention is on a parent node of the reward variable, we report the results for these two different scenarios separately in Figure~\ref{fig:comp1}.

It is seen from Figure~\ref{fig:comp1} that the BA-UCB algorithm outperforms uniformly the UCB algorithm across all scenarios. The cumulative regrets for both algorithms, as well as the difference between the two, increase as $p$ increases. The CN-UCB algorithm is identical to the standard UCB algorithm but with a reduced set of nodes for intervention, which is usually much smaller than $p$. When the optimal arm indeed corresponds to intervention on a parent of the reward variable, it is not surprising that CN-UCB outperforms our method as shown in the lower panel of the figure. However, in practice, one cannot guarantee that this is always the case. As reported in the top panel, the CN-UCB algorithm shows substantially inferior performance when the optimal arm is not an intervention on a parent node, and in such cases the cumulative regret increases linearly.

\section{Latent confounders}\label{sec:confounder_case}

In many applications, some variables in the underlying causal system are unobserved, which may lead to latent confounding among the observed variables. In such settings, there may not be valid backdoor adjustment set for identifying a causal effect. In this section, we extend the BA-UCB algorithm to accommodate latent confounders, using the acyclic directed mixed graph (ADMG) as a model for the observed variables.

This section is organized as follows. Section~\ref{subsec:admg} introduces the Gaussian ADMG model and the necessary preliminaries. Section~\ref{subsec:alg_confounder} presents a modified BA-UCB algorithm assuming the underlying causal model is an ADMG. Section~\ref{subsec:regret_confounder} establishes regret bounds that distinguish between identifiable and non-identifiable arms. Section~\ref{subsec:simulation_confounder} reports simulation results demonstrating the empirical effectiveness of the proposed algorithm.

\subsection{Gaussian DAGs with latent confounders}\label{subsec:admg}

If some variables in a causal DAG are unobserved, the marginal distribution over the observed variables $\mathcal X=\{X_1,\cdots, X_p\}$ can be represented by an ADMG $\mathcal G$ over nodes $V=\{1,\cdots,p\}$ that contains both directed ($\to$) and bidirected ($\leftrightarrow$) edges. Directed edges encode direct causal relations among the observed variables, while a bidirected edge $i \leftrightarrow j$ indicates the existence of one or more latent confounders influencing both $X_i$ and $X_j$. If $i\to j$, then $i$ is a parent of $j$ and $j$ is a child of $i$. If there is a directed path from $i$ to $j$ or $i=j$, then $i$ is an ancestor of $j$ and $j$ is a descendant of $i$. Non-descendants of a node $i$ consist of every node that is not a descendant of~$i$. For each node $i$ in $\mathcal G$, the parent set $\mathrm{pa}(i)=\{j\in V:j\to i\}$. Similarly, we use $\mathrm{ch}(i)$, $\mathrm{an}(i)$, $\mathrm{de}(i)$, and $\mathrm{nd}(i)$ to denote the set of children, ancestors, descendants, and non-descendants of the node~$i$. The district of $i$, denoted as $\mathrm{dis}(i)$, is the maximal set of nodes connected to $X_i$ through a bidirected path, i.e., its bidirected-connected component. The parent set of a district $D$ in the graph $\mathcal G$ is defined as $\mathrm{pa}(D)=\{\cup_{i\in D} \mathrm{pa}(i)\} \setminus D$, which consists of the out-of-district parents of all nodes in the district $D$. %\qzcmt{subscript ${\mathcal{G}}$ used here but not in the definitions of pa, ch etc of a node $i$. Keep them consisitent}

We assume a linear SEM for the observed variables $\mathcal X$ in an ADMG:
\begin{equation}\label{eq:linear_SEM_confounder}
X_j=\sum_{i\in \mathrm{pa}(j)}\beta_{ij}X_i+\varepsilon_j,\quad j=1,\cdots, p,
\end{equation}
where $\beta_{ij}\neq 0$ if $i\in\mathrm{pa}(j)$ and each error variable $\varepsilon_j\sim\mathcal N(0,\omega_j^2)$ with $\omega_{j}^2>0$. Different from the fully observed case in~\eqref{eq:GaussianBN}, the error term $\varepsilon_1,\cdots,\varepsilon_p$ here are not mutually independent. Let $\Omega=(\omega_{ij}^2)_{p\times p}=\operatorname{Cov}(\varepsilon)$ be the covariance matrix of $\varepsilon=(\varepsilon_1,\ldots,\varepsilon_p)^\top$. Then the off-diagonal entry $\omega_{ij}\ne 0$ if and only if there is a bidirected edge $i\leftrightarrow j$. In matrix form, the model can be written as 
\begin{equation}\label{eq:linear_SEM_conf_matrix}
X=B^\top X+\varepsilon, \quad \varepsilon\sim \mathcal N\left(0,\Omega\right),
\end{equation}
where $X=(X_1,\cdots,X_p)^\top$ and $ B=(\beta_{ij})_{p\times p}$. This model implies that $X\sim\mathcal N(0,\Sigma)$, where $\Sigma=( I- B)^{-\top}\Omega( I-  B)^{-1}$. In this setting, assuming $\Sigma \succ 0$, the regression neighborhood $M_i$ of node $i$ is given by 
\begin{equation}\label{eq:reg_neighbor}
    M_i=\{j\in V_{-i}:(\Sigma^{-1})_{ij}\neq 0\}=\{j\in V_{-i}: X_i\notindep X_j|X_{V\setminus\{i,j\}}\}.
\end{equation}
Let $\mathcal N_i$ denote the collection of all subsets of $M_i$.

%\qzcmt{Include some preliminaries: SEM for Gaussian DAG with latent confounders, ADMG and its terminology e.g. siblings, district, and parent of a district}

\subsection{BA-UCB with latent confounders} \label{subsec:alg_confounder}

Our BA-UCB algorithm can be naturally extended to settings where the underlying causal DAG includes unobserved confounders. A key advantage of BA-UCB is that it does not require structure learning of the entire causal graph. Instead, it leverages observational data to generate estimates when the causal effect of interest is identifiable via covariate adjustment, that is, when there exists at least one valid adjustment set, $\mathcal B_i=\{S:\mu_i(S)=\mu_i\}\neq \emptyset$. However, in the presence of latent confounders, such valid adjustment sets may not always exist. For example, if $X$ and $Y$ are connected by a bidirected edge, no observed variable can block the backdoor path $X\leftrightarrow Y$, and consequently, no valid adjustment set exists. In such cases, the original BA-UCB design, $\widehat{\mathcal{B}}_i=\arg \min_{S} d(0, C I_i(t, S))$, will still select a subset even though there is no valid adjustment set, which may lead to biased causal effect estimation.

To address this issue, we modify Line~\ref{alg:line:backid} in Algorithm~\ref{alg:BA-UCB} to:
\begin{equation}\label{eq:backid_confounder}
    \widehat{\mathcal B}_i =\{S \subseteq \widehat{M}_i: 0\in CI_i(t,S)\},
\end{equation}
where $\widehat{M}_i$ is the estimated regression neighborhood of node $i$.
That is, we only consider subsets $S$ of $\widehat{M}_i$ whose confidence interval $CI_i(t,S)$~\eqref{eq:CI} contains zero. %This modification enables the algorithm to detect situations where no valid adjustment set exists and to avoid misleading updates based on invalid adjustments.
If $\widehat{\mathcal B}\neq \emptyset$, the algorithm proceeds as usual with adjustment set sampling and causal effect estimation (Lines~\ref{alg:line:sample} and \ref{alg:line:estimate}). If $\widehat{\mathcal B}_i= \emptyset$, we default to purely interventional estimation using only the estimates in \eqref{eq:est_int}.

This modification ensures that BA-UCB remains theoretically valid and empirically robust in the presence of latent confounders. When valid adjustment sets exist in the regression neighborhood, the algorithm efficiently integrates observational data to improve estimation accuracy and sample efficiency. When no such sets exist, it automatically falls back to unbiased experimental estimation. As we will show in the subsequent sections, this extension preserves strong theoretical guarantees while maintaining superior empirical performance, even when valid adjustment sets exist for only a subset of the variables.

Similar to the fully observed setting, the restriction of the search for adjustment sets to the regression neighborhood $M_i$ of each variable is for computational efficiency. %Specifically, we estimate $M_i$ as $\widehat{M}_i=\{j:\widehat\beta_{ij}\neq 0\text{ in } X_i\sim X_{-i}\}$, and then attempt to identify valid adjustment sets within the estimated neighborhood. 
We next show that under certain conditions an adjustment set is contained in the regression neighborhood. Define $\mathrm{mb}(i):=\mathrm{pa}(\mathrm{dis}(i))\cup (\mathrm{dis}(i)\setminus\{i\})$ as the Markov blanket of $i$ with respect to an ordering of an ADMG~\citep{richardson2003markov}. %\qzcmt{cite}
The following lemma gives a sufficient condition for $\mathrm{mb}(i)$ to be a valid adjustment set for the causal effect of $X_i$ on $Y$.
If $j\in \mathrm{mb}(i)$, then $X_i\notindep X_j|X_{V\setminus\{i,j\}}$ and thus $\mathrm{mb}(i)\subseteq M_i$ by equality~\eqref{eq:reg_neighbor}.% \qzcmt{do we need with pr 1? Even if not faithful, it seems that $mb(i)$ is still a subset of $M_i$ and a subset of $mb(i)$ could be valid for adjustment. Same comment for the DAG case.}. 

\begin{lemma}\label{proposition:mb}
    Assume the Gaussian ADMG model \eqref{eq:linear_SEM_confounder} with $\Sigma\succ0$. If $\mathrm{dis}(i)\cap \mathrm{de}(i)=\{i\}$ and $Y\notin \mathrm{mb}(i)\cup\{i\}$, then $\mathrm{mb}(i)$ is a valid adjustment set for the causal effect of $X_i$ on $Y$.
\end{lemma}
The proof of Lemma~\ref{proposition:mb} is provided in Section~\ref*{appendix:DAG_confounders} in the supplementary materials. In an ADMG, latent confounding is represented by bidirected edges and induces districts. The condition $\mathrm{dis}(i)\cap \mathrm{de}(i)=\{i\}$ ensures that there are no directed paths from $i$ to other nodes in its district. Under this condition, the node $i$ is fixable, and the post-intervention distribution $p(x_{V\setminus\{i\}}\mid do(x_i))$ is identifiable~\citep{Richardson_2023}. Consequently, the Markov blanket $\mathrm{mb}(i)$ provides a valid adjustment set for identifying the causal effect of $X_i$ on $Y$ as stated in Lemma~\ref{proposition:mb}.

\subsection{Regret analysis}\label{subsec:regret_confounder}

The underlying model assumptions for the modified Algorithm~\ref{alg:BA-UCB} are stated formally in Assumption~\ref{ass:gaussian_admg}.
\begin{assumption}\label{ass:gaussian_admg}
The random vector $X=(X_1,\cdots,X_p)$ satisfies the linear SEM~\eqref{eq:linear_SEM_confounder} associated with an ADMG $\mathcal G$ such that $\beta_{ij}\neq 0$ if and only if $i\in \mathrm{pa}(j)$, $\omega_{ij}\neq 0$ if and only if $i\leftrightarrow j $ for $i\neq j$, and $\Omega\succ 0$.
\end{assumption}

Let $\mathcal{I}_0:=\{i\in \overline V:\mu_i(S)=\mu_i\ \text{ for some } S\in \mathcal N_i\}$ be the set of nodes for which there is a valid adjustment set in its regression neighborhood for identifying its causal effect on $Y$, and let $\mathcal{I}_1=\overline V\setminus \mathcal{I}_0$. %$\{i:\forall S\in\mathcal X\setminus\{X_i\}, \mu_i(S)\neq\mu_i\}$ 
%denote those whose causal effects on $Y$ cannot be identified via adjustment restricted within regression neighborhoods.} 
Recall the constants in \eqref{eq:def_notation} in Section~\ref{subsec:regret_analysis}. Further define $\widetilde\sigma=\max_{i\in\overline V}\widetilde{\sigma}_i$. We now establish regret bounds for the BA-UCB algorithm under the ADMG model. The first result provides a case-dependent upper bound that depends on the reward gaps $\Delta_i$, which separates the regrets from arms in $\mathcal{I}_0$ and $\mathcal{I}_1$. % \qzcmt{In Theorems 1-4, we should add the assumptions of Guassian DAG/ADMG for $X$.}

\begin{theorem}\label{thm3}
Let $\mathcal{R}_T$ be the cumulative regret of Algorithm~\ref{alg:BA-UCB} with the modification \eqref{eq:backid_confounder} and parameters $c\geq 4\sqrt{2}\widetilde\phi$, $c_2=2\sqrt{3}$, and $c_3\geq \max\{64,32\psi^2/\delta^2\}$. If Assumptions~\ref{ass:gaussian_admg}, \ref{ass:id} and~\ref{ass:n0} hold,  then after $T$ rounds the cumulative regret $\mathcal{R}_T$ is at most
\begin{equation}\label{eq:thm3}
\begin{aligned}
      \frac{20c^2\log T}{m+n}\sum_{i\in\mathcal I_0}\frac{\widetilde\eta_i^2}{\Delta_i}+\frac{10c^2\log T }{m}\sum_{i\in\mathcal I_1}\frac{\widetilde\sigma_i^2}{\Delta_i}+\left(\frac{c_3\log T}{m}+C_5\right)\sum_{i}\Delta_i,
\end{aligned}
\end{equation}
where $C_5$ is a constant independent of $T$.
\end{theorem}
In general, $C_5$ depends on the covariance matrix $\Sigma$, the initial observational sample size $n_0$, and the per-round observational data size $n$, and its exact expression is provided in Section~\ref*{appendix:constant} of the supplementary materials. See Section~\ref*{appendix:DAG_confounders} in the supplementary materials for a proof of this theorem.

This regret bound explicitly reflects the heterogeneity across arms in terms of causal effect identifiability. For arms in $\mathcal{I}_0$, our algorithm benefits from both observational and experimental data, reflected in the $(m+n)^{-1}$ scaling. For arms in $\mathcal{I}_1$, only experimental samples contribute to reducing uncertainty, leading to a larger regret term. This decomposition generalizes the result from the fully observed setting (Theorem~\ref{thm1}): If the causal effects of all arms are identifiable, i.e., $\mathcal{I}_1=\emptyset$, the bound \eqref{eq:thm3} reduces to the form of Theorem~\ref{thm1}; conversely, if none of them is identifiable, i.e., $\mathcal{I}_0=\emptyset$, the regret scales as $O((\log T)/m)$, where $m$ is the experimental sample size. This latter scaling is identical to that of the standard UCB.

Similar to Theorem~\ref{thm2}, we also establish a $\Delta$-independent upper bound that does not depend on the reward gaps $\Delta_i$, under the additional assumption that all rewards and upper confidence bounds are bounded to $[-1,1]$. 
\begin{theorem}\label{thm4}
    Assume $\mu_i\in[-1,1],\forall i\in\{1,\cdots,p-1\}$. Let $\mathcal R_T$ be the cumulative regret of Algorithm~\ref{alg:BA-UCB} with the modification \eqref{eq:backid_confounder} and parameters $c\geq 4\widetilde\phi$, $c_2=2\sqrt{3}$, and $c_3\geq \max\{32, 32\psi^2/\delta^2\}$. If Assumptions~\ref{ass:gaussian_admg}, \ref{ass:id} and \ref{ass:n0} holds, then after $T$ rounds the cumulative regret $\mathcal{R}_T$ is at most
\begin{equation}\label{eq:thm4}
\begin{aligned}
   2c\sqrt{\log T}\left[\frac{\widetilde{\eta}\left(\sqrt{|\mathcal I_0|(n+m)T+n_0|\mathcal I_0|^2}-\sqrt{n_0}|\mathcal I_0|\right)}{n+m}+\widetilde\sigma\sqrt{\frac{|\mathcal I_1|T}{m}}\right]+\frac{2Kc_3\log T}{m}+C_6,
\end{aligned}
\end{equation}
where $C_6$ is a constant independent of $T$.
\end{theorem}
In general, $C_6$ depends on $\Sigma$, $n_0$, $m$, and $n$, and the exact expression is provided in Section~\ref*{appendix:constant} of the supplementary materials. See Section~\ref*{appendix:DAG_confounders} in the supplementary materials for a proof of this theorem.

This regret bound captures the worst-case scenario across all instances of reward values and avoids dependence on the reward gaps. Similar to Theorem~\ref{thm3}, it separates arms according to identifiability. As discussed in Section~\ref{subsec:regret_implication} , if a sufficient amount of prior observational data is available, i.e., with a large initial sample size $n_0$, the first term in the cumulative regret upper bound in \eqref{eq:thm4} 
\[
\frac{2c\widetilde\eta\sqrt{\log T}}{n+m}\left(\sqrt{|\mathcal I_0|(n+m)T+n_0|\mathcal I_0|^2}-\sqrt{n_0}|\mathcal I_0|\right)=O\left(\sqrt{T\log T/(n+m)}\right).
\]
%\qzcmt{it may not be very clear what you meant by the first term in (20); maybe define the terms in (20)?} 
 As such, it does not depend on the number of identifiable arms $|\mathcal I_0|$, in contrast to the standard MAB bound $O(\sqrt{|\mathcal I_0|T\log T/(n+m)})$. Consequently, the overall cumulative regret can be significantly reduced compared to traditional MAB algorithms that rely solely on experimental data, especially if the number of identifiable arms $|\mathcal I_0|$ is large.

\subsection{Simulation results}\label{subsec:simulation_confounder}

We evaluate the empirical performance of the BA-UCB algorithm in comparison with the standard UCB algorithm when the underlying causal DAG includes latent confounders. DAGs were randomly generated with a total number of variables (including unobserved confounders) $N\in\{10,20,30\}$ and a maximum in- and out-degree of $3$. In each graph, $10 \%$ of the variables were chosen as unobserved confounders, resulting in $p\in\{9,18,27\}$ observed variables. To control the complexity of the induced ADMG, confounders were selected such that they did not share any common child.

For BA-UCB, the initial observational sample size was set to $n_0=160$, $340$, and $520$ for $p=9$, $18$, and $27$, respectively. We generated $20$ experimental samples for each arm before incurring any regret, in order to satisfy the minimum intervention requirement in Algorithm~\ref{alg:BA-UCB} Line~\ref{alg:line:min_time}. %The initial experimental sample size \qzcmt{what is initial experimental sample?} was $m_0=20$ per arm, 
Both observational and experimental sample sizes per round were set to $n=m=5$. We set $c=\sqrt{2}$, $\tau= 0.1 / p$ as the threshold for p-values when estimating the regression neighborhood through linear regression, and $c_2=1/(2 \sqrt{2})$ as the constant in~\eqref{eq:CI}. To reduce computation cost, we only consider $\{S\in\widehat{\mathcal N_i}: |S|\leq 4\}$ in the identification of candidate backdoor adjustment sets. For the standard UCB algorithm, $22$ experimental samples were generated for each arm before incurring regret, and the per-round sample size was $m=10$. Under these choices, the computational costs of thet two algorithms were comparable. %ensuring comparable total sampling budgets. Given that experimental data were assumed to be ten times more costly than observational data, this allocation provided a fair and slightly favorable comparison for the UCB baseline. 
Each algorithm was executed five times on every generated graph.

\begin{figure}
\includegraphics[width=0.8\textwidth]{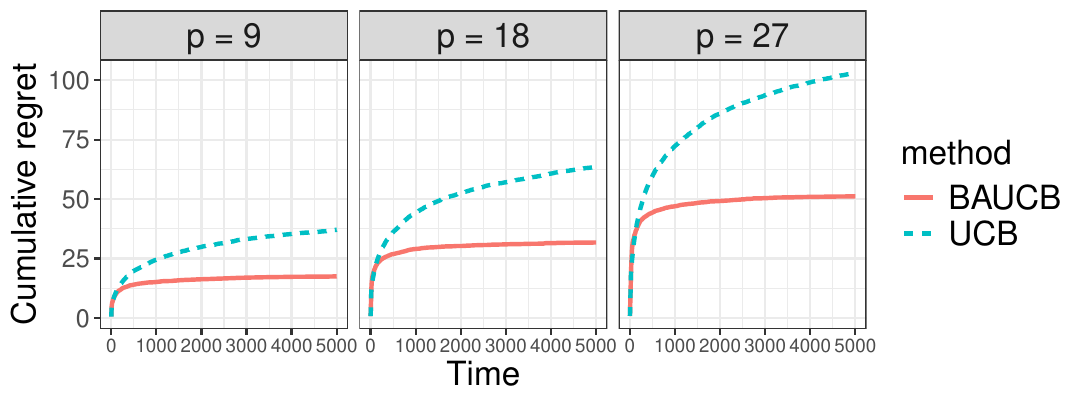}
\caption{Comparison of empirical cumulative regrets over 5000 time steps between the BA-UCB and standard UCB algorithms on graphs with latent confounders. The number of observed variables 
$p=9,18,27$, corresponding to a total of $N=10,20,30$ variables, respectively.}
 \label{fig:comp_confounder}
\end{figure}

Figure~\ref{fig:comp_confounder} summarizes the average results over $100$ independently simulated graphs. For each graph, we computed the median cumulative regret over five runs, and then averaged the median cumulative regrets across the graphs. %(excluding the single worst-performing graph for a robust average). 
The results clearly demonstrate that BA-UCB consistently achieves lower cumulative regret than the standard UCB algorithm for all values of $p$, confirming its advantages in using observational data in the presence of latent confounders.

\begin{figure}
\includegraphics[width=0.7\textwidth]{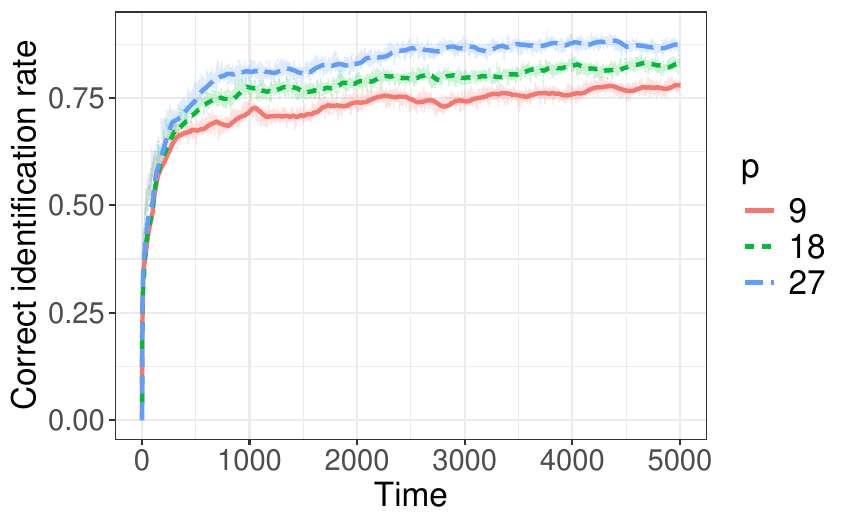}
\caption{Average identification rate of a correct adjustment set for the optimal arm at each time step in the presence of latent confounders. Solid curves are smoothing averages with a 100-step window for better visualization.} %for numbers of observed variables $p=9,18,27$, corresponding to total variable counts of $10$, $20$, and $30$, respectively.}}
    \label{fig:parent_id}
\end{figure}

We also examined whether BA-UCB behaved as expected in terms of identifying adjustment sets. As the number of rounds increases, the algorithm accumulates more observational and experimental data, particularly for the optimal arm, leading to more reliable estimation of causal effects and more accurate identification of valid adjustment sets. %Since BA-UCB was not specifically designed for structure learning, we evaluated its performance based on quantitative closeness: 
In our numerical results, an identified adjustment set for the causal effect $\mu_i$ is regarded as correct (up to certain numerical precision) if %its theoretical observational estimate $\mu_i(S)$ differed from the true causal effect $\mu_i$ by less than 0.01 , i.e., 
$|\mu_i(S)-\mu_i|<0.01$, where $\mu_i(S)$ is the theoretical value of the adjusted estimate of the causal effect  as defined {in \eqref{eq:mu_S_def}}.
%when estimating the effect of $X_i$ on $Y$, and $\mu_i$ represents the true causal effect of $X_i$ on $Y$.
At each time step, we computed the frequency that the sampled adjustment set for the optimal arm is a true adjustment set, %collection $\mathcal B^*$, 
among 5 independent runs and 100 graphs. Figure~\ref{fig:parent_id}  %\qzedit{[This was referred to Figure 5; I corrected. Please check all references are correct]} 
presents this identification rate over time for different graph sizes $p$, showing a steady improvement as BA-UCB accumulates more data. In particular, the rate of correct identification increased to more than $75\%$ near $T=5000$ for all graph sizes.

\section{Discussion}\label{sec:discussion}

In this paper, we studied the causal bandit problem under the setting of Gaussian DAGs. Without assuming any prior knowledge about the underlying DAG structure, we proposed the BA-UCB algorithm which identifies candidate adjustment sets using estimates from both observational and experimental data. These candidate sets are used to construct weighted upper confidence bounds for making sequential decisions. The reason for updating adjustment sets along with more interventions is that the structure of a DAG cannot be fully recovered even with a sufficiently large number of observational data due to Markov equivalence. Therefore, our approach is fundamentally different from other existing works that use prior data or knowledge to reduce the action set before a sequential bandit algorithm.

Our upper bound on the cumulative regret is of order $O(\sqrt{{T\log T}/{(n+m)}})$ with a large number of initial observations, which does not depend on the size of action set and decreases with the
number of data collected at each round. Empirical study shows that the cumulative regret of our algorithm is smaller than the UCB and the CN-UCB algorithms, especially when the optimal arm is not a parent of the reward. Moreover, it is much more scalable with the graph size than the BBB-UCB algorithm.

Finally, we extended the BA-UCB framework to settings with latent confounders. In these scenarios, valid adjustment sets may not exist for every arm. The extended algorithm automatically detects such cases and relies solely on experimental data when necessary, ensuring robustness to unobserved confounding.

Several directions remain open for future work. First, while our analysis is developed for Gaussian DAGs, extensions to linear structural equation models with non-Gaussian or heterogeneous noise distributions deserve further investigation. Second, incorporating richer prior structural knowledge or graph recovery methods could further improve efficiency while maintaining robustness. Lastly, it would be valuable to explore extensions of BA-UCB to nonlinear or nonparametric causal models and to develop regret guarantees under broader model assumptions.

%%%%%%%%%%%%%%%%%%%%%%%%%%%%%%%%%%%%%%%%%%%%%%
%% Single Appendix:                         %%
%%%%%%%%%%%%%%%%%%%%%%%%%%%%%%%%%%%%%%%%%%%%%%
%\begin{appendix}
%\section*{???}%% if no title is needed, leave empty \section*{}.
%\end{appendix}
%%%%%%%%%%%%%%%%%%%%%%%%%%%%%%%%%%%%%%%%%%%%%%
%% Multiple Appendixes:                     %%
%%%%%%%%%%%%%%%%%%%%%%%%%%%%%%%%%%%%%%%%%%%%%%
%\begin{appendix}
%\section{???}
%
%\section{???}
%
%\end{appendix}

%%%%%%%%%%%%%%%%%%%%%%%%%%%%%%%%%%%%%%%%%%%%%%
%% Support information, if any,             %%
%% should be provided in the                %%
%% Acknowledgements section.                %%
%%%%%%%%%%%%%%%%%%%%%%%%%%%%%%%%%%%%%%%%%%%%%%
%\begin{acks}[Acknowledgments]
% The authors would like to thank ...
%\end{acks}
%%%%%%%%%%%%%%%%%%%%%%%%%%%%%%%%%%%%%%%%%%%%%%
%% Funding information, if any,             %%
%% should be provided in the                %%
%% funding section.                         %%
%%%%%%%%%%%%%%%%%%%%%%%%%%%%%%%%%%%%%%%%%%%%%%
\begin{funding}
 This work was supported in part by NSF grant DMS-2305631 and NIH grant R01GM163245.
\end{funding}

%%%%%%%%%%%%%%%%%%%%%%%%%%%%%%%%%%%%%%%%%%%%%%
%% Supplementary Material, including data   %%
%% sets and code, should be provided in     %%
%% {supplement} environment with title      %%
%% and short description. It cannot be      %%
%% available exclusively as external link.  %%
%% All Supplementary Material must be       %%
%% available to the reader on Project       %%
%% Euclid with the published article.       %%
%%%%%%%%%%%%%%%%%%%%%%%%%%%%%%%%%%%%%%%%%%%%%%
% \begin{supplement}
% \stitle{???}
% \sdescription{???.}
% \end{supplement}

%%  MR numbers will be added by VTeX.                      %%
%%                                                         %%
%%  Use \cite{...} to cite references in text.             %%
%%                                                         %%
%%%%%%%%%%%%%%%%%%%%%%%%%%%%%%%%%%%%%%%%%%%%%%%%%%%%%%%%%%%%%

%% if your bibliography is in bibtex format, uncomment commands:
\bibliographystyle{imsart-number} % Style BST file (imsart-number.bst or imsart-nameyear.bst)
\bibliography{bibliography}       % Bibliography file (usually '*.bib')

%% or include bibliography directly:
% \begin{thebibliography}{}
% \bibitem{b1}
% \end{thebibliography}

% \end{document}

\clearpage
% arXiv-friendly supplementary numbering to avoid duplicate hyperref page anchors
\pagenumbering{arabic}
\setcounter{page}{1}
\renewcommand{\thepage}{S\arabic{page}}
\makeatletter
\providecommand*\theHpage{}
\renewcommand*\theHpage{supp.\arabic{page}}
\makeatother
\SuppNumbering

\title{Supplementary Material: Proofs and Technical Details}
% \begin{supplement}
% \stitle{???}
% \sdescription{???.}
% \end{supplement}

% \section{Proofs and technical details}\label{appendix:proof}

% In this section, we first provide proof for the upper bounds on the cumulative regret when the underlying DAG is given. Then we modify the proof to extend it to the DAG unknown case. 

%\appendix
% \qzcmt{Label equations by section so they appear as (1.1), (3.2), etc, to distinguish from the equations in main text}
% \qzcmt{I added 'S' to numbering. See local def.}

\section{Constants in the main theorems}
\label{appendix:constant}

Recall some relevant assumptions and notations in the main paper:
\begin{itemize}
    \item The joint distribution of $X$ is defined by the linear SEM 
    \begin{equation}\label{eq:GaussianBN-supp}
X_j=\sum_{i\in \mathrm{pa}(j)}\beta_{ij}X_i+\varepsilon_j,\quad j=1,\cdots, p,
\end{equation}
where $\beta_{ij}\neq 0$ if $i\in\mathrm{pa}(j)$ and each error variable $\varepsilon_j\sim\mathcal N(0,\omega_j^2)$ with $\omega_{j}^2>0$. This implies that $X\sim\mathcal N( 0,\Sigma)$, where $\Sigma=( I- B)^{-\top}\Omega( I-  B)^{-1}$, $ B=(\beta_{ij})_{p\times p}$, and $\Omega=(\omega_{ij}^2)_{p\times p}=\operatorname{Cov}(\varepsilon)\succ0$.
    \item $Y|do(X_i=x_i)\sim \mathcal N(\mu_i,\widetilde\sigma_i^2)$.
    \item $Y|X_i,X_S \sim \mathcal N\left(X_i\mu_i(S)+\beta_S^\top X_S, \zeta_{i,S}^2\right)$.
    \item $X_i| X_{-i}\sim \mathcal N(\bar\mu_i,\bar\sigma_i^2)$, where $\bar\mu_i=\sum_{j:j\neq i}\bar\beta_{ij}X_j$ and $\bar\beta_{ij}= 0$ iff $X_i\indep X_j| X_{-ij}$.
    \item The following constants have been defined in the main text:
    \begin{equation}\label{eq:def_notation_supp}
    \begin{aligned}
\widetilde{\eta}_i^2&=\underset{S\in\mathcal B_i}{\max}\left\{ \frac{9\zeta_{i,S}^2}{\gamma_{\min}(\Sigma)},\tilde\sigma_i^2\right\},\quad\quad
\widetilde\eta^2=\max_i\widetilde{\eta}_i^2,\\
\widetilde\phi^2 &= \underset{i}{\max}\,\underset{S\in\mathcal B_i}{\max}\max\left\{\frac{9\zeta_{i,S}^2}{\gamma_{\min}(\Sigma)\tilde\sigma_i^2},\frac{25\tilde\sigma_i^2\gamma_{\max}(\Sigma)}{9\zeta_{i,S}^2}\right\},\\
 \psi^2&=\max_i\max_{S'\in \mathcal N_i\setminus\mathcal B_i}\max\left\{\frac{9\zeta_{i,S'}^2}{\gamma_{\min}(\Sigma)},\widetilde\sigma_i^2\right\}.
    \end{aligned}
\end{equation}
    \item $s_0$ denotes the maximum of in- and out-degrees among all nodes except the reward variable.
\end{itemize}

Further, let $s_1=\max_{i}|\mathcal{M}_i|$ denote the maximum size of the regression neighborhood, and $s_2=\underset{i}{\max}|\mathcal N_i\setminus \mathcal B_i|$ be the maximum number of subsets in the regression neighborhood that do not yield the true causal effect of $X_i$ on $Y$. Let
\begin{align*}
    \psi^2&=\max_i\max_{S'\in \mathcal N_i\setminus\mathcal B_i}\max\left\{\frac{9\zeta_{i,S'}^2}{\gamma_{\min}(\Sigma)},\widetilde\sigma_i^2\right\},\\
    \bar\beta_i&=\min_{j:\bar\beta_{ij}\neq 0}|\bar\beta_{ij}|,\\
\alpha&=\min_i\frac{\sqrt{\gamma_{\min}(\Sigma)}\bar\beta_i}{12\sqrt{3}\bar\sigma_i},\\
\ell_0&=\left\lceil \frac{9\operatorname{tr}(\Sigma)}{n\gamma_{\min}(\Sigma)}\right\rceil.
\end{align*} 
Then the constants in Theorems 1, 2, 4, and 5 are given by:
\begin{align*}
C_3&=\pi^2\left(s_2+\frac{4}{3}+\frac{1}{3}\exp\left(-\frac{n_0-s_0-2}{16}\right)\right)+\ell_0+\left(\frac{c_3}{n}\right)^2+1\\
&\quad +\frac{4K}{n\alpha^2\exp\left(\alpha^2(n_0+n\ell_0)\right)}+\frac{36(s_1+s_2+K+3)}{n\exp\left((n_0+n\ell_0-K-1)/18\right)},\\
C_4 &=\pi^2\left(2s_2+\frac{7}{3}+\frac{1}{6}\exp\left(-\frac{n_0-s_0-2}{16}\right)\right) +2\ell_0+2\left(\frac{c_3}{n}\right)^2+2K\\
&\quad +\frac{8K}{n\alpha^2\exp\left(\alpha^2(n_0+n\ell_0)\right)}+\frac{72(s_1+s_2+K+3)}{n\exp\left((n_0+n\ell_0-K-1)/18\right)},\\
C_5&=\pi^2\left(s_2+\frac{2}{3}+\frac{1}{3}\exp\left(-\frac{n_0-s_0-2}{16}\right)\right)+\ell_0+ \left(\frac{c_3}{n}\right)^2+1\\
&\quad +\frac{4K}{n\alpha^2\exp\left(\alpha^2(n_0+n\ell_0)\right)} +\frac{36(s_1+s_2+K+2)}{n\exp\left((n_0+n\ell_0-K-1)/18\right)},\\
C_6&=\pi^2\left(2s_2+\frac{1}{2}+\frac{1}{6}\exp\left(-\frac{n_0-s_0-2}{16}\right)\right)+2\ell_0+2\left(\frac{c_3}{n}\right)^2+2K\\
&\quad +\frac{8K}{n\alpha^2\exp\left(\alpha^2(n_0+n\ell_0)\right)}+\frac{72(s_1+s_2+K+2)}{n\exp\left((n_0+n\ell_0-K-1)/18\right)}.
\end{align*}
Note that when $n_0$ is relatively large, the terms with $\exp(-n_0)$ are usually small. Accordingly, these constants can be simplified when $n_0$ is large:
\begin{align*}
    &C_3=\pi^2\left(s_2+\frac{4}{3}\right)+\ell_0+\left(\frac{c_3}{n}\right)^2+1+O\left(\exp\left(-Mn_0\right)\right),\\
&C_4 =\pi^2\left(2s_2+\frac{7}{3}\right) +2\ell_0+2\left(\frac{c_3}{n}\right)^2+2K+O\left(\exp\left(-Mn_0\right)\right),\\
&C_5=\pi^2\left(s_2+\frac{2}{3}\right)+\ell_0+ \left(\frac{c_3}{n}\right)^2+1+O\left(\exp\left(-Mn_0\right)\right),\\
&C_6=\pi^2\left(2s_2+\frac{1}{2}\right)+2\ell_0+2\left(\frac{c_3}{n}\right)^2+2K+O\left(\exp\left(-Mn_0\right)\right),
\end{align*}
where $M=\min\left\{1/18,\alpha^2\right\}$.

\section{Some general bounds}\label{appendix:rho_bounds}
\begin{definition}\label{def:gaussian-ensemble}
    If each $x_i\in\mathbb R^p$ is drawn independently from $\mathcal{N}(0,\Sigma)$, then the random matrix $\mathbf{X} \in \mathbb{R}^{n \times p}$, with $x_i^\top$ as its $i^{\text {th }}$ row, is said to be drawn from the $\Sigma$-Gaussian ensemble.
\end{definition}

\begin{lemma}\label{thm:rho}
    Let $\mathbf X\in \mathbb R^{N_o(t)\times p}$ be drawn according to the $\Sigma$-Gaussian ensemble with $\Sigma\succ 0$. For any $i\in V=\{1,\cdots, p\}$ and any $S\subseteq V\setminus \{i\}$, let $\mathbf{{X}}_{i,S}=[\mathbf X_i\mid  \mathbf X_S]$ and $\rho_i(t,S)=\left[\left(\frac{\mathbf{{X}}_{i,S}^\top \mathbf{{X}}_{i,S}}{N_o(t)}\right)^{-1}\right]_{i,i}$. Then
    \begin{align*}
      \left(\frac{\sqrt{N_o(t)}}{\sigma_{\max}(\mathbf X)}\right)^2 \leq \rho_i(t,S)\leq \left(\frac{\sqrt{N_o(t)}}{\sigma_{\min}(\mathbf X)}\right)^2.
    \end{align*}
    When $N_o(t)\geq \max\left\{\frac{9\operatorname{tr}(\Sigma)}{\gamma_{\min}(\Sigma)},p\right\}$, we have
    \begin{equation}\label{eq:rho}
        \begin{aligned}
            \mathbb P\left( \rho_i(t,S)\geq \frac{9}{\gamma_{\min}(\Sigma)},\text{ for some }i\in V\text{ and }S\subseteq V\setminus\{i\}\right)&\leq \mathbb P\left( \frac{\sigma_{\min}(\mathbf X)}{\sqrt{N_o(t)}}\leq\frac{1}{3}\gamma_{\min}(\sqrt{\Sigma})\right)\\
            &\leq \exp\left(-\frac{N_o(t)}{18}\right),\\
            \mathbb P\left( \rho_i(t,S)\leq \frac{9}{25\gamma_{\max}(\Sigma)},\text{ for some }i\in V\text{ and }S\subseteq V\setminus\{i\}\right)&\leq \mathbb P\left( \frac{\sigma_{\max}(\mathbf X)}{\sqrt{N_o(t)}}\geq\frac{5}{3}\gamma_{\max}(\sqrt{\Sigma})\right)\\
            &\leq \exp\left(-\frac{N_o(t)}{18}\right).
        \end{aligned}
    \end{equation}
\end{lemma}

\begin{proof}
From Cauchy interlacing theorem, for all $i\in V$ and all $S\subseteq V\setminus\{i\}$,
\begin{align*}
\rho_i(t,S)&=\left[\left(\frac{\mathbf{{X}}_{i,S}^\top \mathbf{X}_{i,S}}{N_o(t)}\right)^{-1}\right]_{i,i}\leq \gamma_{\max}\left[\left(\frac{\mathbf{X}_{i,S}^\top \mathbf{X}_{i,S}}{N_o(t)}\right)^{-1}\right]=\left[\gamma_{\min}\left(\frac{\mathbf{X}_{i,S}^\top \mathbf{X}_{i,S}}{N_o(t)}\right)\right]^{-1}\\
    &\leq\left[\gamma_{\min}\left(\frac{\mathbf X^\top \mathbf X}{N_o(t)}\right)\right]^{-1}= \left(\frac{\sqrt{N_o(t)}}{\sigma_{\min}(\mathbf X)}\right)^2.
\end{align*}
Theorem~6.1 in \cite{wainwright2019high} states that:  Let $\mathbf X\in \mathbb R^{n\times d}$ be drawn according to the $\Sigma$-Gaussian ensemble. Then for all $\delta>0$, the maximum singular value $\sigma_{\max}(\mathbf X)$ satisfies the upper deviation inequality
\begin{align*}
    \mathbb P\left[\frac{\sigma_{\max}(\mathbf X)}{\sqrt n}\geq \gamma_{\max}(\sqrt{\Sigma})(1+\delta)+\sqrt{\frac{\operatorname{tr}(\Sigma)}{n}}\right]\leq e^{-n\delta^2/2}.
\end{align*}
Moreover, for $n\geq d$, the minimum singular value $\sigma_{\min}(\mathbf X)$ satisfies the analogous lower deviation inequality
\begin{align*}
    \mathbb P\left[\frac{\sigma_{\min}(\mathbf X)}{\sqrt n}\leq \gamma_{\min}(\sqrt{\Sigma})(1-\delta)-\sqrt{\frac{\operatorname{tr}(\Sigma)}{n}}\right]\leq e^{-n\delta^2/2}.
\end{align*}
Since $N_o(t)\geq \frac{9\operatorname{tr}(\Sigma)}{\gamma_{\min}(\Sigma)}$, choose $\delta=\frac{2}{3}-\sqrt{\frac{\operatorname{tr}(\Sigma)}{N_o(t)\gamma_{\min}(\Sigma)}}\geq \frac{1}{3}$, and we have 
\begin{align*}
    \mathbb P\left( \frac{\sigma_{\min}(\mathbf X)}{\sqrt{N_o(t)}}\leq\frac{1}{3}\gamma_{\min}(\sqrt{\Sigma})\right)\leq \exp\left(-\frac{N_o(t)\delta^2}{2}\right)\leq \exp\left(-\frac{N_o(t)}{18}\right).
\end{align*}
Similarly, we have 
\begin{align*}
\rho_i(t,S)\geq  \left(\frac{\sqrt{N_o(t)}}{\sigma_{\max}(\mathbf X)}\right)^2,\text{ for all } i\in V \text{ and } S\subseteq V\setminus\{i\},
\end{align*}
and
\begin{align*}
    \mathbb P\left( \frac{\sigma_{\max}(\mathbf X)}{\sqrt{N_o(t)}}\geq\gamma_{\max}(\sqrt{\Sigma})(1+\delta)+\sqrt{\frac{\operatorname{tr}(\Sigma)}{N_o(t)}} \right)\leq \exp\left(-N_o(t)\delta^2/2\right).
\end{align*}
Let $\delta=\frac{2}{3}-\sqrt{\frac{\operatorname{tr}(\Sigma)}{N_o(t)\gamma_{\max}(\Sigma)}}\geq \frac{1}{3}$. Then,
\begin{align*}
    \mathbb P\left( \frac{\sigma_{\max}(\mathbf X)}{\sqrt{N_o(t)}}\geq\frac{5}{3}\gamma_{\max}(\sqrt{\Sigma})\right)\leq \exp\left(-\frac{N_o(t)}{18}\right).
\end{align*}
\end{proof}

\begin{remark}
    Lemma~\ref{thm:rho} applies to not only $\rho_i(t,S)$. In fact, when $\mathbf X$ is drawn from a $\Sigma$-Gaussian ensemble with $\Sigma\succ 0$, we can similarly prove the following results:
    \begin{align*}
            &\mathbb P\left( \left[\left(\frac{\mathbf{{X}}_{S}^\top \mathbf{{X}}_{S}}{N_o(t)}\right)^{-1}\right]_{j,j}\geq \frac{9}{\gamma_{\min}(\Sigma)},\text{ for some }S\subseteq V\text{ and } j\in S\right)\leq \exp\left(-\frac{N_o(t)}{18}\right),\\
            &\mathbb P\left(\left[\left(\frac{\mathbf{{X}}_{S}^\top \mathbf{{X}}_{S}}{N_o(t)}\right)^{-1}\right]_{j,j}\leq \frac{9}{25\gamma_{\max}(\Sigma)},\text{ for some }S\subseteq V\text{ and } j\in S\right)\leq \exp\left(-\frac{N_o(t)}{18}\right).
        \end{align*}
\end{remark}

\begin{lemma}\label{thm:chi_bound}
    If $X~\sim \chi_n^2$, then
   \begin{equation*}
       \begin{aligned}
        &\mbp(X\geq 5n/2)\leq \exp\left(-n/4\right),\\
        & \mbp(X\leq n/3)\leq \exp\left(-n/9\right).
    \end{aligned}
   \end{equation*} 
\end{lemma}
\begin{proof}
   From Lemma 1 in \cite{laurent2000adaptive},
\begin{equation*}
\begin{aligned}
&\mathbb{P}\left(\chi_{n}^{2}-n \geq 2 \sqrt{n} \sqrt{x}+2 x\right) \leq e^{-x}, \\
&\mathbb{P}\left(\chi_{n}^{2}-n \leq-2 \sqrt{n} \sqrt{x}\right) \leq e^{-x}.
\end{aligned}
\end{equation*} 
Let $x=\frac{n}{4}$ and $x=\frac{n}{9}$ respectively, then $\mathbb{P}\left(\chi_{n}^{2} \geq 5n/2\right)\leq e^{-n/4}$ and $\mbp \left(\chi_n^2\leq n/3\right)\leq e^{-n/9}$.
\end{proof}

\begin{lemma}\label{thm:t_bound}
    If $X\sim t_n$, then for any $\delta>0$, 
    \begin{equation*}
        \mathbb{P}(|X|\geq \delta)\leq 2e^{-\delta^2/4}+e^{-n/16}.
    \end{equation*}
\end{lemma}

\begin{proof}
    \begin{align*}
        \mbp(|X|\geq\delta)=\mbp\left(\frac{|\mathcal N(0,1)|}{\sqrt{\chi_{n}^2/n}}\geq\delta\right)\leq\mbp\left(|\mathcal N(0,2)|\geq \delta\right)+\mbp\left(\chi_n^2\leq n/2\right).
    \end{align*}
    From Lemma~1 in \cite{laurent2000adaptive}, $\mathbb{P}\left(\chi_{n}^{2}-n \leq-2 \sqrt{n} \sqrt{x}\right) \leq e^{-x}$. Let $x=n/16$, then 
    \begin{align*}
        \mbp(\chi_n^2\leq n/2)\leq\exp(-n/16).
    \end{align*}
    Combined with Gaussian tail probability bound, we have
    \begin{align*}
        \mbp(|X|\geq\delta)\leq 2\exp(-\delta^2/4)+\exp(-n/16).
    \end{align*}
\end{proof}

\section{When the underlying DAG is given}\label{appendix:DAGgiven}

When the DAG is given, the parent set of each node $X_i$ satisfies the backdoor criterion relative to $(X_i,Y)$. So Algorithm~\ref*{alg:BA-UCB} in the main paper can be reduced to Algorithm~\ref{alg:Combined-UCB}.

\begin{algorithm}
\caption{Combined-UCB for Gaussian Causal Bandits (DAG given)}\label{alg:Combined-UCB}
\begin{algorithmic}[1]
\Require DAG $\mathcal G$, Observational data $\mathbf D_0$, parameters $m$, $n$, $c$, $c_3$, rounds $T$.
\For{$t=1,\cdots, T$}
\If{There is a node $i$ which has been intervened less than $\left\lceil\frac{c_3\log t}{n+m} \right\rceil$ times}
    \State Choose intervention node $A_{t} = i$;\label{alg:line:min_time_given}
\Else
    \State Calculate $\hat\mu_i(t-1, \mathrm{pa}(i))$ and $\hat\sigma_i(t-1,\mathrm{pa}(i))$ for each $i\in\{1,\cdots, K\}$;
    \State Choose intervention node $A_{t}\in \underset{i \in \{1, \ldots, K\}}{\operatorname{ argmax }}\left\{\hat{\mu}_{i}(t-1,\mathrm{pa}(i))+\hat{\sigma}_{i}(t-1,\mathrm{pa}(i)) c \sqrt{\log (t-1)}\right\}$.
\EndIf
\State Sample $m$ data points under intervention $A_{t+1}$ and $n$ observational data.
\EndFor
\Ensure Optimal intervention $A_T$ and estimated reward $\hat \mu^*$.
\end{algorithmic}
\end{algorithm}

Define the following constants:
\begin{align*}
    &\phi^2= \underset{1\leq i\leq K}{\max}\max\left\{\frac{9\zeta_{i}^2}{\gamma_{\min}(\Sigma)\tilde\sigma_i^2},\frac{25\tilde\sigma_i^2\gamma_{\max}(\Sigma)}{9\zeta_{i}^2}\right\}, \text{ where }\zeta_i^2=\zeta_{i,\mathrm{pa}(i)}^2,\\  
    &\eta^2=\max_i\eta_i^2,\text{ where }{\eta}_i^2=\max\left\{ \frac{9\zeta_{i}^2}{\gamma_{\min}(\Sigma)},\widetilde\sigma_i^2\right\},\\
    &\ell_0 =\left\lceil\frac{9\operatorname{tr}(\Sigma)}{n\gamma_{\min}(\Sigma)}\right\rceil.
\end{align*}
The theoretical results on the cumulative regret are summarized in the following theorems. 

\subsection{Case-dependent bound}

\begin{theorem}\label{thm:DAGgiven_casedepend}
  Under Assumptions~\ref*{ass:gaussian_dag}, \ref*{ass:id}, and \ref*{ass:n0} in the main paper, the cumulative regret $\mathcal R_T$ of Algorithm~\ref{alg:Combined-UCB}, with parameters $c\geq 4\sqrt{2}\phi$ and $c_3\geq 64$, after $T$ rounds is at most
\begin{equation}\label{eq:regret_dependent_DAG}
    \log T\left(\frac{20c^2}{n+m}\sum_{i:\mu_i<\mu^*}\frac{\eta_i^2}{\Delta_i}+\frac{c_3\sum_{i=1}^K\Delta_i}{n+m}\right)+C_1\left(\sum_{j=1}^K\Delta_j\right),
\end{equation}
where $C_1=\ell_0+1+\frac{\pi^2}{3}\left(1+\exp\left(-\frac{n_0-s_0-2}{16}\right)\right) +\frac{36}{n}\exp\left(-\frac{n_0+n\ell_0}{18}\right)$.  
\end{theorem}

\begin{proof}
 The key idea of the proof follows the proof of Theorem 4 in \cite{auer2002MAB}. Denote the optimal intervention node as $a^*$, with parent set $\mathrm{pa}^*$. Let $c_t=c\sqrt{\log t}$, $n^*(t)$ be the number of times that the optimal intervention is picked, $\widehat\mu_u^*(t)$ be the estimated mean reward $\widehat\mu^*(t,\mathrm{pa}^*)$ of the optimal intervention when $n^*(t)=u$, and $\widehat\mu_{i,u_i}(t)$ be the estimated mean reward $\widehat\mu_i(t,\mathrm{pa}_i)$ of arm $i$ when $n_i(t)=u_i$. Let $\widehat\sigma_u^*(t)$ be the estimated standard error of $\widehat\mu^*(t)$ when $n^*(t)=u$ and $\widehat\sigma_{i,u_i}(t)$ be the estimated standard error of $\widehat\mu_i(t)$ when $n_i(t)=u_i$. Then for any $\ell\geq0$, we have
\begin{align*}
     &n_i(T) =\sum_{t=1}^{T}\mathbbm 1\left\{A_{t}=i\right\} \leq \ell+\sum_{t=\ell+1}^{T}\mathbbm 1\left\{A_{t}=i, n_i(t-1) \geq \ell\right\} \\ 
     & \leq \ell+\sum_{t=\ell}^{T-1}\mathbbm 1\{\widehat{\mu}^{*}(t)+\widehat{\sigma}^*(t)c_{t} \leq \widehat{\mu}_i(t)+\widehat{\sigma}_i(t)c_{t},n_i(t-1) \geq \ell\} \\ 
     & \leq \ell+\sum_{t=\ell}^{T-1}\mathbbm 1\left\{\min _{1\leq u\leq t} \widehat{\mu}_{u}^{*}(t)+\widehat\sigma_u^*(t) c_{t} \leq \max _{\ell \leq u_{i}\leq t} \widehat{\mu}_{i, u_{i}}(t)+\widehat\sigma_{i,u_i}(t) c_{t}\right\} \\ 
     & \leq \ell+\sum_{t=\ell}^{T-1} \sum_{u=1}^{t} \sum_{u_{i}=\ell}^{t}\mathbbm 1\left\{\widehat{\mu}_{u}^{*}(t)+\widehat\sigma_u^*(t) c_{t} \leq  \widehat{\mu}_{i, u_{i}}(t)+\widehat\sigma_{i,u_i}(t) c_{t}\right\} .
\end{align*}
When $\widehat{\mu}_{u}^{*}(t)+\widehat\sigma_u^*(t) c_{t} \leq  \widehat{\mu}_{i, u_{i}}(t)+\widehat\sigma_{i,u_i}(t) c_{t}$, i.e.,
\begin{align*}
    0&\leq \widehat{\mu}_{i, u_{i}}(t)+\widehat\sigma_{i,u_i}(t) c_{t}-\left(\widehat{\mu}_{u}^{*}(t)+\widehat\sigma_u^*(t) c_{t}\right)\\
    &=\left(\widehat{\mu}_{i, u_{i}}(t)-\mu_i-\widehat\sigma_{i,u_i}(t) c_{t}\right)+\left(\mu_i+2\widehat\sigma_{i,u_i}(t) c_{t}-\mu^*\right)+\left(\mu^*-\widehat{\mu}_{u}^{*}(t)-\widehat\sigma_u^*(t) c_{t}\right),
\end{align*}
at least one of the following three events must happen
\begin{align*}
    \widehat{\mu}_{i, u_{i}}(t)-\mu_i-\widehat\sigma_{i,u_i}(t) c_{t}&\geq 0,\\
    \mu_i+2\widehat\sigma_{i,u_i}(t) c_{t}-\mu^* &\geq 0,\\
    \mu^*-\widehat{\mu}_{u}^{*}(t)-\widehat\sigma_u^*(t) c_{t}&\geq 0.
\end{align*}
Therefore,
\begin{align*} 
n_i(T) \leq & \ell+\sum_{t=\ell}^{T-1} \sum_{u=1}^{t} \sum_{u_{i}=\ell}^{t}\Big[\mathbbm 1\left\{\widehat\mu_u^{*}(t) \leq \mu^{*}-\widehat\sigma_u^*(t)c_{t}\right\}\\
& +\mathbbm 1\left\{\widehat\mu_{i, u_{i}}(t) \geq \mu_{i}+\widehat\sigma_{i,u_i}(t)c_{t}\right\} +\mathbbm 1\left\{\mu^{*}\leq\mu_{i}+2 \widehat\sigma_{i,u_i}(t)c_{t}\right\}\Big],
\end{align*}
and
\begin{equation}\label{eq:proof_1}
    \begin{aligned} 
\mbe \left[n_i(T)\right] \leq & \ell+\sum_{t=\ell}^{T-1} \sum_{u=1}^{t} \sum_{u_{i}=\ell}^{t}\Big[\mbp\left(\widehat\mu_u^{*}(t) \leq \mu^{*}-\widehat\sigma_u^*(t)c_{t}\right)\\
& +\mbp\left(\widehat\mu_{i, u_{i}}(t) \geq \mu_{i}+\widehat\sigma_{i,u_i}(t)c_{t}\right) +\mbp \left(\mu^{*}\leq\mu_{i}+2 \widehat\sigma_{i,u_i}(t)c_{t}\right)\Big].
\end{aligned}
\end{equation}
In what follows, we bound the three probabilities in \eqref{eq:proof_1}, which provide an upper bound on $\mbe[n_i(T)]$ for $i\neq a^*$. Then this translates into a bound on $\mathcal R_T=\sum_i\Delta_i \mbe[n_i(T)]$. Let $s=|\mathrm{pa}(i)|$ and $s_0= \underset{i}{\max} |\mathrm{pa}(i)|$.

\textbf{(a) Bound on $\mbp \left(\mu^{*}\leq\mu_{i}+2 \widehat\sigma_{i,u_i}(t)c_{t}\right)$. }

Plugging in $c_t=c\sqrt{\log t}$, we have
\begin{align*}
    \mathbb P\left( \mu^{*}\leq\mu_{i}+2 \widehat\sigma_{i,u_i}(t)c_{t}\right)= \mathbb P\left(\widehat\sigma_{i,u_i}^2(t)\geq\frac{\Delta_i^2}{4c^2\log(t)}\right),
\end{align*}
where 
\begin{align*}
\widehat\sigma_{i,u_i}^2(t)=\frac{(n_0+nt)^2\widehat{\operatorname{Var}}(\hat{\mu}_{i,\mathrm{obs}}(t))+(mu_i)^2\widehat{\operatorname{Var}}(\hat{\mu}_{i,\mathrm{int}}(t))}{(n_0+nt+mu_i)^2},
\end{align*}
and $\widehat{\operatorname{Var}}(\hat{\mu}_{i,\mathrm{int}}(t))$ and $\widehat{\operatorname{Var}}(\hat{\mu}_{i,\mathrm{obs}}(t))$ are defined in Equations \ref*{eq:est_int} and \ref*{eq:est_obs} in the main paper respectively. From Gaussian sample variance,  
\begin{align*}
    \widehat{\operatorname{Var}}(\hat{\mu}_{i,\mathrm{int}}(t))=\frac{s_{i,t}^2}{N_i(t)}\sim \frac{\widetilde\sigma_{i}^2}{mu_i(mu_i-1)}\chi_{mu_i-1}^2.
\end{align*}
From Gaussian linear regression,
\begin{align*}
&   \widehat{\operatorname{Var}}(\hat{\mu}_{i,\mathrm{obs}}(t))=\frac{1}{n_0+nt} \rho_i(t,\mathrm{pa}(i))\widehat{\zeta}_{i}^2(t),\\
  & \widehat{\zeta}_{i}^2(t)=\left.\frac{\|\mathbf Y-\mathbf X_i\hat\beta_i-\mathbf X_{\mathrm{pa}(i)}\hat\beta_{\mathrm{pa}(i)}\|^2}{N_o(t)-s-1}\right|\mathbf X_{i,\mathrm{pa}(i)} \sim \frac{\zeta_{i}^2}{n_0+nt-s-1}\chi_{n_0+nt-s-1}^2.
\end{align*}
Given the independence between observational data and interventional data, we have
\begin{align*}
     \widehat\sigma_{i,u_i}^2(t)\mid\mathbf X_{i,\mathrm{pa}(i)}\sim \frac{\left[\frac{( n_0+nt)\rho_i(t,\mathrm{pa}(i))}{ n_0+nt-s-1}\zeta_{i}^2\chi_{ n_0+nt-s-1}^2+\frac{mu_i}{mu_i-1}\widetilde\sigma_{i}^2\chi_{mu_i-1}^2\right]}{(n_0+ nt+mu_i)^2}.
\end{align*}
Let $\ell\geq\ell_0=\left\lceil\frac{9\operatorname{tr}(\Sigma)}{n\gamma_{\min}(\Sigma)} \right\rceil$ and $\ell\geq 2$. Since $t\geq \ell$ and $u_i\geq \ell$, we have $\frac{n_0+nt}{n_0+nt-s-1}\leq 2$ and $\frac{mu_i}{mu_i-1}\leq 2$. Define $\eta_i^2 = \max\{\rho_i(t,\mathrm{pa}(i))\zeta_i^2,\widetilde\sigma_i^2\}$. Then we have
\begin{align*}
    \frac{( n_0+nt)\rho_i(t,\mathrm{pa}(i))}{ n_0+nt-s-1}\zeta_{i}^2\chi_{ n_0+nt-s-1}^2+\frac{mu_i}{mu_i-1}\widetilde\sigma_{i}^2\chi_{mu_i-1}^2\leq 2\eta_i^2\chi_{n_0+nt+mu_i-s-2}^2.
\end{align*}
Therefore,
\begin{align*}
    \mathbb P\left(\widehat\sigma_{i,u_i}^2(t)\geq\frac{\Delta_i^2}{4c^2\log t}\middle| \mathbf X_{i,\mathrm{pa}(i)}\right)
    \leq \mathbb P\left(\chi_{  n_0+nt+mu_i-s-2}^2\geq\frac{(n_0+nt+mu_i)^2\Delta_i^2}{8c^2\eta_i^2\log t}\right).
\end{align*}

According to Lemma~\ref{thm:chi_bound}, when $\frac{(n_0+nt+mu_i)^2\Delta_i^2}{8c^2\eta_i^2\log t}\geq \frac{5}{2}(n_0+nt+mu_i-s-2)$, i.e., $u_i\geq \ell\geq \frac{20c^2\eta_i^2\log t}{(m+n)\Delta_i^2}$, we have
\begin{align*}
   \mathbb P\left(\chi_{  n_0+nt+mu_i-s-2}^2\geq\frac{(n_0+nt+mu_i)^2\Delta_i^2}{8c^2\eta_i^2\log t}\right)&\leq \exp\left(-(n_0+nt+mu_i-s-2)/4\right)\\
  &\leq \exp\left(-(n_0-s-2+(m+n)u_i)/4\right).
\end{align*}
Let $u_i\geq \frac{16\log t}{n+m}$. Then,
\begin{equation*}
   \mathbb P\left(\chi_{  n_0+nt+mu_i-s-2}^2\geq\frac{(n_0+nt+mu_i)^2\Delta_i^2}{8c^2\eta_i^2\log t}\right)\leq t^{-4}\exp\left(-(n_0-s-2)/4\right).
\end{equation*}
Therefore, if $u_i\geq \frac{16\log t}{n+m}$ and $\ell\geq \max\big\{\frac{20c^2\eta_i^2\log t}{(m+n)\Delta_i^2}, 2,  \ell_0\big\}$, then for any $t\geq u_i\geq \ell$, we have
\begin{equation}\label{eq:bound_sigma2}
     \mathbb P\left( \mu^{*}\leq\mu_{i}+2 \widehat\sigma_{i,u_i}(t)c_{t}\middle| \mathbf X_{i,\mathrm{pa}(i)}\right)\leq t^{-4}\exp\left(-(n_0-s_0-2)/4\right).
\end{equation}

\textbf{(b) Bounds on $\mathbb P(\widehat\mu_{i, u_{i}}\geq \mu_{i}+\widehat\sigma_{i,u_i}c_{t})$ and $\mbp\left(\widehat\mu_u^{*}(t) \leq \mu^{*}-\widehat\sigma_u^*(t)c_{t}\right)$. }

Define $\phi_i^2=\min \{\rho_i(t,\mathrm{pa}(i))\zeta_i^2,\widetilde\sigma_i^2\}$. Then, 
\begin{equation}\label{eq:chi_ineq_proof}
     \frac{N_o(t)\rho_i(t,\mathrm{pa}(i))}{ N_o(t)-s-1}\zeta_{i}^2\chi_{ N_o(t)-s-1}^2+\frac{N_i(t)}{N_i(t)-1}\widetilde\sigma_{i}^2\chi_{N_i(t)-1}^2\geq \phi_i^2\chi_{N_o(t)+N_i(t)-s-2}^2.
\end{equation}
Below drop $t$ for notational brevity. We have 
\begin{align*}
&\widehat\mu_{i,u_i}=\left.\frac{N_o\widehat\mu_{i,\mathrm{obs}}+N_i\widehat\mu_{i,\mathrm{int}}}{N_o+N_i}\right| \mathbf X_{i,\mathrm{pa}(i)}\sim \mathcal N\left(\mu_i,\frac{N_i\tilde\sigma_i^2+N_o\zeta_{i}^2\rho_i(\mathrm{pa}(i))}{(N_i+N_o)^2}\right),\\
 &\left.\widehat\sigma_{i,u_i}^2\right|\mathbf X_{i,\mathrm{pa}(i)}\sim \frac{\left[\frac{N_o\rho_i(\mathrm{pa}(i))}{ N_o-s-1}\zeta_{i}^2\chi_{ N_o-s-1}^2+\frac{N_i}{N_i-1}\widetilde\sigma_{i}^2\chi_{N_i-1}^2\right]}{(N_o+N_i)^2}.
\end{align*}
Therefore,
\begin{align*}
    \left.\frac{\widehat\mu_{i, u_{i}}-\mu_{i}}{\widehat\sigma_{i,u_i}} \right| \mathbf X_{i,\mathrm{pa}(i)}\sim \frac{\mathcal N(0,1)\sqrt{N_i\tilde\sigma_i^2+N_o \zeta_{i}^2\rho_i(\mathrm{pa}(i))}}{\sqrt{\frac{N_o\rho_i(\mathrm{pa}(i))}{ N_o-s-1}\zeta_{i}^2\chi_{ N_o-s-1}^2+\frac{N_i}{N_i-1}\widetilde\sigma_{i}^2\chi_{N_i-1}^2}}.
\end{align*}
Recall that $\eta_i=\max\left\{\rho_i(\mathrm{pa}(i))\zeta_i^2,\widetilde\sigma_i^2\right\}$. Given \eqref{eq:chi_ineq_proof}, we have
\begin{align*}
   \mathbb P\left(\widehat\mu_{i, u_{i}}\geq \mu_{i}+\widehat\sigma_{i,u_i}c_{t}\middle|\mathbf X_{i,\mathrm{pa}(i)}\right) & = \mathbb P\left(\frac{\widehat\mu_{i, u_{i}}-\mu_{i}}{\widehat\sigma_{i,u_i}}\geq c_{t}\middle| \mathbf X_{i,\mathrm{pa}(i)}\right)\\
  &\leq \mathbb P\left(\frac{\eta_i\mathcal N(0,1)\sqrt{N_i+N_o}}{\sqrt{\phi_{i}^2\chi_{N_o+N_i-s-2}^2}}\geq c_{t}\middle| \mathbf X_{i,\mathrm{pa}(i)}\right)\\
  &= \mathbb P\left(t_{N_o+N_i-s-2}\geq \frac{c \phi_i\sqrt{\log t}}{\sqrt{2}\eta_i}\middle| \mathbf X_{i,\mathrm{pa}(i)}\right).
\end{align*}
According to Lemma~\ref{thm:t_bound}, we have
\begin{equation}\label{eq:muhat-t-bound}
    \begin{aligned}
    \mathbb P\left(\widehat\mu_{i, u_{i}}\geq \mu_{i}+\widehat\sigma_{i,u_i}c_{t}\middle| \mathbf X_{i,\mathrm{pa}(i)}\right)&\leq \exp\left(-\frac{c^2\phi_i^2\log t}{8\eta_i^2}\right)+\frac{1}{2}\exp\left(-\frac{n_0+nt+mu_i-s-2}{16}\right)\\
    &\leq t^{-c^2\phi_i^2/8\eta_i^2}+\frac{1}{2}\exp\left(-\frac{n_0+nt+mu_i-s-2}{16}\right).
\end{aligned}
\end{equation}
Let $c\geq 4\sqrt{2}\eta_i/\phi_i$. If $u_i\geq \frac{64\log t}{n+m}$, then 
\begin{equation}\label{eq:bound_mui}
    \mathbb P\left(\widehat\mu_{i, u_{i}}\geq \mu_{i}+\widehat\sigma_{i,u_i}c_{t}\middle| \mathbf X_{i,\mathrm{pa}(i)}\right)\leq t^{-4}\left(1+\frac{1}{2\exp((n_0-s_0-2)/16)}\right).
\end{equation}
% Set minimum intervention times as $\frac{64\log t}{n+m}$, for all arms, when $c\geq 4\sqrt{2}\eta^*/\phi^*$, we have 
% \begin{equation}\label{eq:bound_mustar}
%      \mbp\left(\widehat\mu_u^{*}(t) \leq \mu^{*}-\widehat\sigma_u^*(t)c_{t}\mid \mathbf X_{a^*,\mathrm{pa}^*}\right)\leq t^{-4}\left(1+\frac{1}{2\exp((n_0-s_0-2)/16)}\right).
% \end{equation}
Define $$\phi^2=\underset{i}{\max} \eta_i^2/\phi_i^2=\underset{i}{\max}\max\left\{\frac{\rho_i(t,\mathrm{pa}(i))\zeta_i^2}{\widetilde\sigma_i^2},\frac{\widetilde\sigma_i^2}{\rho_i(t,\mathrm{pa}(i))\zeta_i^2}\right\}.$$ Recall that $\overline V=\{1,2,\cdots, p-1\}$. Conditioning on $\mathbf X_{i,\mathrm{pa}(i)}$, $\rho_i(t,\mathrm{pa}(i))$ is a constant. Plugging \eqref{eq:bound_sigma2} and \eqref{eq:bound_mui} into \eqref{eq:proof_1}, then $\mbe [n_i(T)\mid \mathbf X_{\overline V}]$ can be bounded by 
\begin{equation}\label{eq:final_bound_0}
  \mbe \left[n_i(T)\middle| \mathbf X_{\overline V}\right]\leq \ell+ \sum_{t=\ell+1}^T t^{-2}\left(2+2\exp\left(-\frac{  n_0-s_0-2}{16}\right)\right),
\end{equation}
where
\begin{equation}\label{eq:min_int_1}
    \ell \geq \max\left\{\left\lceil\frac{20c^2\frac{\eta_i^2}{\Delta_i^2}\log t}{m+n}\right\rceil, \left \lceil \frac{c_3\log t}{n+m}\right\rceil,\left\lceil\frac{9\operatorname{tr}(\Sigma)}{n\gamma_{\min}(\Sigma)}\right\rceil, \ 2\right\},
\end{equation}
$c\geq 4\sqrt{2}\phi$, and $c_3\geq 64$. Note that choosing $c_3\geq64$ ensure that $n_i=n_i(t)\geq\frac{64\log t}{n+m}$ for all $i$ by Line~\ref*{alg:line:min_time_given} of Algorithm~\ref{alg:Combined-UCB}.

Averaging over $\mathbf X_{\overline V}$, $\rho_i(t,\mathrm{pa}(i))$ and $\rho^*(t,\mathrm{pa}^*)$ can be bounded by Lemma~\ref{thm:rho} in Section~\ref{appendix:rho_bounds}. In fact, for any $t\geq \left\lceil\frac{9\operatorname{tr}(\Sigma)}{n\gamma_{\min}(\Sigma)}\right\rceil=\ell_0$, we have
\begin{equation}\label{eq:rho_all}
    \begin{aligned}
 &\mathbb P\left(\rho_i(t,\pa(i))\geq\frac{9}{\gamma_{\min}(\Sigma) } \text{ or } \rho_i(t,\pa(i))\leq\frac{9}{25\gamma_{\max}(\Sigma) },\text{ for some } i\in\overline V\right)\\
    \leq & \mathbb P\left(\frac{\sigma_{\min}(\mathbf X)}{\sqrt{N_o(t)}}\leq\frac{1}{3}\gamma_{\min}(\sqrt{\Sigma}) \text{ or } \frac{\sigma_{\max}(\mathbf X)}{\sqrt{N_o(t)}}\geq\frac{5}{3}\gamma_{\max}(\sqrt{\Sigma})\right)\\
    \leq & 2 \exp\left(-\frac{n_0+nt}{18}\right).
 \end{aligned}
\end{equation}
 Therefore, for any $\ell\geq \left\lceil\frac{9\operatorname{tr}(\Sigma)}{n\gamma_{\min}(\Sigma)}\right\rceil=\ell_0$, we have
\begin{equation}\label{eq:rho_sum}
     \begin{aligned}
    &\sum_{t=\ell}^T \mathbb P\left(\rho_i(t,\pa(i))\geq\frac{9}{\gamma_{\min}(\Sigma) } \text{ or } \rho_i(t,\pa(i))\leq\frac{9}{25\gamma_{\max}(\Sigma) } ,\text{ for some } i\in\overline V\right)\\
    \leq & 2\sum_{t=\ell}^{\infty}\exp\left(-\frac{n_0+nt}{18}\right)\leq 2\int_{\ell}^{+\infty}\exp\left(-\frac{n_0+nt}{18}\right)dt\leq \frac{36}{n}\exp\left(-\frac{n_0+n\ell}{18}\right)
\end{aligned}
 \end{equation}
Now re-define 
\begin{align*}
   \eta_i^2&=\max\left\{ \frac{9\zeta_{i}^2}{\gamma_{\min}(\Sigma)},\tilde\sigma_i^2\right\},\\
   \phi^2 &= \max_i\max\left\{\frac{9\zeta_{i}^2}{\gamma_{\min}(\Sigma)\tilde\sigma_i^2},\frac{25\tilde\sigma_i^2\gamma_{\max}(\Sigma)}{9\zeta_{i}^2}\right\},
\end{align*}
and let $c\geq 4\sqrt{2}\phi$. Given the conditions on $\ell$~\eqref{eq:min_int_1}, the final upper bound of $\mathbb E[n_i(T)]$ is 
\begin{equation}\label{eq:bound_Ri}
    \begin{aligned}
    \mathbb E[n_i(T)]&\leq\ell  + \sum_{t=\ell}^T t^{-2}\left(2+2\exp\left(-\frac{  n_0-s_0-2}{16}\right)\right)+\eqref{eq:rho_sum}\\
    &\leq 20c^2\frac{\eta_i^2}{\Delta_i^2}\frac{\log T}{n+m}+\frac{c_3\log T}{n+m}+\frac{9\operatorname{tr}(\Sigma)}{n\gamma_{\min}(\Sigma)}+2+\frac{\pi^2}{3}\left(1+\exp\left(-\frac{n_0-s_0-2}{16}\right)\right)\\
    &\quad +\frac{36}{n}\exp\left(-\frac{n_0+n\ell_0}{18}\right),
\end{aligned}
\end{equation}
where $\ell_0=\left\lceil\frac{9\operatorname{tr}(\Sigma)}{n\gamma_{\min}(\Sigma)}\right\rceil$.

Then the upper bound of $\mathcal R_T$ can be expressed as
\begin{equation}\label{eq:upperbound_DAG_dep}
    \log T\left(\frac{20c^2}{n+m}\sum_{i:\mu_i<\mu^*}\frac{\eta_i^2}{\Delta_i}+\frac{c_3\sum_{i=1}^K\Delta_i}{n+m}\right)+C_1\left(\sum_{j=1}^K\Delta_j\right),
\end{equation}
where 
\begin{equation}\label{eq:C1}
        C_1=\ell_0+1+\frac{\pi^2}{3}\left(1+\exp\left(-\frac{n_0-s_0-2}{16}\right)\right) +\frac{36}{n}\exp\left(-\frac{n_0+n\ell_0}{18}\right).
\end{equation}
\end{proof}

\subsection{$\Delta$-independent bound}\label{sec:case_indep_DAG_given}

Assuming $\mu_i\in[-1,1]$ for all $i$, we restrict $U_i(t)\in[-1,1]$ for all $i$ and $t$ in Algorithm~\ref{alg:Combined-UCB}.

\begin{theorem}\label{thm:thm2_DAG}
  Assume $\mu_i\in[-1,1],\forall i\in\{1,\cdots,p-1\}$. Then under Assumptions~\ref*{ass:gaussian_dag}, \ref*{ass:id}, and \ref*{ass:n0}, the cumulative regret of Algorithm~\ref{alg:Combined-UCB}, with parameters $c\geq 4\phi$ and $c_3\geq 32$, after $T$ rounds is at most
\begin{equation}\label{eq:regret_indep_DAG}
\begin{aligned}
      &2c\eta\sqrt{\log T}\min\left\{\frac{\sqrt{(n+m)KT+K^2n_0}-\sqrt{K^2n_0}}{n+m},\frac{\sqrt{n_0+nT}-\sqrt{n_0}}{n}\mathbbm 1\{n\geq 1\}\right\}\\
      &\quad +\frac{2Kc_3\log T}{n+m}
        +C_2
\end{aligned}
\end{equation}
where $C_2 = 2\ell_0+\frac{\pi^2}{3}+2K+\frac{\pi^2}{6}\exp\left(-\frac{n_0-s_0-2}{16}\right) + \frac{72}{n}\exp\left(-\frac{n_0+n\ell_0}{18}\right)$.
\end{theorem}

\begin{proof} 
Let $\ell\geq K\left\lceil\frac{c_3\log T}{n+m}\right\rceil$. Then, for any $t\geq \ell+1$, the arm chosen $A_t$ is the arm with the largest upper confidence bound. Let $U_t(i)=\hat\mu_i(t,S)+\hat\sigma_i(t,S)c\sqrt{\log t}$ be the constructed upper confidence bound of arm $i$ after $t$ rounds. All symbols with superscript $*$ denote quantities for the optimal arm, such as $A^*$ and $\mu^*$ for the optimal arm and the largest expected reward, respectively. Since $\mu_i\in[-1,1]$ and $U_t(i)\in [-1,1]$ by construction, $|\mu_i-U_t(i)|\leq 2$, $\forall i, t$. Therefore,
\begin{equation}\label{eq:indep_inequality}
    \begin{aligned}
     \mathcal R_T&= \sum_{t=1}^T \mbe[\mu^*-\mu_{A_t}] \leq 2\ell+\sum_{t\geq \ell+1}\mbe[\mu^*-U_{t-1}(A_t)+U_{t-1}(A_t)-\mu_{A_t}]\\
    &\leq 2\ell+\sum_{t\geq \ell+1}\mbe[\mu^*-U_{t-1}(A^*)]+\sum_{t\geq \ell+1}\mbe[U_{t-1}(A_t)-\mu_{A_t}]\\
    &\leq 2\ell+2\sum_{t=\ell}^{T-1}\mbp \left(\mu^*>U_t(A^*)\right)+\sum_{t=\ell+1}^T\mbe[U_{t-1}(A_t)-\mu_{A_t}].
\end{aligned}
\end{equation}
According to Lemma~\ref{thm:rho}, \eqref{eq:rho_sum} holds for any  $S(i)\subseteq\mathcal V\setminus\{i\}$ in place of $\mathrm{pa}(i)$ and any $t\geq \ell_0=\left\lceil\frac{9\operatorname{tr}(\Sigma)}{n\gamma_{\min}(\Sigma)}\right\rceil$. 

When $c\geq 4\phi$, $c_3\geq 32$, and $\rho_i(t,\mathrm{pa}(i))$ is between $\frac{9}{25\gamma_{\max}(\Sigma)}$ and $\frac{9}{\gamma_{\min}(\Sigma)}$ conditioning on $\mathbf X_{i,\mathrm{pa}(i)}$,  we have 
\begin{align*}
    \mbp \left(\mu^*>U_t(A^*)\middle| \mathbf X_{a^*,\mathrm{pa}^*}\right)\leq t^{-4}\left(1+\frac{1}{2\exp((n_0-s_0-2)/16)}\right)
\end{align*}
from \eqref{eq:bound_mui}. %\qzedit{[This does not go to S3.9, check]}. 
Then 
\begin{equation}\label{eq:indep_b2}
    \sum_{t=K}^{T-1}\mbp \left(\mu^*>U_t(A^*)\middle| \mathbf X_{a^*,\mathrm{pa}^*}\right)\leq \frac{\pi^2}{6}\left(1+\frac{1}{2}\exp(-(n_0-s_0-2)/16)\right).
\end{equation}
Let $\eta^2=\max_i\max\left\{\frac{9\zeta_{i,\mathrm{pa}(i)}^2}{\gamma_{\min}(\Sigma)},\widetilde\sigma_i^2\right\}$,  $\ell\geq \ell_0=\left\lceil\frac{9\operatorname{tr}(\Sigma)}{n\gamma_{\min}(\Sigma)}\right\rceil$, and $\mathcal T_a=\{t:A_t=a, \ell +1\leq t\leq T\}$. Then 
\begin{align*}
    \sum_{t=\ell+1}^T\mbe\left(U_{t-1}(A_t)-\mu_{A_t}\right)
     \leq \sum_{a\in \mathcal A}\sum_{t\in \mathcal T_a}\mbe\left(U_{t-1}(a)-\mu_{a}\right).
\end{align*}
For $t\geq  \ell$ and $\forall a\in \mathcal A$,
\begin{align*}
    \mbe [U_t(a)-\mu_{a}|\mathbf X_{\overline V}]&=\mbe (\widehat\sigma_a|\mathbf X_{\overline V})c\sqrt{\log t}\leq c\sqrt{\log t}\sqrt{\mbe (\widehat\sigma_a^2|\mathbf X_{\overline V})}\\
    &=c\sqrt{\log t}\frac{\sqrt{N_o(t)\zeta_{i}^2\rho_i(t,\mathrm{pa}(i))+N_a(t)\widetilde\sigma_i^2}}{N_o(t)+N_a(t)}\leq c\eta\sqrt{\frac{\log t}{N_o(t)+N_a(t)}}.
\end{align*}
Since $N_o(t)+N_a(t)=n_0+nt+mn_a(t)\geq n_0+(n+m)n_a(t)$,
\begin{equation}\label{eq:indep_b3}
    \begin{aligned}
   \sum_{a\in \mathcal A}\sum_{t\in \mathcal T_a}\sqrt{\frac{\log t}{N_o(t)+N_a(t)}}&\leq \sum_{a\in \mathcal A}\sum_{t\in \mathcal T_a}\sqrt{\frac{\log t}{n_0+(n+m)n_a(t)}}\leq \sum_{a\in\mathcal A}\sum_{\nu=0}^{n_a(T)}\sqrt{\frac{\log T}{n_0+(n+m)\nu}}\\
    &\leq\sqrt{\log T}\sum_{a\in \mathcal A}\int_0^{n_a(T)}\frac{1}{\sqrt{n_0+(n+m)\nu}}d\nu\\
    &\leq \frac{2\sqrt{\log T}}{n+m}\sum_{a\in\mathcal A}\left(\sqrt{n_0+(m+n)n_a(T)}-\sqrt{n_0}\right).
\end{aligned}
\end{equation}
By Cauchy-Schwartz inequality, 
\begin{align*}
    \sum_{a\in\mathcal A}\sqrt{n_0+(n+m)n_a(T)}\leq \sqrt{K\left(\sum_{a\in\mathcal A}(n_0+(n+m)n_a(T))\right)}=\sqrt{(n+m)KT+K^2n_0}.
\end{align*}
Since $N_o(t)+N_a(t)\geq N_o(t)$, when $n\geq 1$, we also have 
\begin{equation}\label{eq:indep_b4}
    \begin{aligned}
    \sum_{t=\ell}^T\sqrt{\frac{\log t}{N_o(t)+N_a(t)}}&\leq  \sum_{t=\ell}^T\sqrt{\frac{\log t}{n_0+nt}}\leq \sqrt{\log T}\int_{0}^T\frac{1}{\sqrt{n_0+nt}}dt\\
    &=\frac{2\sqrt{\log T}}{n}\left(\sqrt{n_0+nT}-\sqrt{n_0}\right).
\end{aligned}
\end{equation}
Plugging the upper bounds on $\sum_{t=\ell+1}^T\mbe[U_{t-1}(A_t)-\mu_{A_t}]$ and $\sum_{t=\ell}^{T-1}\mbp \left(\mu^*>U_t(A^*)\right)$ into \eqref{eq:indep_inequality},
\begin{equation*}
    \begin{aligned}
        \mathcal R_T \leq &2\ell+c\eta\min\left\{\eqref{eq:indep_b3},\eqref{eq:indep_b4}\right\}+2\times \eqref{eq:rho_sum} + 2\times \eqref{eq:indep_b2}\\
        \leq &2\ell+2c\eta\sqrt{\log T}\min\left\{\frac{\sqrt{(n+m)KT+K^2n_0}-\sqrt{K^2n_0}}{n+m},\frac{\sqrt{n_0+nT}-\sqrt{n_0}}{n}\right\} \\
        & +\frac{\pi^2}{6}\left(2+\exp\left(-\frac{n_0-s_0-2}{16}\right)\right)+\frac{72}{n}\exp\left(-\frac{n_0+n\ell}{18}\right).
    \end{aligned}
\end{equation*}
Since $\ell\geq K\lceil\frac{c_3\log T}{n+m}\rceil$ and $\ell\geq \ell_0$,
\begin{equation}\label{eq:upperbound_DAG_indep}
    \begin{aligned}
        \mathcal R_T &\leq 2c\eta\sqrt{\log T}\min\left\{\frac{\sqrt{(n+m)KT+K^2n_0}-\sqrt{K^2n_0}}{n+m},\frac{\sqrt{n_0+nT}-\sqrt{n_0}}{n}\mathbbm 1\{n\geq 1\}\right\}\\
        &\quad +\frac{2Kc_3\log T}{n+m}
        +C_2,
    \end{aligned}
\end{equation}
where
\begin{equation}\label{eq:C2}
    C_2 = 2\ell_0+\frac{\pi^2}{3}+2K+\frac{\pi^2}{6}\exp\left(-\frac{n_0-s_0-2}{16}\right) + \exp\left(-\frac{n_0+n\ell_0}{18}\right)\frac{72}{n}.
\end{equation}
\end{proof}

\section{Proof of main results in Section~\ref*{sec:regret}}\label{appendix:DAGunknown}

% \begin{algorithm}[h]
% \caption{BA-UCB for Gaussian Causal Bandits}\label{alg:BA-UCB_appendix}
% \begin{algorithmic}
% \Require Observational data $\mathbf D_0$ of size $n_0$, parameters  $c$, $c_3$, $n$, $m$, rounds $T$.
% \For{$t=1,\cdots, T$}
% \If{There is a node $i$ which has been intervened less than $\lceil\frac{c_3\log t+1}{m} \rceil$ times}
%     \State Choose intervention node $A_{t} = i$;
% \Else
% \For {$i=1,\cdots, p-1$}
%   \State Estimate the regression neighborhood $\widehat M_i$ of each node $X_i$ and find $\widehat{\mathcal N}_i$;
%   \State Identify candidate backdoor adjustment sets $\widehat{\mathcal B}_i = \mathrm{argmin}_{S \in \widehat{\mathcal{N}}_i} d(0, CI_i(t-1,S))$;
%   \State Sample set $S$ from $\widehat{\mathcal B}_i$, calculate $\hat{\mu}_{i,\mathrm{obs}}(t-1,S)$ and $\hat\sigma_{i,\mathrm{obs}}(t-1,S)$;
%    \State Calculate estimates $\hat\mu_i(t-1,S)$ and $\hat\sigma_i^2(t-1,S)$;
% \EndFor
%     \State Choose $A_{t}=\underset{i \in \{1, \ldots, p-1\}}{\operatorname{argmax}}\left\{\hat{\mu}_{i}(t-1,S)+\hat{\sigma}_{i}(t-1,S) c \sqrt{\log (t-1)}\right\}$; 
% \EndIf
% \State Sample $m$ data points under intervention $A_{t}$ and $n$ observational data.
% \EndFor
% \Ensure Optimal intervention $A_T$ and estimated reward $\hat \mu^*$.
% \end{algorithmic}
% \end{algorithm} 

When the underlying DAG is unknown, we take an additional step to estimate the backdoor adjustment sets. In this section, we modify the proof for the DAG given setting to establish our main theoretical results.  

We adopt a sample-splitting strategy to decouple adjustment set identification from parameter estimation used to construct upper confidence bounds and select arms. Let $M_i$ denote the regression neighborhood of a node $X_i$, $\mathcal N_i$ denote the collection of all subsets of $M_i$, $\mathcal B_i$ denote the collection of valid adjustment sets for $(X_i, Y)$, and $\widetilde{\mathbf X}$ denote the observational data used exclusively for identifying adjustment sets. Also, denote the sampled set $S$ in Algorithm~\ref*{alg:BA-UCB} Line~\ref*{alg:line:sample} at time step $t$ as $S(i,t)$ for node $X_i$ and $S^*(t)$ for the optimal arm.

For any $S\in \mathcal N_i$, we have
\begin{equation}\label{eq:dist_sample_splitting}
    \begin{aligned}
    \widehat{\mu}_i(t,S)-\widehat{\mu}_{i,\mathrm{int}}(t)\mid \widetilde{\mathbf X}_{i,S}& \sim \mathcal N\left( \mu_i(S)-\mu_i, \frac{\widetilde \sigma_i^2}{N_i(t)} + \frac{\zeta_{i,S}^2\rho_i(t,S)}{N_o(t)}\right),\\
\widehat{\operatorname{Var}}\left(\widehat{\mu}_i(t,S)-\widehat{\mu}_{i,\mathrm{int}}(t)\right) &= \left.\frac{\widehat\zeta_{i,S}^2\rho_i(t,S)}{N_o(t)}+ \frac{s_{i,t}^2}{N_i(t)}\right| \widetilde{\mathbf X}_{i,S}\\
&\sim \frac{\zeta_{i,S}^2\rho_i(t,S)}{N_o(t)(N_o(t)-s-1)}\chi_{N_o(t)-s-1}^2+ \frac{\widetilde\sigma_i^2}{N_i(t)(N_i(t)-1)}\chi_{N_i(t)-1}^2.
\end{aligned}
\end{equation}
As displayed in the main paper's Equation~\ref*{eq:CI}, the confidence intervals $CI_i(t,S)$ at round t for $\mu_i(S)-\mu_i$ are
\begin{equation}\label{eq:CI_constructed}
     \widehat\mu_{i,\mathrm{obs}}(t,S) - \widehat\mu_{i,\mathrm{int}}(t) \pm c_2\sqrt{\log t}\sqrt{ \widehat{\operatorname{Var}}(\hat{\mu}_{i,\mathrm{obs}}(t,S))+ \widehat{\operatorname{Var}}(\hat{\mu}_{i,\mathrm{int}}(t))}.
\end{equation}

We now prove Theorem~\ref*{thm1} and Theorem~\ref*{thm2}.

\subsection{Proof of Theorem~\ref*{thm1}} \label{sec:case_dep_DAG_unknown}
% \textit{Proof.} \qzcmt{see below and change other similar places}
We first introduce and prove some lemmas that form the essential building blocks of the proof of Theorem~\ref*{thm1}.

\begin{lemma}\label{thm:CI_pa}
    Under the same setting of Theorem~\ref*{thm1} and Assumptions~\ref*{ass:gaussian_dag} and \ref*{ass:n0},  with parameters $c_2=2\sqrt{3}$ and $c_3\geq 18$, for any $i\in\overline V$ and $t$,
    \begin{equation}\label{eq:dep_dag_b1}
    \mathbb P\left(0\notin CI_i(t,\mathrm{pa}(i))\middle|\widetilde{\mathbf X}_{i,\pa(i)}\right)\leq 3t^{-2}+\exp\left( -\frac{N_o(t)-s_0-1}{9}\right).
\end{equation}
\end{lemma}
\begin{proof}
    We drop the time index $t$ for simplicity in notation. We have
    \begin{equation*}
    \mathbb P\left(0\notin CI_i(\mathrm{pa}(i))\middle|\widetilde{\mathbf X}_{i,\pa(i)}\right)=\mathbb P\left(\left|\frac{\widehat{\mu}_i(\mathrm{pa}(i))-\widehat{\mu}_{i,\mathrm{int}}}{\sqrt{\widehat{\operatorname{Var}}\left(\widehat{\mu}_i(\mathrm{pa}(i))-\widehat{\mu}_{i,\mathrm{int}}\right)}}\right|>c_2\sqrt{\log t}\middle|\widetilde{\mathbf X}_{i,\pa(i)}\right).
\end{equation*}
Let $V_i=\frac{\zeta_{i,\pa(i)}^2\rho_i(\pa(i))}{N_0}+\frac{\widetilde{\sigma}_i^2}{N_i}$,
$$Z_i=\left.\frac{\widehat{\mu}_i(\pa(i))-\widehat{\mu}_{i,\mathrm{int}}}{\sqrt{V_i}}\right| \widetilde{\mathbf X}_{i,\pa(i)}\sim \mathcal N(0,1).$$ Let $\widehat V_i=\widehat{\operatorname{Var}}\left(\widehat{\mu}_i(\mathrm{pa}(i))-\widehat{\mu}_{i,\mathrm{int}}\right)$. Then,
\begin{align*}
    \frac{Z_i}{\sqrt{\widehat{V}_i/V_i}}>c_2\sqrt{\log t}\Rightarrow \sqrt{3}|Z_i|>c_2\sqrt{\log t} \text{ or }\widehat{V}_i<V_i/3,
\end{align*}
and we have
\begin{align*}
    \mathbb P\left(0\notin CI_i(\mathrm{pa}(i))\middle|\widetilde{\mathbf X}_{i,\pa(i)}\right)
    &\leq \mbp\left(\left |\mathcal N(0,1)\right |>c_2\sqrt{\log t/3}\right)\\
    &\  +\mbp \left(\widehat{\operatorname{Var}}\left(\widehat{\mu}_i(\pa(i))-\widehat{\mu}_{i,\mathrm{int}}\right) \leq \frac{\zeta_{i,\pa(i)}^2\rho_i(\pa(i))}{3N_o}+\frac{\widetilde \sigma_i^2}{3N_i}\middle| \widetilde{\mathbf X}_{i,\pa(i)}\right).
\end{align*}
According to Lemma~\ref{thm:chi_bound} and \eqref{eq:dist_sample_splitting}, for any $S\in\mathcal N_i$,
\begin{align*}
&\mbp \left(\widehat{\operatorname{Var}}\left(\widehat{\mu}_i(S)-\widehat{\mu}_{i,\mathrm{int}}\right) \leq \frac{\zeta_{i,S}^2\rho_i(S)}{3N_o}+\frac{\widetilde \sigma_i^2}{3N_i}\middle| \widetilde{\mathbf X}_{i,S}\right)\\
\leq \ &\mbp\left(\chi_{N_o-s-1}^2\leq \frac{N_o-s-1}{3}\right)+\mbp\left(\chi_{N_i-1}^2\leq \frac{N_i-1}{3}\right)\\
\leq \ &\exp\left(-\frac{N_o-s-1}{9}\right) + \exp\left(-\frac{N_i-1}{9}\right).
\end{align*}
Combining it with 
\begin{align*}
    \mbp\left(\left |\mathcal N(0,1)\right |>c_2\sqrt{\log t/3}\right)\leq2\exp\left(-\frac{c_2^2\log t}{6}\right),
\end{align*}
we have
\begin{align*}
    \mathbb P\left(0\notin CI_i(\mathrm{pa}(i))\middle| \widetilde{\mathbf X}_{i,\pa(i)}\right)
    &\leq  2\exp\left(-\frac{c_2^2\log t}{6}\right)+\exp\left(-\frac{N_o-s-1}{9}\right) +\exp\left(-\frac{N_i-1}{9}\right)\\
    & \leq 2t^{-c_2^2/6} +\exp\left( -\frac{n_0+nt-s_0-1}{9}\right) + t^{-2},
\end{align*}
where we used $N_i\geq c_3\log t+1\geq 18\log t+1$ and $s=|\mathrm{pa}(i)|\leq s_0$.

Let $c_2= 2\sqrt{3}$. Then
\begin{equation*}
    \mathbb P\left(0\notin CI(\mathrm{pa}(i))\middle|\widetilde{\mathbf X}_{i,\pa(i)}\right)\leq 3t^{-2}+\exp\left( -\frac{n_0+nt-s_0-1}{9}\right).
\end{equation*}
\end{proof}

Define constants $\psi^2=\max_i\max_{S'\in \mathcal N_i\setminus\mathcal B_i}\max\left\{\rho_i(t,S')\zeta_{i,S'}^2,\widetilde\sigma_i^2\right\}$, $\ell_0=\lceil\frac{9\operatorname{tr}(\Sigma)}{n\gamma_{\min}(\Sigma)}\rceil$, and $s_2=\max_i|\mathcal N_i\setminus\mathcal B_i|$.
\begin{lemma}\label{thm:CI_S}
    Under the same settings of Theorem~\ref*{thm1} and Assumptions~\ref*{ass:gaussian_dag}, \ref*{ass:id}, and \ref*{ass:n0}, with parameters $c_2=2\sqrt{3}$ and $c_3\geq \max\{32\psi^2/\delta^2,8\}$, for any $i\in\overline V$ and $t\geq(c_3/n)^2$,
   \begin{equation}\label{eq:dep_dag_b2}
    \begin{aligned}
    \sum_{S'\in\mathcal N_i\setminus\mathcal B_i}\mbp \left(d(0,CI_i(t,S')) = 0\middle|\widetilde{\mathbf X}_{i,S'}\right)
    \leq  s_2\left(3t^{-2}+ \exp\left(-\frac{N_o(t)-p}4\right)\right).
\end{aligned}
\end{equation}
\end{lemma}
\begin{proof}
    We drop the time index $t$ for simplicity in notation. We have
    \begin{align*}
    \mathbb P\left(d(0,CI_i(S')) = 0\middle| \widetilde{\mathbf X}_{i,S'}\right)= \mathbb P\left(\left|\frac{\widehat{\mu}_i(S')-\widehat{\mu}_{i,\mathrm{int}}}{\sqrt{\widehat{\operatorname{Var}}\left(\widehat{\mu}_i(S')-\widehat{\mu}_{i,\mathrm{int}}\right)}}\right|\leq c_2\sqrt{\log t}\middle| \widetilde{\mathbf X}_{i,S'}\right).
\end{align*}
Let $V_i(S')=\frac{\zeta_{i,S'}^2\rho_i(S')}{N_0}+\frac{\widetilde{\sigma}_i^2}{N_i}$,
$Z_i(S')=\frac{\widehat{\mu}_i(S')-\widehat{\mu}_{i,\mathrm{int}}}{\sqrt{V_i(S')}}$, and $\widehat V_i(S')=\widehat{\operatorname{Var}}\left(\widehat{\mu}_i(S')-\widehat{\mu}_{i,\mathrm{int}}\right)$. Then,
\begin{align*}
    \frac{Z_i(S')}{\sqrt{\widehat{V}_i(S')/V_i(S')}}\leq c_2\sqrt{\log t}\Rightarrow |Z_i(S')|\leq c_2\sqrt{3\log t} \text{ or }\widehat{V}_i(S')\geq 3V_i(S'),
\end{align*}
and we have
\begin{align*}
    \mathbb P\left(d(0,CI_i(S')) = 0\middle| \widetilde{\mathbf X}_{i,S'}\right)&\leq \mathbb P\left(\widehat{\operatorname{Var}}\left(\widehat{\mu}_i(S')-\widehat{\mu}_{i,\mathrm{int}}\right)\geq 3\left(\frac{\zeta_{i,S'}^2\rho_i(S')}{N_o}+ \frac{\widetilde\sigma_i^2}{N_i}\right)\middle| \widetilde{\mathbf X}_{i,S'}\right) \\
    &\quad + \mbp\left(\left |Z_i(S')\right|>c_2\sqrt{3\log t}\middle| \widetilde{\mathbf X}_{i,S'}\right).
\end{align*}
According to Lemma~\ref{thm:chi_bound} and \eqref{eq:dist_sample_splitting}, 
\begin{align*}
    \mathbb P\left(\widehat{\operatorname{Var}}\left(\widehat{\mu}_i(S')-\widehat{\mu}_{i,\mathrm{int}}\right)\geq \frac{3\zeta_{i,S'}^2\rho_i(S')}{N_o}+ \frac{3\widetilde\sigma_i^2}{N_i}\middle| \widetilde{\mathbf X}_{i,S'}\right)
   & \leq \exp\left(-\frac{N_o-s-1}4\right) + \exp\left(-\frac{N_i-1}4\right)\\
    & \leq \exp\left(-\frac{N_o-p}4\right) + t^{-2},
\end{align*}
where $s=|S'|$ and $N_i\geq c_3\log t+1\geq 8\log t +1$.

Note that
\begin{align*}
    \widehat{\mu}_i(S')-\widehat{\mu}_{i,\mathrm{int}}\mid \widetilde{\mathbf X}_{i,S'}\sim \mathcal N\left(\mu_i(S')-\mu_i, V_i(S')\right).
\end{align*}
According to \eqref{eq:dist_sample_splitting} and the identifiability assumption $|\mu_i(S')-\mu_i|\geq 2\delta$ for all $S'\in\mathcal N_i\setminus\mathcal B_i$, with $c_2=2\sqrt{3}$, we have
\begin{align*}
   \mbp\left(\left |Z_i(S')\right|>c_2\sqrt{3\log t}\middle| \widetilde{\mathbf X}_{i,S'}\right)&= \mathbb P\left(\left|\widehat{\mu}_i(S')-\widehat{\mu}_{i,\mathrm{int}}\right|\leq c_2\sqrt{3\log t V_i(S')}\middle| \widetilde{\mathbf X}_{i,S'}\right)\\
    &=  \mathbb P\left(\left|\mathcal N\left( \mu_i(S')-\mu_i, V_i(S')\right)\right|\leq 6\sqrt{\log t V_i(S')}\middle| \widetilde{\mathbf X}_{i,S'}\right)\\
     &\leq \mathbb P\left(\left|\mu_i(S')-\mu_i\right|-\left|\mathcal N\left( 0, V_i(S')\right)\right|\leq 6\sqrt{\log tV_i(S')}\middle|\widetilde{\mathbf X}_{i,S'}\right)\\
     &\leq  \mathbb P\left(\left|\mathcal N\left( 0, V_i(S')\right)\right|\geq 2\delta -6\sqrt{\log tV_i(S')}\middle| \widetilde{\mathbf X}_{i,S'} \right)\\
     &\leq  2\exp\left( -\frac{\left( 2\delta -6\sqrt{\log tV_i(S')}\right)^2}{2V_i(S')}\right).
\end{align*}
Since $t\geq (c_3/n)^2$, we have $\log t/(n_0+nt)\leq 1/c_3$. Given $N_o=n_0+nt$, $N_i\geq c_3\log t+1$, and $c_3\geq 32\psi^2/\delta^2$, we get
\begin{align*}
   \sqrt{\log t\left( \frac{\zeta_{i,S'}^2\rho_i(S')}{N_o}+\frac{\widetilde\sigma_i^2}{N_i}\right)}\leq \sqrt{\frac{\zeta_{i,S'}^2\rho_i(S')+\widetilde\sigma_i^2}{c_3}}\leq \sqrt{\frac{2\psi^2}{c_3}}\leq \frac{\delta}{4}.
\end{align*}
Therefore, according to Lemma~\ref{thm:rho},
\begin{align*}
   \mbp\left(\left |Z_i(S')\right|>c_2\sqrt{3\log t}\middle| \widetilde{\mathbf X}_{i,S'}\right)
    \leq  2\exp\left( -\frac{\left( 2\sqrt{\log tV_i(S')}\right)^2}{2V_i(S')} \right)\leq 2t^{-2}.
\end{align*}
Combining the upper bounds for the two parts, we have
\begin{equation*}
    \begin{aligned}
    \sum_{S'\in\mathcal N_i\setminus\mathcal B_i}\mbp \left(d(0,CI_i(t,S')) = 0\middle| \widetilde{\mathbf X}_{i,S'}\right)\leq s_2\left(3t^{-2}+\exp\left(-\frac{N_o(t)-p}4\right)\right).
\end{aligned}
\end{equation*}
\end{proof}

By assumption, $X_i = \sum_{j:j\neq i}\bar\beta_{ij}X_j+\varepsilon_i$, where $\varepsilon\sim \mathcal N(0,\bar{\sigma}_i^2)$. Define constants $s_1=\max_i|M_i|$ and $\bar\beta_i=\min_{j:\bar\beta_{ij}\neq 0}|\bar\beta_{ij}|$. Denote $\rho_{ij}(t)=[(\widetilde{\mathbf X}_{-i}^\top \widetilde{\mathbf X}_{-i}/N_o(t))^{-1}]_{j,j}$. The regression neighborhood of $X_i$ at time $t$ is determined by 
\[
\widehat{M}_i(t)=\{j \neq i:|\widehat{\beta}_{i j}(t)|/v_{ij}>\bar\alpha\sqrt{N_o(t)}\},
\]
where $\bar\alpha$ is a constant, $\widehat{\beta}_{i j}$ is the fitted coefficient for $X_j$ in the linear regression $X_i\sim X_{-i}$ using $\widetilde{\mathcal D}_o(t)$, and $v_{ij}^2=\widehat{\operatorname{Var}}(\widehat{\beta}_{i j}(t))$ is the estimated variance of $\widehat{\beta}_{i j}(t)$ from the standard results in Gaussian linear regression.
%$=\frac{RSS}{N_o(t)-p}\frac{\rho_{ij}(t)}{N_o(t)}$ from the standard results in Gaussian linear regression.}

\begin{lemma}\label{thm:regression_neighbor}
    Suppose Assumptions~\ref*{ass:gaussian_dag} and \ref*{ass:n0} hold. Consider Algorithm~\ref*{alg:BA-UCB} with the same parameters as in Theorem~\ref*{thm1}. If $\bar\alpha\leq \min_{i,j}\frac{\bar\beta_i}{2\sqrt{3{\bar\sigma}_i^2\rho_{ij}(t)}}$ and $N_o(t)\geq p$, then for any $i\in\overline V$,
    \begin{equation}\label{eq:dep_dag_b3}
    \begin{aligned}
         \mathbb P\left(\widehat M_i(t)\neq M_i\middle|\widetilde{\mathbf X}_{-i}\right)
    &\leq (p-1)\left[2\exp\left( - \frac{\bar\alpha^2 N_o(t)}{4}\right)+\exp\left(-\frac{N_o(t)-p}{16}\right)\right]\\
    &\quad +s_1\exp\left(-\frac{N_o(t)-p}{4}\right).
    \end{aligned}
    \end{equation}
\end{lemma}
\begin{proof}
    
    Write the linear regression in matrix form: $\widehat {\bar{\beta}}_i=\operatorname{argmin}_{\beta_i}\|\widetilde X_i -\widetilde{\mathbf X}_{-i}\beta_i\|$. We drop the time index $t$ in notation for simplicity. From standard results in Gaussian linear regression, 
    \begin{align*}
    &\widehat{\bar\beta}_{ij}\mid\widetilde{\mathbf X}_{-i}\sim \mathcal N\left(\bar\beta_{ij}, \bar\sigma_i^2 \frac{\rho_{ij}}{N_o}\right),\\
    &\widehat{\bar\sigma}_i^2 =\left.\frac{\|\mathbf X_i-\mathbf X_{-i}\widehat\beta_i\|^2}{N_o-p}\right|\widetilde{\mathbf X}_{-i}\sim \bar\sigma_i^2\frac{\chi_{N_o-p}^2}{N_o-p},\\
    &\frac{\widehat{\bar\beta}_{ij}-\bar\beta_{ij}}{\sqrt{\widehat{\operatorname{Var}}(\widehat{\beta}_{i j})}}=\left.\frac{\widehat{\bar\beta}_{ij}-\bar\beta_{ij}}{\sqrt{\widehat{\bar\sigma}_i^2\frac{\rho_{ij}}{N_o}}}\right|\widetilde{\mathbf X}_{-i}\sim t_{N_o-p}.
\end{align*}
    
 Then $\forall j$ with $\bar{\beta}_{ij}\neq 0$,
\begin{align*}
    \mathbb P\left(X_j\notin \widehat{M}_i\middle|\widetilde{\mathbf X}_{-i}\right) &= \mathbb P\left(\left|\frac{\bar\beta_{ij}}{\sqrt{\widehat{\bar\sigma}_i^2\frac{\rho_{ij}}{N_o}}}+ t_{N_o-p}\right|\leq \bar\alpha\sqrt{N_o}\middle|\widetilde{\mathbf X}_{-i}\right)\\
    &\leq \mathbb P\left(|t_{N_o-p}|\geq \left(\frac{\bar\beta_i}{\sqrt{\widehat{\bar\sigma}_i^2\rho_{ij}}}-\bar\alpha\right)\sqrt{N_o}\middle|\widetilde{\mathbf X}_{-i}\right).
\end{align*}
Let $\bar\alpha\leq \frac{\bar\beta_i}{2\sqrt{3{\bar\sigma}_i^2\rho_{ij}}}$. According to Lemma~\ref{thm:chi_bound} and \ref{thm:t_bound},
\begin{align*}
    &\mathbb P\left(X_j\notin \widehat{M}_i\middle|\widetilde{\mathbf X}_{-i}\right)\\ \leq &\mathbb P\left(|t_{N_o-p}|\geq \left(\frac{\bar\beta_i}{\sqrt{\widehat{\bar\sigma}_i^2\rho_{ij}}}-\bar\alpha\right)\sqrt{N_o}\middle|\widetilde{\mathbf X}_{-i}\right)\\
    \leq &\mathbb P\left(|t_{N_o-p}|\geq \left(\frac{\bar\beta_i}{\sqrt{3{\bar\sigma}_i^2\rho_{ij}}}-\bar\alpha\right)\sqrt{N_o}\right)+\mbp\left(\widehat{\bar\sigma}_i^2\geq 3\bar\sigma_i^2\middle|\widetilde{\mathbf X}_{-i}\right)\\
    \leq & 2\exp\left(-\frac{\bar\beta_i^2N_o}{48{\bar\sigma}_i^2\rho_{ij}}\right)+\exp\left(-\frac{N_o(t)-p}{16}\right)+\mathbb{P}\left(\chi_{N_o-p}^2>3\left(N_o-p\right)\right)\\
    \leq & 2\exp\left(-\frac{\bar\alpha^2N_o}{4}\right)+\exp\left(-\frac{N_o(t)-p}{16}\right)+\exp\left(-\frac{N_o(t)-p}{4}\right).
\end{align*}
Similarly, $\forall j$ with $\bar\beta_{ij}=0$, 
\begin{align*}
    \mathbb P\left(X_j\in \widehat{M}_i\middle|\widetilde{\mathbf X}_{-i}\right)=\mathbb P\left( |t_{N_o-p}|\geq \bar\alpha \sqrt{N_o}\right)\leq 2\exp\left( - \frac{\bar\alpha^2 N_o}{4}\right)+\exp\left(-\frac{N_o-p}{16}\right).
\end{align*}
Therefore,
\begin{align*}
    \mathbb P\left(\widehat M_i\neq M_i\middle|\widetilde{\mathbf X}_{-i}\right)&\leq \sum_{j:\bar\beta_{ij}\neq 0} \mathbb P\left(X_j\notin \widehat{M}_i\middle|\widetilde{\mathbf X}_{-i}\right) + \sum_{j:\bar\beta_{ij}= 0} \mathbb P\left(X_j\in \widehat{M}_i\middle|\widetilde{\mathbf X}_{-i}\right)\\
    &\leq (p-1)\left[2\exp\left( - \frac{\bar\alpha^2 N_o}{4}\right)+\exp\left(-\frac{N_o-p}{16}\right)\right]\\
    &\quad +s_1\exp\left(-\frac{N_o(t)-p}{4}\right).
\end{align*}
\end{proof}

\begin{remark}
    Under our Gaussian DAG model,
    \begin{equation}\label{eq:cond_dist}
        X_i| X_{-i}\sim \mathcal N(\bar\mu_i,\bar\sigma_i^2), \text{ where } \bar\mu_i=\sum_{j:j\neq i}\bar\beta_{ij}X_j.
    \end{equation}
If $N_o(t)<p$, we can consider using Lasso-based neighborhood regression \citep{wainwright2019high} instead, i.e., 
\begin{equation*}
\widehat{\bar{\beta}}_i=\operatorname{argmin}_{\beta_i\in\mathbb R^{p-1}}\left\{\frac{1}{2N_o(t)}\| X_i -\mathbf X_{-i}{\beta}_i\|^2+\lambda_{N_o}\|{\beta}_i\|_1\right\}.
\end{equation*}
Consider the Gaussian graphical model constructed with the edge set $E=\{(i,j):\bar\beta_{ij}\neq 0\}$. %associated with the conditional distributions in \eqref{eq:cond_dist}, where an edge $(i,j)$ is present if and only if $\bar\beta_{ij}\neq 0$. 
According to Theorem 11.2 in \cite{wainwright2019high}, if the nonzero coefficients $\bar\beta_{ij}$ are bounded below in absolute value, then the estimated edge set $\widehat E=\{(i,j):\widehat{\bar\beta}_{ij}\neq 0\}$ equals the true edge set $E$ with high probability, provided that $N_o(t)\gg 2s_0\log p$ and $\lambda_{N_o}\asymp \sqrt{{\log p}/{N_o(t)}}$. More precisely, $\mathbb P(\widehat E=E)\geq 1-c_4\exp\left(-c_5N_o(t)\right)$, where $c_4$ and $c_5\leq 1$ are positive constants. Since each neighborhood $M_i$ corresponds to the set of neighbors of node $i$, this implies
\[
\mathbb P\Big(\bigcup_{i\in[p]}\{\widehat M_i \neq M_i\}\Big)
\le c_4 \exp\left(-c_5 N_o(t)\right).
\]
Therefore, using Lasso-based neighborhood regression, we can replace the union of the upper bound in Lemma~\ref{thm:regression_neighbor} by the above bound.
\end{remark}

% According to Theorem 11.2 in \cite{wainwright2019high}, if the non-zero coefficients $\bar\beta_{ij}$ in \eqref{eq:cond_dist} are bounded below in absolute value, then the estimated edge set $\hat E=\{(i,j):\widehat{\bar\beta}_{ij}\neq 0\}$ equals the actual edge set $E=\{(i,j):\bar\beta_{ij}\neq 0\}$ with high probability, assuming $N_o(t)\gg 2s_0\log p$ and $\lambda_{N_o}\asymp \sqrt{\frac{\log p}{N_o(t)}}$. More specifically, $\mathbb P(\hat E=E)\geq 1-c_4\exp\left(-c_5N_o(t)\right)$, where $c_4$ and $c_5\leq 1$ are constants. \qzedit{[QZ: (1) We didn't mention GGM, so the meaning of the edge sets here is not clear. (2) Based on this result of GGM, the upper bound is applicable to the union of all $M_i$ over $i\in[p]$ ]} Therefore, if we use Lasso based neighborhood regression, we have 
% \begin{equation*}
%     \mathbb P\left(\widehat M_i\neq M_i\right)\leq c_4\exp(-c_5N_o(t)).
% \end{equation*}

\newpage
Now we prove Theorem~\ref*{thm1}.
\begin{proof}
Let $S(i,t)$ denote the backdoor adjustment set sampled from $\widehat{\mathcal B}_i$ for node $X_i$ at time step $t$. Then,
\begin{equation}\label{eq:dep_decomp}
    \begin{aligned}
    &\mcr_T=\sum_{i=1}^K\Delta_i\mathbb E[n_i(T)] =\sum_{i=1}^K\Delta_i \sum_{t=1}^{T}\mathbb P\left(A_{t}=i\right)\\
    \leq &\sum_{i=1}^K\Delta_i \sum_{t=0}^{T-1}\Big[\mathbb P\left(A_{t+1}=i|S(i,t)\in \mathcal B_i,S^*(t)\in \mathcal B^*\right)+ \mathbb P\left(S(i,t)\notin \mathcal B_i\right)+ \mathbb P\left(S^*(t)\notin \mathcal B^*\right)\Big].
\end{aligned}
\end{equation}

Define 
\begin{align*}
   \widetilde{\eta}_i^2&=\max_{S\in\mathcal B_i}\max\left\{ \frac{9\zeta_{i,S}^2}{\gamma_{\min}(\Sigma)},\tilde\sigma_i^2\right\},\\
    \widetilde\phi^2 &= \max_i\max_{S\in\mathcal B_i}\max\left\{\frac{9\zeta_{i,S}^2}{\gamma_{\min}(\Sigma)\tilde\sigma_i^2},\frac{25\tilde\sigma_i^2\gamma_{\max}(\Sigma)}{9\zeta_{i,S}^2}\right\}.
\end{align*}
We have proved that under the correct backdoor adjustment sets, the upper bound of $\mcr_T$ can be expressed as in \eqref{eq:upperbound_DAG_dep}, i.e., the first term is bounded. Due to sample splitting, the bound \eqref{eq:upperbound_DAG_dep} is applicable here.

For any $i$ and $t$, since
\begin{align*}
    \left\{0\in CI_i(t,\pa(i))\right\}\cap \left\{d(0, CI_i(t,S')>0, \forall S'\in\mathcal N_i\setminus\mathcal B_i\right\}\cap \left\{\widehat{M}_i=M_i\right\}\implies S\in\mcb_i,
\end{align*}
we have
\begin{align*}
     \mathbb P\left(S(i,t)\notin \mathcal B_i\right)\leq \mathbb P\left(0\notin CI_i(t,\mathrm{pa}(i))\right)+\mathbb P\left(\widehat M_i\neq M_i\right) + \sum_{S'\in\mathcal N_i\setminus\mathcal B_i}\mathbb P\left(d(0,CI_i(t,S')) = 0\right).
\end{align*}

 Re-define the following constants:
 \begin{equation}\label{eq:notations_rep}
     \begin{aligned}
    &\psi^2=\max_i\max_{S'\in \mathcal N_i\setminus\mathcal B_i}\max\left\{\frac{9\zeta_{i,S'}^2}{\gamma_{\min}(\Sigma)},\widetilde\sigma_i^2\right\},\\
    &\bar\alpha=\min_i\frac{\sqrt{\gamma_{\min}(\Sigma)}\bar\beta_i}{6\sqrt{3}\bar\sigma_i},\\
    &\alpha=\bar\alpha/2,\\
    &s_0=\max_{i,S\in \mathcal B_i}|S|.
\end{aligned}
 \end{equation}
For any $t\geq \ell_0$, we have the following bound from Lemma~\ref{thm:rho},
\begin{equation}\label{eq:rho_one_side}
    \mbp \left(\exists i,j\in V, i\neq j, \rho_{ij}(t)\geq\frac{9}{\gamma_{\min}(\Sigma)}\right)\leq \mathbb P\left( \frac{\sigma_{\min}(\widetilde{\mathbf X})}{\sqrt{N_o(t)}}\leq\frac{1}{3}\gamma_{\min}(\sqrt{\Sigma})\right)\leq\exp\left(-\frac{N_o(t)}{18}\right).
\end{equation}
Incorporating Lemma~\ref{thm:CI_pa}, \ref{thm:CI_S}, and \ref{thm:regression_neighbor} and averaging over $\widetilde{\mathbf X}_{-i}$, we have that for any $i$ and $t\geq \ell_0$,
\begin{equation}\label{eq:bound_S_B}
    \begin{aligned}
     & \mbp\left(S(i,t)\notin \mathcal B_i\right)\\
      \leq & \mbp\left(0\notin CI_i(\mathrm{pa}(i))\middle|\widetilde{\mathbf X}_{-i}\right)+\mbp\left(\widehat M_i\neq M_i\middle|\widetilde{\mathbf X}_{-i}\right) + \sum_{S'\in\mathcal N_i\setminus\mathcal B_i}\mbp\left(d(0,CI_i(S')) = 0\middle|\widetilde{\mathbf X}_{-i}\right)\\
      &\quad + \mathbb P\left( \frac{\sigma_{\min}(\widetilde{\mathbf X})}{\sqrt{N_o(t)}}\leq\frac{1}{3}\gamma_{\min}(\sqrt{\Sigma})\right)\\
     \leq & 3t^{-2}+\exp\left( -\frac{N_o(t)-s_0-1}{9}\right) +s_2\left(3t^{-2}+ \exp\left(-\frac{N_o(t)-p}4\right)\right)+\exp\left(-\frac{N_o(t)}{18}\right)\\ 
    &  + (p-1) \left(2\exp\left( - \frac{\bar\alpha^2N_o(t)}{4}\right) +\exp\left(-\frac{N_o(t)-p}{16}\right)\right)+s_1 \exp\left(-\frac{N_o(t)-p}{4}\right)\\
    & \leq   3(s_2+1)t^{-2} + (s_1+s_2+p+1)\exp\left(-\frac{n_0+nt-p}{18}\right) + 2(p-1) \exp\left( - \alpha^2(n_0+nt)\right).
\end{aligned}
\end{equation}
Therefore,
\begin{equation}\label{eq:bound_dep_dag_id}
    \begin{aligned}
     &\sum_{t=\ell}^T \mathbb P\left(S(i,t)\notin \mathcal B_i\right)\\
      \leq & \frac{(s_2+1)\pi^2}{2} + \frac{18(s_1+s_2+p+1)}{n}\exp\left(-\frac{n_0+n\ell-p}{18}\right)+ \frac{2(p-1)}{\alpha^2 n}\exp\left( - \alpha^2 (n_0+n\ell)\right).
\end{aligned}
\end{equation}

Let $c\geq 4\sqrt{2}\widetilde\phi$ and $c_3\geq \max\left\{64, \frac{32\psi^2}{\delta^2}\right\}$. Aside from the condition in \eqref{eq:min_int_1}, assume further that $\ell\geq\left(\frac{c_3}{n}\right)^2$ in order to satisfy the condition on $t$ in Lemma~\ref{thm:CI_S}. The upper bound still has the form as in \eqref{eq:upperbound_DAG_dep}, i.e.,
\begin{equation*}
   \log T\left(\frac{20c^2}{n+m}\sum_{i:\mu_i<\mu^*}\frac{\widetilde\eta_i^2}{\Delta_i}+\frac{c_3\sum_{i=1}^K\Delta_i}{m}\right)+C_3\left(\sum_{j=1}^K\Delta_j\right).
\end{equation*}
Since $p=K+1$,
\begin{equation}\label{eq:C3}
    \begin{aligned}
        C_3&=\left(\frac{c_3}{n}\right)^2+C_1+\sum_{t=\ell_0}^T \mbp\left(S(i,t)\notin \mathcal B_i\right)+\sum_{t=\ell_0}^T \mbp \left(S^*(t)\notin \mathcal B^*\right)\\
        &=\left(\frac{c_3}{n}\right)^2+\eqref{eq:C1} +2\times\eqref{eq:bound_dep_dag_id}\\
        &=\ell_0+1+\pi^2\left(s_2+\frac{4}{3}+\frac{1}{3}\exp\left(-\frac{n_0-s_0-2}{16}\right)\right)+\left(\frac{c_3}{n}\right)^2\\
        &\quad +  \frac{4K}{n\alpha^2\exp\left(\alpha^2(n_0+n\ell_0)\right)}+\frac{36(s_1+s_2+K+3)}{n\exp\left((n_0+n\ell_0-K-1)/18\right)}.
    \end{aligned}
\end{equation}
\end{proof}

\newpage
\subsection{Proof of Theorem~\ref*{thm2}}
\begin{proof}
We start with the inequality \eqref{eq:indep_inequality} in the proof of Theorem~\ref{thm:thm2_DAG}:
\begin{align*}
    \mcr_T&\leq 2\ell+2\sum_{t=\ell}^{T-1}\mbp \left(\mu^*>U_t(A^*)\right)+\sum_{t=\ell+1}^T\mbe\left(U_{t-1}(A_t)-\mu_{A_t}\right).
\end{align*}
If DAG is unknown, we have the following bounds:
\begin{align*}
     &\mbp \left(\mu^*>U_t(A^*)\right)
     \leq \mbp \left(\mu^*>U_t(A^*)|S^*(t)\in \mathcal B^{*}\right)+ \mbp \left(S^*(t)\notin \mathcal B^*\right),\\
    &\mbe\left(U_{t-1}(A_t)-\mu_{A_t}\right)\leq  \mbe\left(U_{t-1}(A_t)-\mu_{A_t}|S(A_t,t-1)\in \mathcal B_{A_t}\right) +2\mbp \left(S(A_t,t-1)\notin \mathcal B_{A_t}\right).
\end{align*}
Let $\widetilde\eta^2=\max_i\widetilde\eta_i^2$, $c\geq 4\widetilde\phi$, $c_3\geq \max\left\{32, \frac{32\psi^2}{\delta^2}\right\}$, and $\ell\geq \left\lceil\frac{9\operatorname{tr}(\Sigma)}{n\gamma_{\min}(\Sigma)}\right\rceil=\ell_0$. In Section~\ref{sec:case_dep_DAG_unknown}, we have proved an upper bound \eqref{eq:bound_S_B} for $\mathbb P(S(i,t)\notin \mathcal B_i)$, for all $i$ and $t\geq \ell$. Combined with the upper bound proved under the DAG given setting in Section~\ref{sec:case_indep_DAG_given}, we have the following result
\begin{equation*}
    \begin{aligned}
        \mcr_T &\leq 2c\widetilde\eta\sqrt{\log T}\min\left\{\frac{\sqrt{(n+m)KT+K^2n_0}-\sqrt{K^2n_0}}{n+m},\frac{\sqrt{n_0+nT}-\sqrt{n_0}}{n}\mathbbm 1\{n\geq 1\}\right\}\\
        &\quad +\frac{2Kc_3\log T}{m}
        +C_4,
    \end{aligned}
\end{equation*}
where 
\begin{align*}
    C_4 &= 2\left(\frac{c_3}{n}\right)^2+C_2+2\sum_{t=\ell}^{T-1}\mbp \left(S^*(t)\notin \mathcal B^*\right)+2\sum_{t=\ell+1}^{T}\mbp \left(S(A_t,t-1)\notin \mathcal B_{A_t}\right)\\
    &=2\left(\frac{c_3}{n}\right)^2+\eqref{eq:C2}+4\times\eqref{eq:bound_dep_dag_id}\\
    &=2\ell_0+\left(2s_2+\frac{7}{3}\right)\pi^2+2K+2\left(\frac{c_3}{n}\right)^2+\frac{\pi^2}{6}\exp\left(-\frac{n_0-s_0-2}{16}\right)\\
    &\quad +\frac{72(s_1+p+s_2+2)}{n}\exp\left(-\frac{n_0+n\ell_0-p}{18}\right)+ \frac{8(p-1)}{\alpha^2 n}\exp\left( - \alpha^2 (n_0+n\ell_0)\right).
\end{align*}
Replacing $p$ by $K+1$, we have 
\begin{align*}
    C_4 &=\pi^2\left(2s_2+\frac{7}{3}+\frac{1}{6}\exp\left(-\frac{n_0-s_0-2}{16}\right)\right) +2\ell_0+2\left(\frac{c_3}{n}\right)^2+2K\\
&\quad +\frac{8K}{n\alpha^2\exp\left(\alpha^2(n_0+n\ell_0)\right)}+\frac{72(s_1+s_2+K+3)}{n\exp\left((n_0+n\ell_0-K-1)/18\right)}
\end{align*}
\end{proof}

\section{Proofs of results in Section \ref*{sec:confounder_case}}\label{appendix:DAG_confounders}

\subsection{Proof of Lemma \ref*{proposition:mb}}\label{sec:proposition_admg}

\begin{proof}
Let $\prec$ be a topological order of the ADMG $\mathcal G$, and denote by $[v]$ the rank of $v$ under $\prec$. Define $V_{[i]}:=\{v\in V: [v]\le [i]\}$ and let $\mathcal G_{[i]}$ be the induced subgraph of $\mathcal G$ on $V_{[i]}$. If $\mathrm{dis}(i)\cap \mathrm{de}(i)=\{i\}$, then $i$ is fixable in $\mathcal G(V)$ \citep{Richardson_2023}. According to Definition~19 in \cite{Richardson_2023}, for a fixable vertex $i$, we have 
\begin{equation}\label{eq:truncated}
    p(x_{V\setminus\{i\}}|do(x_i))=\frac{p(x_V)}{p(x_i|x_{\mathrm{mb}(i,\mathcal G_{[i]})})},
\end{equation}
where $\mathrm{mb}(i,\mathcal G_{[i]})$ is the Markov blanket of the node $i$ in the induced graph $\mathcal{G}_{[i]}$.

Since $\mathrm{dis}(i)\cap \mathrm{de}(i)=\{i\}$, the node $i$ can be sorted as the last vertex in its district. Therefore, under this topological sorting, $\mathrm{mb}(i,\mathcal G_{[i]})=\mathrm{mb}(i)$. Then we have
%\qzcmt{first equality needs some explanation/citation}
\begin{align*}
p\left(x_{V\setminus \{i\}} \middle| do(x_i)\right) & =\frac{p(x_V)}{p\left(x_i \middle| x_{\mathrm{mb}(i)}\right)} \\
&=\frac{p\left(x_V\right)}{p(x_i,x_{\mathrm{mb}(i)})/p\left(x_{\mathrm{mb}(i)}\right)} =\frac{p\left(x_V\right)}{p(x_i, x_{\mathrm{mb}(i)})} p\left(x_{\mathrm{mb}(i)}\right) \\
& =p\left(x_{V\setminus \{i, \mathrm{mb}(i)\}}|x_i,x_{\mathrm{mb}(i)}\right) p\left(x_{\mathrm{mb}(i)}\right).
\end{align*}
Therefore
\begin{align*}
    p(y\mid do(x_i))&=\int p\left(x_{V\setminus \{i\}} \middle| do(x_i)\right) \mathrm{d} x_{\overline V\setminus \{i\}} \\
    &=\int p\left(x_{V\setminus \{i, \mathrm{mb}(i)\}}\middle| x_i,x_{\mathrm{mb}(i)}\right) p\left(x_{\mathrm{mb}(i)}\right)\mathrm{d}x_{\overline V\setminus \{i\}} \\
    &=\int p\left(x_{\mathrm{mb}(i)}\right)\left[\int p\left(x_{V\setminus \{i, \mathrm{mb}(i)\}}\middle| x_i,x_{\mathrm{mb}(i)}\right)dx_{\overline V\setminus \{i, \mathrm{mb}(i)\}}\right]\mathrm{d}x_{\mathrm{mb}(i)}\\
    &=\int p\left(x_{\mathrm{mb}(i)}\right)p\left(y\middle| x_i,x_{\mathrm{mb}(i)}\right)\mathrm{d}x_{\mathrm{mb}(i)},
\end{align*}
which shows that $\mathrm{mb}(i)$ is a valid adjustment set for $p(y\mid do(x_i))$.
%\qzcmt{for Gaussian, summation should be changed to integral. Check if other regularity conditions are needed, say nonzero denominators (probably all implied by positive densities}
\end{proof}

\subsection{Proof of Theorem~\ref*{thm3}}

\begin{proof}
    The proof starts from 
\begin{align*}
    \mcr_T=\sum_{i=1}^K\Delta_i\mathbb E[n_i(T)] =\sum_{i=1}^K\Delta_i \sum_{t=0}^{T-1}\mathbb P\left(A_{t+1}=i\right).
\end{align*}
Note that $\mathbb P(A_{t+1}=i)$ can be upper bounded in the same way as \eqref{eq:dep_decomp}. As long as the identification of backdoor adjustment sets is correct, we can decompose 
\begin{align*}
    \sum_{t=0}^{T-1}\mathbb P\left(A_{t+1}=i\right)&\leq \ell+\sum_{t=\ell}^{T-1} \sum_{u=1}^{t} \sum_{u_{i}=\ell}^{t}\Big[\mbp\left(\widehat\mu_u^{*}(t) \leq \mu^{*}-\widehat\sigma_u^*(t)c_{t}\right)+\mbp\left(\widehat\mu_{i, u_{i}}(t) \geq \mu_{i}+\widehat\sigma_{i,u_i}(t)c_{t}\right)\\
& \quad  +\mbp \left(\mu^{*}<\mu_{i}+2 \widehat\sigma_{i,u_i}(t)c_{t}\right)\Big],
\end{align*}
by similar reasoning as for \eqref{eq:proof_1}. Let 
\[
\widetilde\phi^2 = \underset{i\in\mathcal I_0}{\max}\underset{S\in\mathcal B_i}{\max}\max\left\{\frac{9\zeta_{i,S}^2}{\gamma_{\min}(\Sigma)\tilde\sigma_i^2},\frac{25\tilde\sigma_i^2\gamma_{\max}(\Sigma)}{9\zeta_{i,S}^2}\right\} \quad\text{and}\quad s_0= \underset{i\in\mathcal I_0,S\in\mathcal B_i}{\max} |S|.
\] 

If a node $i\in\mathcal I_0$, i.e., $\mathcal B_i\neq \emptyset$, as proved in \eqref{eq:bound_sigma2} of Section~\ref{appendix:DAGgiven}, when $u_i\geq \ell\geq \frac{20c^2\widetilde\eta_i^2\log t}{(m+n)\Delta_i^2}$ and $u_i\geq \frac{16\log t}{n+m}$, we have
\begin{equation}\label{eq:part1}
\mathbb P\left( \mu^{*}<\mu_{i}+2 \widehat\sigma_{i,u_i}(t)c_{t}\middle|\mathbf X_{\overline V}\right)\leq t^{-4}\exp\left(-\frac{n_0-s_0-2}{4}\right)\coloneq F_1,
\end{equation}
When $u_i\geq \frac{64\log t}{n+m}$ and $c\geq 4\sqrt{2}\widetilde{\phi}$, as proved in \eqref{eq:bound_mui} of Section~\ref{appendix:DAGgiven}, we have
\begin{equation}\label{eq:part2}
    \mathbb P\left(\widehat\mu_{i, u_{i}}\geq \mu_{i}+\widehat\sigma_{i,u_i}c_{t}\middle|\mathbf X_{\overline V}\right)\leq t^{-4}\left(1+\frac{1}{2}\exp\left(-\frac{n_0-s_0-2}{16}\right)\right)\coloneq F2.
\end{equation}

If $i\in \mathcal I_1$, when $u_i\geq\ell\geq \frac{10c^2\widetilde\sigma_i^2\log t +1}{m\Delta_i^2}$ and $u_i\geq \frac{16\log t + 1}{m}$, we have
\begin{equation}\label{eq:part1_int}
    \begin{aligned}
    \mathbb P\left( \mu^{*}<\mu_{i}+2 \widehat\sigma_{i,u_i}(t)c_{t}\middle|\mathbf X_{\overline V}\right)&= \mathbb P\left(\widehat\sigma_{i,u_i}^2(t)>\frac{\Delta_i^2}{4c^2\log(t)}\middle|\mathbf X_{\overline V}\right)\\
    &\leq \mathbb P\left(\chi_{mu_i-1}^2>\frac{(mu_i)(mu_i-1)\Delta_i^2}{4c^2\widetilde\sigma_i^2\log t}\right)\\
    &\leq t^{-4}\coloneq F_3.
\end{aligned}
\end{equation}
When $u_i\geq \frac{16\log t + 1}{m}$ and $c\geq 4$, we have %\qzedit{[Need conditioning on $\mathbf X_{\overline V}$ below?]}
\begin{equation}\label{eq:part2_int}
    \begin{aligned}
    \mathbb P(\widehat\mu_{i, u_{i}}\geq \mu_{i}+\widehat\sigma_{i,u_i}c_{t}\mid \mathbf X_{\overline V})=\mathbb P\left(t_{mu_i-1}\geq c\sqrt{\log t}\right)\leq t^{-c^2/4}+\frac{1}{2}e^{-\frac{mu_i-1}{16}}\leq \frac{3}{2}t^{-4}\coloneq F_4.
\end{aligned}
\end{equation}
Aside from the above bounds, for $t\geq \ell\geq \ell_0$, we also have
\begin{equation}\label{eq:part3}
\begin{aligned}
    \mathbb P\left(\rho_i(t,S)\geq\frac{9}{\gamma_{\min}(\Sigma) } \text{ or } \rho_i(t,S)\leq\frac{9}{25\gamma_{\max}(\Sigma)} \text{ for some }i\in\overline V, S\subseteq\overline V\setminus\{i\} \right) \\
    \leq  2\exp\left(-\frac{N_o(t)}{18}\right)\coloneq F_5.
\end{aligned}
\end{equation}
Now we continue to bound the probability of $S(i,t)\notin\mathcal B_i$ for any arm $i$ including the optimal arm. For any node $i\in\mathcal I_0$, $\mathcal B_i\neq \emptyset$ and
\begin{align*}
    \mathbb P\left(S(i,t)\notin \mathcal B_i\right)&\leq  \mathbb P\left(S(i,t)\notin \mathcal B_i\text{ and }\widehat M_i= M_i \right)+ \mathbb P\left(\widehat M_i\neq M_i \right)\\
    &\leq \mathbb P\left(\exists S\in \mathcal N_i\setminus\mathcal B_i, 0\in CI_i(t, S)\right)+ \mathbb P\left(\widehat M_i\neq M_i \right)\\
    &\leq \sum_{S\in \mathcal N_i\setminus\mathcal B_i}\mathbb P\left(0\in CI_i(t, S)\right)+ \mathbb P\left(\widehat M_i\neq M_i \right).
\end{align*}
For $i\in\mathcal I_1$, $\mathcal B_i=\emptyset$. The probability of incorrect adjustment is bounded by
\begin{align*}
    \mathbb P\left(\widehat{\mathcal B}_i\neq \emptyset\right)&\leq  \mathbb P\left(\widehat{\mathcal B}_i\neq \emptyset\text{ and }\widehat M_i= M_i \right)+ \mathbb P\left(\widehat M_i\neq M_i \right)\\
    &\leq \mathbb P\left(\exists S\in \mathcal N_i\setminus\mathcal B_i, 0\in CI_i(t, S)\right)+ \mathbb P\left(\widehat M_i\neq M_i \right)\\
    &\leq \sum_{S\in \mathcal N_i\setminus\mathcal B_i}\mathbb P\left(0\in CI_i(t, S)\right)+ \mathbb P\left(\widehat M_i\neq M_i \right).
\end{align*}
It is seen that the upper bound of incorrect backdoor adjustment set sampled is the same for $i\in\mathcal I_0$ and $i\in\mathcal I_1$. Incorporating the results in Lemmas~\ref{thm:CI_S} and \ref{thm:regression_neighbor}, with parameters $c_2=2\sqrt{3}$ and $c_3\geq \max\{32\psi^2/\delta^2,8\}$, when $N_i(t)\geq c_3\log t+1$ and $N_o(t)\geq p$, for any $t\geq \max\left\{\frac{9\operatorname{tr}(\Sigma)}{n\gamma_{\min}(\Sigma)},(c_3/n)^2\right\}$,  we have 
\begin{equation}\label{eq:bound_final_id}
\begin{aligned}
    &\mathbb P\left(\text{Incorrect }S(i,t)\right)\leq \text{RHS of } [\eqref{eq:dep_dag_b2}+\eqref{eq:dep_dag_b3}+\eqref{eq:rho_one_side}]\\
    =& 3s_2t^{-2} + (s_1+s_2+K+1)\exp\left(-\frac{n_0+nt-K-1}{18}\right) + 2K \exp\left( - \alpha^2(n_0+nt)\right)\\
    \coloneq &F_6.
\end{aligned}
\end{equation}
Now we summarize all conditions as follows for the upper bound on the cumulative regret.

For $i\in\mathcal I_0$, we need $\ell\geq\max\left\{ \frac{c_3\log t+1}{m}, \frac{20c^2\widetilde\eta_i^2\log t}{(m+n)\Delta_i^2}, \frac{9\operatorname{tr}(\Sigma)}{n\gamma_{\min}(\Sigma)}, 2, \left(\frac{c_3}{n}\right)^2\right\}$, where $c\geq 4\sqrt{2}\widetilde{\phi}$ and $c_3\geq \max\left\{64,32\psi^2/\delta^2\right\}$. Thus, it suffices to have
\begin{align*}
    \ell\geq \frac{20c^2\widetilde\eta_i^2\log T}{(m+n)\Delta_i^2}+\frac{c_3\log T}{m}+\lceil\frac{9\operatorname{tr}(\Sigma)}{n\gamma_{\min}(\Sigma)}\rceil+\left(\frac{c_3}{n}\right)^2+1.
\end{align*}
If the optimal arm $a^*\in\mathcal I_0$, for $ t\geq \ell$,
\begin{align*}
    \mathbb P(A_{t+1}=i)&\leq\mathbb P\left(A_{t+1}=i|S(i,t)\in \mathcal B_i,S^*(t)\in \mathcal B^*\right)+ \mathbb P\left(S(i,t)\notin \mathcal B_i\right)+ \mathbb P\left(S^*(t)\notin \mathcal B^*\right)\\
    &\leq t^2\left(F_1 +2F2\right)+ F_5 + 2F_6\\
    &=2t^{-2}\left(1+\exp\left(-\frac{n_0-s_0-2}{16}\right)\right)+ F_5 + 2F_6.
\end{align*}
If the optimal arm $a^*\in\mathcal I_1$, for $t\geq \ell$,
\begin{align*}
    \mathbb P(A_{t+1}=i)&\leq\mathbb P\left(A_{t+1}=i|S(i,t)\in \mathcal B_i,S^*(t)\in \mathcal B^*\right)+ \mathbb P\left(S(i,t)\notin \mathcal B_i\right)+ \mathbb P\left(S^*(t)\notin \mathcal B^*\right)\\
    &\leq t^2\left(F_1+F_2+F_4\right)+ F_5 + 2F_6\\
    &=t^{-2}\left(\frac{5}{2}+\frac{3}{2}\exp\left(-\frac{n_0-s_0-2}{16}\right)\right)+ F_5 + 2F_6.
\end{align*}
If $i\in\mathcal I_1$, we need $\ell\geq \max\left\{\frac{10c^2\widetilde\sigma_i^2\log t +1}{m\Delta_i^2},  \frac{c_3\log t+1}{m}, \frac{9\operatorname{tr}(\Sigma)}{n\gamma_{\min}(\Sigma)}, \left(\frac{c_3}{n}\right)^2\right\}$, i.e.,
\begin{align*}
    \ell\geq \frac{10c^2\widetilde\sigma_i^2\log T }{m\Delta_i^2}+\frac{c_3\log T}{m}+\frac{9\operatorname{tr}(\Sigma)}{n\gamma_{\min}(\Sigma)}+ \left(\frac{c_3}{n}\right)^2+1.
\end{align*}
If $a^*\in\mathcal I_0$, for $ t\geq \ell$,
\begin{align*}
    \mathbb P(A_{t+1}=i)&\leq\mathbb P\left(A_{t+1}=i|\widehat{\mathcal B}_i=\emptyset,S^*(t)\in \mathcal B^*\right)+ \mathbb P\left(\widehat{\mathcal B}_i\neq\emptyset\right)+ \mathbb P\left(S^*(t)\notin \mathcal B^*\right)\\
    &\leq t^2\left(F_2+F_3+F_4\right)+ F_5 + 2F_6\\
    &=t^{-2}\left(\frac{7}{2}+\frac{1}{2}\exp\left(-\frac{n_0-s_0-2}{16}\right)\right)+ F_5 + 2F_6.
\end{align*}
If $a^*\in\mathcal I_1$, for $t\geq \ell$,
\begin{align*}
    \mathbb P(A_{t+1}=i)&\leq\mathbb P\left(A_{t+1}=i|\widehat{\mathcal B}_i=\emptyset,\widehat B^*=\emptyset\right)+ \mathbb P\left(\widehat{\mathcal B}_i\neq\emptyset\right)+ \mathbb P\left(\widehat B^*\neq\emptyset\right)\\
    &\leq t^2\left(F_3+2F_4\right)+ F_5 + 2F_6\\
    &=4t^{-2}+ F_5 + 2F_6.
\end{align*}
Therefore, for any $i$ and $ t\geq \ell$, we have
\begin{align*}
     \mathbb P(A_{t+1}=i)&\leq t^{-2}\left(4+2\exp\left(-\frac{n_0-s_0-2}{16}\right)\right)+F_5 + 2F_6\\
     &=t^{-2}\left(6s_2+4+2\exp\left(-\frac{n_0-s_0-2}{16}\right)\right)  + 4K \exp\left( - \alpha^2(n_0+nt)\right)\\
     &+2\exp\left(-\frac{n_0+nt}{18}\right) + 2(s_1+s_2+K+1)\exp\left(-\frac{n_0+nt-K-1}{18}\right)\\
     &\leq t^{-2}\left(6s_2+4+2\exp\left(-\frac{n_0-s_0-2}{16}\right)\right)\\
      &+\frac{2(s_1+s_2+K+2)}{\exp\left((n_0+nt-K-1)/18\right)} + 4K \exp\left( - \alpha^2(n_0+nt)\right).
\end{align*}
Then
\begin{align*}
    \sum_{t\geq \ell} \mathbb P(A_{t+1}=i)&\leq \pi^2\left(s_2+\frac{2}{3}+\frac{1}{3}\exp\left(-\frac{n_0-s_0-2}{16}\right)\right)\\
      &+\frac{36(s_1+s_2+K+2)}{n\exp\left((n_0+n\ell-K-1)/18\right)} + \frac{4K}{n\alpha_2^2\exp\left(\alpha_2^2(n_0+n\ell)\right)}.
\end{align*}
Incorporating the condition for $\ell$, we have that for $i\in\mathcal I_0$,
\begin{align*}
    \mathbb E[n_i(T)]\leq \frac{20c^2\widetilde\eta_i^2\log T}{(m+n)\Delta_i^2}+\frac{c_3\log T}{m}+C_5,
\end{align*}
and for $i\in\mathcal I_1$,
\begin{align*}
    \mathbb E[n_i(T)]\leq \frac{10c^2\widetilde\sigma_i^2\log T }{m\Delta_i^2}+\frac{c_3\log T}{m}+C_5,
\end{align*}
where 
\begin{align*}
    C_5&=\ell_0+ \left(\frac{c_3}{n}\right)^2+1+\pi^2\left(s_2+\frac{2}{3}+\frac{1}{3}\exp\left(-\frac{n_0-s_0-2}{16}\right)\right)\\
      &+\frac{36(s_1+s_2+K+2)}{n\exp\left((n_0+n\ell_0-K-1)/18\right)} + \frac{4K}{n\alpha_2^2\exp\left(\alpha_2^2(n_0+n\ell_0)\right)}.
\end{align*}
Thus, the final cumulative regret bound is
\begin{align*}
    \mcr_T
    \leq \frac{20c^2\log T}{m+n}\sum_{i\in\mathcal I_0}\frac{\widetilde\eta_i^2}{\Delta_i}+\frac{10c^2\log T }{m}\sum_{i\in\mathcal I_1}\frac{\widetilde\sigma_i^2}{\Delta_i}+\left(\frac{c_3\log T}{m}+C_5\right)\left(\sum_{i}\Delta_i\right).
\end{align*} 
\end{proof}

\subsection{Proof of Theorem~\ref*{thm4}}
\begin{proof}
As shown in \eqref{eq:indep_inequality},
    \begin{equation}\label{eq:indep_ineq_conf}
    \mcr_T\leq 2\ell+2\sum_{t=\ell}^{T-1}\mbp \left(\mu^*>U_t(A^*)\right)+\sum_{t=\ell+1}^T\mbe\left(U_{t-1}(A_t)-\mu_{A_t}\right),
\end{equation}
where 
\begin{align*}
      &\mbp \left(\mu^*>U_t(A^*)\right)
     \leq \mbp \left(\mu^*>U_t(A^*)|S^*(t)\in \mathcal B^{*}\right)+ \mbp \left(\text{Incorrect }S^*(t)\right),\\
   &\mbe\left(U_{t-1}(A_t)-\mu_{A_t}\right)\leq  \mbe\left(U_{t-1}(A_t)-\mu_{A_t}|S(A_t,t-1)\in \mathcal B_{A_t}\right)+2\mbp \left(\text{Incorrect }S(A_t,t-1)\right).
\end{align*}
If $A^*\in \mathcal I_0$, then
\begin{align*}
    \mbp \left(\mu^*>U_t(A^*)|S^*(t)\in \mathcal B^{*}\right)\leq F_2\leq t^{-2}\left(1+\frac{1}{2}\exp\left(-\frac{n_0-s_0-2}{16}\right)\right).
\end{align*}
If $A^*\in \mathcal I_1$, then
\begin{align*}
    \mbp \left(\mu^*>U_t(A^*)| \widehat{\mathcal B^{*}}=\emptyset\right)\leq F_4\leq \frac{3}{2}t^{-2}.
\end{align*}
Therefore,
\begin{equation}\label{eq:part1_indep}
\begin{aligned}
     \sum_{t=\ell}^{T-1} \mbp \left(\mu^*>U_t(A^*)|S^*(t)\in \mathcal B^{*}\right)&\leq \sum_{t=\ell}^{T-1}  t^{-2}\left(\frac{3}{2}+\frac{1}{2}\exp\left(-\frac{n_0-s_0-2}{16}\right)\right)\\
     &\leq \frac{\pi^2}{6}\left(\frac{3}{2}+\frac{1}{2}\exp\left(-\frac{n_0-s_0-2}{16}\right)\right).
\end{aligned}
\end{equation}
Let $\mathcal T_a=\{t:A_t=a, \ell +1\leq t\leq T\}$. Then, 
\begin{align*}
    \sum_{t=\ell+1}^T\mbe\left(U_{t-1}(A_t)-\mu_{A_t}|S(A_t,t-1)\in \mathcal B_{A_t}\right)&\leq \sum_{a\in \mathcal I_0}\sum_{t\in \mathcal T_a}\mbe\left(U_{t-1}(a)-\mu_{a}|S(a,t-1)\in \mathcal B_{a}\right)\\
    &\quad +\sum_{a\in \mathcal I_1}\sum_{t\in \mathcal T_a}\mbe\left(U_{t-1}(a)-\mu_{a}|S(a,t-1)\in \mathcal B_{a}\right).
\end{align*}
If $a\in\mathcal I_0$, then 
\begin{align*}
   \mbe\left(U_{t-1}(a)-\mu_{a}|S(a,t-1)\in \mathcal B_{a}\right)&\leq c\widetilde\eta\sqrt{\frac{\log t}{N_o(t)+N_a(t)}}.
\end{align*}
Similar to \eqref{eq:indep_b3} and by Cauchy-Schwarz inequality,
\begin{align*}
    \sum_{a\in \mathcal I_0}\sum_{t\in \mathcal T_a}\mbe\left(U_{t-1}(a)-\mu_{a}|S(a,t-1)\in \mathcal B_{a}\right)& \leq \frac{2c\widetilde{\eta}\sqrt{\log T}}{n+m}\sum_{a\in\mathcal I_0}\left(\sqrt{n_0+(m+n)n_a(T)}-\sqrt{n_0}\right)\\
    % \leq &\quad \frac{2c\widetilde{\eta}\sqrt{\log T}}{n+m}\left(\sqrt{|\mathcal I_0|\left(\sum_{a\in\mathcal I_0}(n+m)n_a(T)\right)+n_0|\mathcal I_0|^2}-\sqrt{n_0}|\mathcal I_0|\right)\\
    % = &\quad\frac{2c\widetilde{\eta}\sqrt{\log T}}{n+m}\left(\sqrt{|\mathcal I_0|(n+m)n_{\mathcal I_0}(T)+n_0|\mathcal I_0|^2}-\sqrt{n_0}|\mathcal I_0|\right)\\
    &\leq \frac{2c\widetilde{\eta}\sqrt{\log T}}{n+m}\left(\sqrt{|\mathcal I_0|(n+m)T+n_0|\mathcal I_0|^2}-\sqrt{n_0}|\mathcal I_0|\right).
\end{align*}
If $a\in\mathcal I_1$, then 
\begin{align*}
   \mbe\left(U_{t}(a)-\mu_{a}|S(a,t-1)\in \mathcal B_{a}\right)\leq c\widetilde\sigma\sqrt{\frac{\log t}{N_a(t)}},
\end{align*}
and by Cauchy-Schwarz inequality,
\begin{align*}
    \sum_{a\in \mathcal I_1}\sum_{t\in \mathcal T_a}\mbe\left(U_{t-1}(a)-\mu_{a}|S(a,t-1)\in \mathcal B_{a}\right)&\leq \frac{2c\widetilde\sigma\sqrt{\log T}}{m}\sum_{a\in\mathcal I_1}\left(\sqrt{mn_a(T)}\right)\\
    % &\leq \frac{2c\widetilde\sigma\sqrt{\log T}}{m}\sqrt{|\mathcal I_1|\left(\sum_{a\in\mathcal I_1}mn_a(T)\right)}\\
    % &=\frac{2c\widetilde\sigma\sqrt{\log T}}{m}\sqrt{|\mathcal I_1|mn_{\mathcal I_1}(T)}\\
    &\leq \frac{2c\widetilde\sigma\sqrt{\log T}}{m}\sqrt{|\mathcal I_1|mT}.
\end{align*}
Therefore,
\begin{equation}\label{eq:part2_indep}
    \begin{aligned}
         &\sum_{t=\ell+1}^T\mbe\left(U_{t-1}(A_t)-\mu_{A_t}|S(A_t,t-1)\in \mathcal B_{A_t}\right)\\
         \leq& \frac{2c\widetilde{\eta}\sqrt{\log T}}{n+m}\left(\sqrt{|\mathcal I_0|(n+m)T+n_0|\mathcal I_0|^2}-\sqrt{n_0}|\mathcal I_0|\right)+\frac{2c\widetilde\sigma\sqrt{\log T}}{m}\sqrt{|\mathcal I_1|mT}.
    \end{aligned}
\end{equation}
Note that 
\[\ell\geq \max\left\{\frac{9\operatorname{tr}(\Sigma)}{n\gamma_{\min}(\Sigma)}, \left(\frac{c_3}{n}\right)^2, K\left\lceil\frac{c_3\log T+1}{m}\right\rceil\right\}.
\]
Let $\ell_0=\lceil \frac{9\operatorname{tr}(\Sigma)}{n\gamma_{\min}(\Sigma)}\rceil$. Plugging the relevant bounds into \eqref{eq:indep_ineq_conf}, the final cumulative regret bound is
\begin{align*}
   \mcr_T& \leq 2\ell_0+2\left(\frac{c_3}{n}\right)^2+\frac{2Kc_3\log T}{m}+2K \\
   &\quad\quad +\text{RHS of }[2\times \eqref{eq:rho_sum} + 2\times\eqref{eq:part1_indep} +\eqref{eq:part2_indep}]+\sum_{t\geq \ell_0}4F_6 \\
   &\leq 2c\sqrt{\log T}\left(\frac{\widetilde{\eta}\left(\sqrt{|\mathcal I_0|(n+m)T+n_0|\mathcal I_0|^2}-\sqrt{n_0}|\mathcal I_0|\right)}{n+m}+\widetilde\sigma\sqrt{\frac{|\mathcal I_1|T}{m}}\right)\\
   &\quad\quad \quad +\frac{2Kc_3\log T}{m}+C_6,
\end{align*}
where 
\begin{align*}
    C_6&=2\ell_0+2\left(\frac{c_3}{n}\right)^2+2K+\pi^2\left(\frac{1}{2}+2s_2+\frac{1}{6}\exp\left(-\frac{n_0-s_0-2}{16}\right)\right)\\
    & + \frac{8K}{n\alpha^2\exp\left(\alpha^2(n_0+n\ell_0)\right)}+\frac{72(s_1+s_2+K+2)}{n\exp\left((n_0+n\ell_0-K-1)/18\right)}.
\end{align*}
\end{proof}

% \end{document}

\end{document}